\documentclass{article} 
\usepackage{iclr2023_conference,times}
\usepackage{amssymb}
\usepackage{amsmath}
\usepackage{bbm}
\usepackage{algpseudocode}
\usepackage{algorithm}
\usepackage{algcompatible}
\usepackage{stmaryrd}
\usepackage{booktabs}
\usepackage{tabularx}

\usepackage{amsmath,amsfonts,bm}
\usepackage{amsthm}
\usepackage{amssymb}
\usepackage{algorithm}
\usepackage{algpseudocode}
\usepackage{mathrsfs}

\usepackage{graphicx}
\usepackage{subfigure}

\newtheorem{lemma}{Lemma}
\newtheorem{thm}{Theorem}
\newtheorem{rmk}{Remark}
\newtheorem{prpst}{Proposition}
\newtheorem{claim}{Claim}
\newtheorem{definition}{Definition}

\newtheorem{assumption}{Assumption}

\def\bfa{{\boldsymbol a}}
\def\bfb{{\boldsymbol b}}

\def\bfe{{\boldsymbol e}}

\def\bfn{{\boldsymbol n}}
\def\bfo{{\boldsymbol o}}
\def\bfp{{\boldsymbol p}}
\def\bfq{{\boldsymbol q}}
\def\bfr{{\boldsymbol r}}

\def\bfx{{\boldsymbol x}}
\def\bfy{{\boldsymbol y}}
\def\bfz{{\boldsymbol z}}
\def\bfmu{{\boldsymbol \mu}}

\def\bfA{{\boldsymbol A}}

\def\bfF{{\boldsymbol F}}

\def\bfI{{\boldsymbol I}}

\def\bfP{{\boldsymbol P}}
\def\bfQ{{\boldsymbol Q}}
\def\bfR{{\boldsymbol R}}

\def\bfU{{\boldsymbol U}}
\def\bfV{{\boldsymbol V}}
\def\bfW{{\boldsymbol W}}
\def\bfX{{\boldsymbol X}}

\def\bfZ{{\boldsymbol Z}}

\newcommand{\be}{\begin{equation}}
\newcommand{\ee}{\end{equation}}

\def\eqref#1{equation~\ref{#1}}

\def\1{\bm{1}}

\DeclareMathAlphabet{\mathsfit}{\encodingdefault}{\sfdefault}{m}{sl}
\SetMathAlphabet{\mathsfit}{bold}{\encodingdefault}{\sfdefault}{bx}{n}

\usepackage{hyperref}
\usepackage{url}

\usepackage{color}

\title{A Theoretical Understanding of shallow Vision Transformers: Learning, Generalization, and Sample Complexity}

\author{Hongkang Li 
\thanks{Rensselaer Polytechnic Institute. Email: \texttt{lih35@rpi.edu}}
\And
Meng Wang \thanks{Rensselaer Polytechnic Institute. Email: \texttt{wangm7@rpi.edu}}
\And
Sijia Liu \thanks{Michigan State University \& IBM Research.  Email: \texttt{liusiji5@msu.edu }}
\And
Pin-Yu Chen \thanks{IBM Research. Email: \texttt{Pin-Yu.Chen@ibm.com}}
}

\iclrfinalcopy 
\begin{document}

\maketitle

\begin{abstract}

Vision Transformers (ViTs) with self-attention modules have recently achieved great empirical success in many vision tasks. Due to non-convex interactions across layers, however, the theoretical learning and generalization analysis is mostly elusive. Based on a data model characterizing both label-relevant  and label-irrelevant  
tokens, this paper provides the first theoretical analysis of training a shallow ViT, i.e., one self-attention layer followed by a two-layer perceptron, for a classification task. We characterize the sample complexity to achieve a zero generalization error. Our sample complexity bound is positively correlated with the inverse of the fraction of label-relevant tokens, the token noise level, and the initial model error. 
We also prove that  a training process using stochastic gradient descent (SGD) leads to a sparse attention map, which is a formal  verification of the general intuition about the success of attention. Moreover,  this paper indicates that a proper token  sparsification can improve the test performance by removing label-irrelevant and/or noisy tokens, including spurious correlations. 
Empirical experiments on synthetic data and CIFAR-10 dataset justify our theoretical results and generalize to deeper ViTs.  
\end{abstract}

\section{Introduction}
As the backbone of Transformers \citep{VSPU17}, the self-attention mechanism \citep{BCB14} computes the feature representation by  globally modeling long-range interactions within the input.   Transformers have demonstrated tremendous empirical success in numerous areas, including natural language processing \citep{DCLT19, RWCL19, RNSS18, BMRS20}, recommendation system \citep{ZZSF18, CZLH19, SLWP19}, and reinforcement learning \citep{CLRL21, JLL21, ZZG22}. Starting from the advent of Vision Transformer (ViT) \citep{DBKW20}, Transformer-based models \citep{TCDM21, JHYZ21, WXLF21, LLCH21} gradually replace convolutional neural network (CNN) architectures and become prevalent in vision tasks. Various techniques have been developed to train ViT efficiently. 
Among them, token sparsification \citep{PZLH21, RZLL21, LGTS22, THWX22, YVAM22} removes redundant tokens (image patches) of data to improve the computational complexity while maintaining a comparable learning performance. For example, \citet{LGTS22, THWX22} prune tokens following criteria designed based on the magnitude of the  attention map.  
Despite the remarkable   empirical success, one fundamental question about training Transformers is still vastly open, which is
\begin{center}
 \textit{Under what conditions does a Transformer  achieve satisfactory generalization?}
\end{center}
Some recent works analyze Transformers theoretically from the perspective of proved Lipschitz constant of self-attention \citep{VAC20, KPM21}, properties of the neural tangent kernel \citep{HBSN20, Y20} and expressive power and Turing-completeness \citep{DGVU18, YBRR19, BAG20, BPG20, EGKZ22, DCL21, LCW21, CLJ19, LWSB20} with statistical guarantees \citep{SZKS21, WCM21}. 
\citet{LCW21} showed a model complexity for the function approximation of the self-attention module. \citet{CLJ19} provided sufficient and necessary conditions for multi-head self-attention structures to simulate convolution layers. None of these works, however, characterize the generalization performance of the learned model theoretically. Only \cite{EGKZ22} theoretically proved that a single self-attention head can represent a sparse function of the input with a sample complexity for a generalization gap between the training loss and the test loss, but no discussion is provided regarding what algorithm to train the Transformer to achieve a desirable loss. 

\textbf{Contributions}: To the best of our knowledge, this paper provides the first learning and generalization analysis of training a basic shallow Vision Transformer using 
stochastic gradient descent (SGD). 
This paper focuses on a binary classification problem on structured data, where tokens with discriminative 
patterns determine the label from a majority vote, while tokens with non-discriminative  patterns do not affect the labels. We train a ViT containing a self-attention layer followed by a two-layer perceptron using SGD from a  proper initial model. This paper explicitly characterizes the required number of training samples to achieve a desirable generalization performance, referred to as the sample complexity. Our sample complexity bound is positively correlated with the inverse of the fraction of label-relevant tokens, the token noise level, and the error from the initial model, 
indicating a better generalization performance on data with fewer label-irrelevant patterns and less noise  from a better initial model. The highlights of our technical contributions include: 


\textbf{First,} \textbf{this paper proposes a new analytical framework to tackle  the non-convex optimization and generalization for shallow ViTs. } 
Due to the more involved non-convex interactions of learning parameters and diverse activation functions across layers, the ViT model, i.e.,  a three-layer neural network with one self-attention layer, considered in this paper is more complicated to analyze than three-layer CNNs 
considered in \cite{ALL19, AL19}, the most complicated neural network model that has been analyzed so far for across-layer nonconvex interactions. We consider a structured data model with relaxed assumptions from existing models and establish a new analytical framework to   overcome the new technical challenges to handle ViTs. 

\textbf{Second,} \textbf{this paper theoretically 
depicts the evolution of the attention map during the training and characterizes how ``attention'' is paid to different tokens during the training.} Specifically, 
we show that under the structured data model,  the learning parameters of the self-attention module  grow in the direction that projects the data to the label-relevant patterns, 
resulting in an increasingly sparse attention map.  
This insight provides a theoretical justification of the 
magnitude-based token pruning methods such as \citep{LGTS22, THWX22} for efficient learning. 

\textbf{Third,} \textbf{we provide a theoretical explanation for the improved generalization using token sparsification.} We quantitatively  show that if a token sparsification method can remove class-irrelevant and/or highly noisy tokens, 
then the sample complexity is reduced while achieving the same testing accuracy. Moreover, token sparsification can also remove spurious correlations to improve the testing accuracy \citep{LXSC21, ZPXS21}. This insight provides a guideline in designing token sparsification  and 
few-shot learning methods for  Transformer \citep{HLZZ22, GMLT22}. 

\subsection{Background and Related Work}\label{subsec: related_work}
\textbf{Efficient ViT learning.} 
To alleviate the memory and computation burden in training 
\citep{DBKW20, TCDM21, WJZY22}, various acceleration techniques have been developed other than token sparsification. \citet{ZTH21} identifies the importance of different dimensions in
each layer of ViTs and then executes model pruning. \citet{LWHZ21, LZSL22, LYWC22} quantize weights and inputs to compress the learning model. \citet{LZWL22} studies automated progressive learning that 
automatically increases the model capacity on-the-fly. Moreover, modifications of attention modules, such as the network architecture based on local attention \citep{WXLF21, LLCH21, CTWZ21}, can simplify the computation of global attention for acceleration.

\textbf{Theoretical analysis of learning and generalization of neural networks.} One line of research \citep{ZSJB17, FCL20, ZSD17, ZWLC20, ZWXL20, LZW22} analyzes the generalization performance 
when the number of neurons is smaller than the number of training samples. 
The neural-tangent-kernel (NTK) analysis \citep{JGH18, ALL19, ALS19, ADHL19, CG19, ZG19, DZPS19, CCGZ20, LWLC22} considers strongly overparameterized networks  
and eliminates the nonconvex interactions across layers by linearizing the neural network around the initialization. The generalization performance is independent of the feature distribution 
and cannot explain the advantages of self-attention modules. 

\textbf{Neural network learning on structured data.} 
\citet{LL18} provide the generalization analysis of a fully-connected neural network when the data comes from 
separated distributions. \citet{DM20, SWL21, KWLS21, BG21, ZWCL23} study fully connected networks and convolutional neural networks assuming  
that data contains discriminative patterns and background patterns. \citet{AL22} illustrates the robustness of adversarial training by introducing the feature purification mechanism, in which neural networks with non-linear activation functions can memorize the data-dependent features. \citet{WL21} extends this framework to the area of self-supervised contrastive learning. All these works consider one-hidden-layer neural networks without self-attention. 

\textbf{Notations}: Vectors are in bold lowercase, and matrices and tensors are in bold uppercase. Scalars are in normal fonts. Sets are in calligraphy font. For instance, $\bfZ$ is a matrix, and $\bfz$ is a vector. $z_i$ denotes the $i$-th entry of $\bfz$, and $Z_{i,j}$ denotes the $(i,j)$-th entry of $\bfZ$. 
$[K]$ ($K>0$) denotes the set including integers from $1$ to $K$. 
We follow the convention that $f(x)=O(g(x))$ (or $\Omega(g(x))$, $\Theta(g(x)))$ means that $f(x)$ increases at most, at least, or in the order of $g(x)$, respectively. 
\section{Problem Formulation and Learning Algorithm}

We study a 
binary classification problem \footnote{Extension to multi-classification is briefly discussed in Section \ref{subsec: ext_multi}.} 
following the common setup in \citep{DBKW20, TCDM21, JHYZ21}. Given $N$ 
 training samples $\{(\bfX^n, y^n)\}_{n=1}^N$ generated from an unknown distribution $\mathcal{D}$ and a fair initial model, the goal is to find an improved model that maps $\bfX$ to $y$ for any $(\bfX, y)\sim \mathcal{D}$. Here each data point contains $L$ tokens $\bfx_1^n,\ \bfx_2^n,\cdots,\bfx_L^n$, i.e.,  $\bfX^n=[\bfx_1^n,\cdots,\bfx_L^n]\in\mathbb{R}^{d\times L}$, where each token is $d$-dimensional and unit-norm. $y^n\in\{+1,-1\}$ is a scalar. 
 A token can be an image patch \citep{DBKW20}. We consider a general setup that also applies to  token sparsification, where some tokens are set to zero to reduce the computational time. Let  $\mathcal{S}^n \subseteq [L]$ denote the set of indices of remaining tokens in $\bfX^n$ after sparsification. Then $|\mathcal{S}^n|\leq L$, and $\mathcal{S}^n=[L]$ without token sparsification. 

Learning is performed over a basic shallow Vision Transformer, a neural network with a single-head self-attention layer and a two-layer fully connected network, as shown in (\ref{eqn: attention}). This is a simplified model 
of practical Vision Transformers \citep{DBKW20} to avoid unnecessary complications in analyzing the most critical component of ViTs, the self-attention. 

\begin{equation}
    F(\bfX^n)=\frac{1}{|\mathcal{S}^n|}\sum_{l\in\mathcal{S}^n}\bfa_{(l)}^\top\text{Relu}(\bfW_O\bfW_V\bfX^n\text{softmax}({\bfX^n}^\top\bfW_K^\top\bfW_Q\bfx_l^n)),\label{eqn: attention}
\end{equation}

where the queue weights   $\bfW_Q$ in $\mathbb{R}^{m_b\times d}$, the key weights  $\bfW_K$ in $\mathbb{R}^{m_b\times d}$, and the value weights  $\bfW_V$ in $\mathbb{R}^{m_a\times d}$ in the attention unit are multiplied with $\bfX^n$ to obtain the queue vector  $\bfW_Q \bfX^n$, the key vector $\bfW_K \bfX^n$, and the value vector $\bfW_V \bfX^n$, respectively \citep{VSPU17}.   $\bfW_O$ is in $\mathbb{R}^{m\times m_a}$ and $\bfA=(\bfa_{(1)},\bfa_{(2)},\cdots,\bfa_{L})$ where $\bfa_{(l)}\in\mathbb{R}^{m},\ l\in[L]$ are the hidden-layer and output-layer weights of the two-layer perceptron, respectively. $m$ is the number of neurons in the hidden layer. $\text{Relu}: \mathbb{R}^{m}\rightarrow \mathbb{R}^m$ where $\text{Relu}(\bfx)=\max\{\bfx,0\}$. $\text{softmax}: \mathbb{R}^{L}\rightarrow \mathbb{R}^L$ where $\text{softmax}(\bfx)=(e^{x_1}, e^{x_2}, \cdots, e^{x_L})/\sum_{i=1}^L e^{x_i}$. Let $\psi=(\bfA,\bfW_O,\bfW_V,\bfW_K,\bfW_Q)$ denote the set of  parameters to train.  The training problem minimizes the   empirical risk $f_N(\psi)$, 
\begin{equation}
    \min_{\psi}: f_N(\psi)=\frac{1}{N}\sum_{n=1}^N \ell(\bfX^n,y^n;\psi), \label{eqn: min_train}
\end{equation}
where $\ell(\bfX^n,y^n;\psi)$ is the Hinge loss function, i.e.,
\begin{equation}
    \ell(\bfX^n,y^n;\psi)=\max\{1-y^n\cdot F(\bfX^n),0\}.
\end{equation}
The generalization performance of a learned model $\psi$ 
is evaluated by the population risk $f(\psi)$, where 
\begin{equation}
    f(\psi)=f(\bfA, \bfW_O,\bfW_V,\bfW_K, \bfW_Q)=\mathbb{E}_{(\bfX,y)\sim\mathcal{D}}[\max\{1-y\cdot F(\bfX),0\}].
\end{equation}

The training problem (\ref{eqn: min_train}) is solved via a mini-batch stochastic gradient descent (SGD),  
as summarized in Algorithm \ref{alg: sgd}. At iteration $t$, $t=0,1,2,\cdots,T-1$, the gradient is computed using a mini-batch $\mathcal{B}_t$ with $|\mathcal{B}_t|=B$. The step size is $\eta$. 

Similar to  \citep{DBKW20, TCDM21, JHYZ21}, $\bfW_{V}^{(0)}$, $\bfW_{Q}^{(0)}$, and $\bfW_{K}^{(0)}$ come from an initial model. Every entry of $\bfW_{O}$ is generated from  $\mathcal{N}(0,\xi^2)$. 
Every entry of $\bfa_{l}^{(0)}$ is sampled from $\{+\frac{1}{\sqrt{m}},-\frac{1}{\sqrt{m}}\}$ with equal probability. $\bfA$ does not update during the training\footnote{It is common to fix the output layer weights as the random initialization  in  the theoretical analysis of neural networks, including NTK \citep{ALL19, ADHL19}, model recovery \citep{ZSJB17}, and feature learning \citep{KWLS21, AL22} type of approaches. 
The optimization problem here of $\bfW_Q$, $\bfW_K$, $\bfW_V$, and $\bfW_O$ with non-linear activations is still highly non-convex and challenging.}. 

\section{Theoretical Results}

\subsection{Main Theoretical insights}\label{subsec: insights}

Before formally introducing our data model and main theory, we first summarize the major insights. We consider a data model where tokens are noisy versions of \textit{label-relevant} patterns that determine the data label and \textit{label-irrelevant} patterns that do not affect the label. $\alpha_*$ is the fraction of label-relevant tokens. $\sigma$ represents the initial model error, and $\tau$ characterizes the token noise level. 

\textbf{(P1). A Convergence and sample complexity analysis of SGD to achieve zero generalization error.} We prove SGD with a proper initialization  converges to a model with zero generalization error. 
The required number of iterations is proportional to  $1/\alpha_*$ and $1/(\Theta(1)-\sigma-\tau)$.  Our sample complexity bound is linear in $\alpha_*^{-2}$ and $(\Theta(1)-\sigma-\tau)^{-2}$. Therefore, 
the learning performance is improved, in the sense of a faster convergence and fewer training samples to achieve a desirable generalization, with a larger fraction of label-relevant patterns, a better initial model, and less token noise. 

\textbf{(P2). A theoretical characterization of increased sparsity of the self-attention module during  training.}  We prove that  the attention weights, which are softmax values of each token in the self-attention module, become increasingly   sparse during the training, with non-zero weights concentrated at  label-relevant tokens. This formally justifies the general intuition that  the attention layer makes the neural network focus on the most important part of data. 

\textbf{(P3). A theoretical guideline of designing token sparsification methods to reduce sample complexity.}   
Our sample complexity bound indicates that the required number of samples to achieve zero generalization can be reduced if a token sparsification method removes some label-irrelevant tokens  (reducing $\alpha_*$), or tokens with large noise 
(reducing $\sigma$), or both. This insight provides a guideline to design proper token sparsification methods. 

\textbf{(P4). A new theoretical framework to analyze the nonconvex interactions in shallow ViTs}. This paper develops a new framework to analyze ViTs based on a more general data model than existing works like \citep{BG21, KWLS21, WL21}. Compared with the nonconvex interactions in three-layer feedforward neural networks, analyzing ViTs has technical challenges that the softmax activation is highly non-linear, and the gradient computation on token correlations is complicated. 
We develop new tools to handle this problem by exploiting structures in the data and proving that SGD iterations increase the magnitude of label-relevant tokens only rather than label-irrelevant tokens. 
This theoretical framework  is of independent interest and can be potentially applied to analyze different variants of Transformers and attention mechanisms. 

\subsection{Data Model}\label{subsec: data_model}

There are  $M\ (2<M<m_a,m_b)$ distinct patterns $\{\bfmu_1,\ \bfmu_2,\cdots, \bfmu_M\}$ in $\mathbb{R}^d$,  where $\bfmu_1,\bfmu_2$ are \textit{discriminative patterns} that determine the binary labels, 
and the remaining $M-2$ patterns $\bfmu_3,\ \bfmu_4,\cdots,\bfmu_M$ are \textit{non-discriminative patterns} that do not affect the labels.   Let $\kappa=\min_{1\leq i\neq j\leq M}\|\bfmu_i-\bfmu_j\|>0$ denote the minimum distance between patterns.  Each token $\bfx_l^n$ of $\bfX^n$ is a noisy version of one of the patterns, i.e., 
\begin{equation}\min_{j\in[M]}\|\bfx_l^n-\bfmu_j\|\leq  \tau, 
\label{eqn: mu_j}\end{equation}
and the noise level $\tau < \kappa/4$. We take $\kappa-4\tau$ as $\Theta(1)$ for the simplicity of presentation.

The label $y^n$  is determined by the tokens that correspond to discriminative patterns through a majority vote. If  the number of tokens that are noisy versions of $\bfmu_1$ is larger than the number of tokens that correspond to $\bfmu_2$ in $\bfX^n$, then $y^n=1$. In this case that the label $y^n=1$, the tokens that are noisy $\bfmu_1$ are refereed to as \textit{label-relevant} tokens, and the tokens that are noisy $\bfmu_2$ are referred to as \textit{confusion} tokens. Similarly, if there are more tokens that are noisy $\bfmu_2$ than those that are noisy $\bfmu_1$, the former are label-relevant tokens,  the latter are confusion tokens, 
and $y^n=-1$. All other tokens that are not label-relevant are called label-irrelevant tokens.  

Let $\alpha_*$ and $\alpha_\#$ as the average fraction of the label-relevant and the confusion tokens over the  distribution $\mathcal{D}$, respectively.  
We consider a balanced dataset. Let
$\mathcal{D}_+=\{(\bfX^n,y^n)|y^n=+1, n\in[N]\}$ and $\mathcal{D}_-=\{(\bfX^n,y^n)|y^n=-1, n\in[N]\}$ denote  the sets of positive and negative labels, respectively. Then 
$\Big||\mathcal{D}^+|-|\mathcal{D}^-|\Big|=O(\sqrt{N})$.

Our model is motivated by and generalized from those used in the state-of-art analysis of neural networks on structured data  \citep{LL18, BG21, KWLS21}. All the existing models require that only one discriminative pattern  exists in each sample, i.e., either $\bfmu_1$ or $\bfmu_2$, but not both, while our model allows both patterns to appear in the same sample.

\subsection{Formal Theoretical Results}\label{subsec: theory}

Before presenting our main theory below, we first characterize the behavior of the initial model through Assumption \ref{asmp: VQK_initialization}. Some important notations are summarized in Table \ref{tab: notation}. 
\begin{table}[h!]
  \begin{center}
        \caption{Some important notations}    \label{tab: notation}
    \begin{tabular}{l|c|l|c} 
 \hline
 \small $\sigma$ & \scriptsize{Initialization error for value vectors} & \small$\delta$ & \scriptsize{Initialization error for query and key vectors}\\
 \hline
 \small$\kappa$ & \scriptsize{Minimum of $\|\bfmu_i-\bfmu_j\|$ for any $i,j\in[M],\ i\neq j$.} &
 \small$\tau$ & \scriptsize{Token noise level}\\
 \hline
  \small$M$ & \scriptsize{Total number of patterns}
  & \small$m$ & \scriptsize{The number of neurons in $\bfW_O$}\\
 \hline
  $\alpha_*$ & \scriptsize{Average fraction of label-relevant tokens}
  & $\alpha_\#$ &   \scriptsize{Average fraction of confusion tokens} \\
 \hline
    \end{tabular}

  \end{center}
\end{table}

\begin{assumption}\label{asmp: VQK_initialization}
Assume $\max (\|\bfW_V^{(0)}\|, \|\bfW_K^{(0)}\|, \|\bfW_Q^{(0)}\|) \leq 1$ without loss of generality. 
There exist three (not necessarily different) sets  of orthonormal bases
  $\mathcal{P}=\{\bfp_1,\bfp_2,\cdots,\bfp_M\}$,  $\mathcal{Q}=\{\bfq_1,\bfq_2,\cdots,\bfq_M\}$, and $\mathcal{R}=\{\bfr_1,\bfr_2,\cdots,\bfr_M\}$, 
  where $\bfp_l\in\mathbb{R}^{m_a}$, $\bfq_l,\ \bfr_l\in\mathbb{R}^{m_b}$,  $ \forall l\in[M]$,    $\bfq_1=\bfr_1$,  and $\bfq_2=\bfr_2$\footnote{The condition   $\bfq_1=\bfr_1$ and $\bfq_2=\bfr_2$ is to eliminate the trivial case that the initial attention value is very small. This condition can be relaxed but we keep this form to simplify the representation.} such that 
\begin{equation}\label{eqn:initial}
      \|\bfW_V^{(0)}\bfmu_j-\bfp_j\|\leq \sigma,  \quad
      \|\bfW_K^{(0)}\bfmu_j-\bfq_j\|\leq \delta, \quad  \textrm{and } \|\bfW_Q^{(0)}\bfmu_j-\bfr_j\|\leq \delta.
 \end{equation}
hold for some $\sigma=O(1/M)$ and $\delta<1/2$. 
\end{assumption}

Assumption \ref{asmp: VQK_initialization}  characterizes  the distance of query, key, and value vectors of patterns $\{\bfmu_j\}_{j=1}^M$ to orthonormal vectors.  The requirement on $\delta$ is minor because $\delta$ can be in the same order as $\|\bfmu_j\|$.  

\begin{thm}[Generalization of ViT]\label{thm: main_theory}
Suppose 
Assumption \ref{asmp: VQK_initialization} holds; $\tau \leq \min(\sigma, \delta)$; 
 a sufficiently large model with 
\begin{equation}
m\gtrsim M^2\log N,
\end{equation}
the average fraction of label-relevant patterns satisfies
\begin{equation}
    \alpha_*\geq \frac{\alpha_\#}{ e^{-(\delta+\tau)}(1-(\sigma+\tau))},\label{eqn: alpha*}
\end{equation}
, and the mini-batch size and the number of sampled tokens of each data $\bfX^n,\ n\in[N]$ satisfy
\begin{equation}
    B\geq \Omega(1),\ \ \ \ |\mathcal{S}^n|\geq \Omega(1)
\end{equation}
Then, after $T$ number of iterations   such that
\begin{equation}
    T=\Theta(\eta^{-3/5}\alpha_*^{-1})\label{eqn: iterations}
\end{equation}, as long as  the number of training samples  $N$ satisfies
\begin{equation}
    N\geq \Omega(\frac{1}{(\alpha_*-c'(1-\zeta)-c'' (\sigma+\tau))^{2}})\label{eqn: sample_complexity}
\end{equation}
for some constant $c',c''>0$, and $\zeta\gtrsim 1-\eta^{10}$, 
 with a probability of at least $0.99$, the returned model achieves zero generalization error as
\begin{equation}
    f(\bfA^{(0)}, \bfW_O^{(T)},\bfW_V^{(T)},\bfW_K^{(T)},\bfW_Q^{(T)})=0
\end{equation}
\end{thm}

Theorem \ref{thm: main_theory} characterizes under what condition of the data the neural network with self-attention in (\ref{eqn: attention}) trained with Algorithm \ref{alg: sgd} can achieve zero generalization error. To show that the self-attention layer can improve the generalization performance by reducing the required sample complexity to achieve zero generalization error, we also quantify the sample complexity when there is no self-attention layer in the following proposition. 

\begin{prpst}[Generalization without self-attention]\label{prpst: ViT_CNN}
Suppose assumptions in Theorem \ref{thm: main_theory} hold. 
    When there is no self-attention layer, i.e., 
    $\bfW_K$ and $\bfW_Q$ are not updated during the training,  if $N$ satisfies
    \begin{equation}\label{eqn:no_attention}
    N\geq \Omega(\frac{1}{(\alpha_*(\alpha_*-\sigma-\tau))^2})
    \end{equation}
    then after $T$ iterations with $T$ in (\ref{eqn: iterations}), the returned model achieves zero generalization error as 
    \begin{equation}
    f(\bfA^{(0)}, \bfW_O^{(T)},\bfW_V^{(T)},\bfW_K^{(0)},\bfW_Q^{(0)})=0
\end{equation}
    \end{prpst}

\begin{rmk} (Advantage of the self-attention layer) Because $m\gg m_a, m_b, d$, the number of trainable parameter   remains almost the same with or without updating the attention layer. Combining Theorem \ref{thm: main_theory} and Proposition \ref{prpst: ViT_CNN}, we can see that with the additional self-attention layer, the sample complexity\footnote{The sample complexity bounds in (\ref{eqn: sample_complexity}) and (\ref{eqn:no_attention}) are sufficient but not necessary. Thus, rigorously speaking, one can not compare two cases based on sufficient conditions only. In our analysis, however, these two bounds are derived with   exactly the same technique with the only difference in handling the self-attention layer. Therefore, we believe it is fair to compare these two bounds to show the advantage of ViT.} is reduced by a factor $1/\alpha_*^2$ with an approximately equal number of network parameters. 
\end{rmk}

\begin{rmk}\label{rmk: sparsification} (Generalization improvement by token sparsification). (\ref{eqn: sample_complexity})  and (\ref{eqn: iterations}) show that the sample complexity $N$ and the required number of iterations $T$ scale with $1/\alpha_*^2$  and $1/\alpha_*$, respectively. Then, increasing $\alpha_*$, the fraction of label-relevant tokens, can reduce the sample complexity and speed up the convergence. Similarly,  $N$ and $T$ scale with  $1/(\Theta(1)-\tau)^2$  and $1/(\Theta(1)-\tau)$. Then decreasing $\tau$, the noise in the tokens, can also improve the generalization. Note that a properly designed token sparsification method can both increase $\alpha_*$ by removing label-irrelevant tokens and decrease $\tau$ by removing noisy tokens, thus improving the generalization performance. 
\end{rmk}

\begin{rmk}\label{rmk: initial_model} (Impact of the initial model) The initial model  $\bfW_V^{(0)}$, $\bfW_K^{(0)}$, $\bfW_Q^{(0)}$ affects the learning performance through $\sigma$ and $\delta$, both of which decrease as the initial model is improved. Then from (\ref{eqn: sample_complexity}) and (\ref{eqn: iterations}), the sample complexity reduces and the convergence speeds up for a better initial model.  
\end{rmk}

Proposition \ref{lemma: concentration} shows that the attention weights are increasingly concentrated on label-relevant tokens during the training.  Proposition  \ref{lemma: concentration} is a critical component in proving Theorem \ref{thm: main_theory} and is of independent interest.  

\begin{prpst}\label{lemma: concentration}
The attention weights for each token become increasingly concentrated on those correlated with tokens of the label-relevant pattern during the training, i.e.,
\begin{equation}\label{eqn:attention}
    \sum_{i\in\mathcal{S}_*^n}\text{softmax}({\bfX^n}^\top{\bfW_K^{(t)}}^\top\bfW_Q^{(t)}\bfx_l^n)_{i}=\sum_{i\in\mathcal{S}_*^n}\frac{\exp({\bfx_i^n}^\top{\bfW_K^{(t)}}^\top\bfW_Q^{(t)}\bfx_l^n)}{\sum_{r\in\mathcal{S}^n}\exp({\bfx_r^n}^\top{\bfW_K^{(t)}}^\top\bfW_Q^{(t)}\bfx_l^n)}\rightarrow 1-\eta^C
\end{equation}
at a sublinear rate of $O(1/t)$ when $t$ is large for a large $C>0$ and all $l\in\mathcal{S}^n$ and $n\in[N]$.
\end{prpst}

Proposition \ref{lemma: concentration} indicates that only label-relevant tokens are highlighted by the learned attention of ViTs, 
while other tokens have less weight. This provides a theoretical justification of 
magnitude-based token sparsification methods. $\text{softmax}(\cdot)_i$ in (\ref{eqn:attention}) denotes the $i$-th entry of $\text{softmax}(\cdot)$.


\textbf{Proof idea sketch}: The main proof idea is to show that 
the SGD updates scale up value, query, and key vectors of discriminative patterns, while  keeping the magnitude of the projections of non-discriminative patterns and the initial model error almost unchanged. To be more specific, by Lemma \ref{lemma: initial_WU}, \ref{lemma: update_WU}, we can identify two groups of neurons in the hidden layer  $\bfW_{O}$, 
where one group only learns the positive pattern, and the other group only learns the negative pattern. Claim \ref{clm: W_O} of Lemma \ref{lm: training} states that   during the SGD updates, the neuron weights in these two groups evolve in the direction of 
projected discriminative patterns,  $\bfp_1$ and $\bfp_2$, respectively. Meanwhile,  Claim \ref{clm: W_QK} of Lemma \ref{lm: training} indicates that  $\bfW_K$ and $\bfW_Q$ update in the direction of increasing the magnitude of the query and key vectors of label-relevant tokens from $1$ to $\Theta(\log T)$, such that the attention weights correlated with label-relevant tokens gradually become dominant. 
Moreover, by Claim \ref{clm: W_V} of Lemma \ref{lm: training}, the update of $\bfW_V$ increases the magnitude of the value vectors of label-relevant tokens, by adding partial neuron weights of $\bfW_O$ that are aligned with the value vectors 
to these  vectors. 
Due to the above properties during the training, one can simplify the training process to show that 
the output of neural network (\ref{eqn: attention}) changes linearly in the iteration number $t$. From the above analysis, we can develop the sample complexity and the required number of iterations for the zero generalization guarantee.

\textbf{Technical novelty}: Our proof technique is inspired by the feature learning technique in analyzing fully connect networks and convolution neural networks \citep{SWL21, BG21}. Our paper makes new technical contributions from the following aspects. First, we provide a new framework of studying the nonconvex interactions of multiple weight matrices in a shallow ViT 
while other feature learning works \citep{SWL21, BG21, KWLS21, AL22, WL21, ZWCL23} only study one trainable 
weight matrix in the hidden layer of a two-layer network. Second, we 
analyze the updates of the self-attention module with the softmax function during the training, while other papers either ignore this issue  without exploring convergence analysis \citep{EGKZ22} or oversimplify the analysis by applying the neural-tangent-kernel (NTK) method 
that considers impractical over-parameterization and updates the weights only around initialization.  \citep{HBSN20, Y20, ALL19, ADHL19}. 
Third, we consider a more general 
data model, where discriminative patterns of multiple classes can exist in the same data sample, 
but 
the data models in \citep{BG21, KWLS21} require one discriminative pattern only in each sample. 
\section{Numerical experiments}\label{sec: experiments}
\subsection{Experiments on Synthetic datasets}
We first verify the theoretical bounds in  Theorem \ref{thm: main_theory} on synthetic data. We set the dimension of data and attention embeddings to be $d=m_a=m_b=10$. Let $c_0=0.01$. Let the total number of patterns $M=5$, and $\{\bfmu_1,\bfmu_2, \cdots,\bfmu_M\}$ be a set of orthonormal bases.  To satisfy Assumption \ref{asmp: VQK_initialization}, we generate every token that is a noisy version of $\bfmu_i$ from a Gaussian distribution $\mathcal{N}(\bfmu_i, c_0^2\cdot I)$ with the mean $\bfmu_i$ and covariance $c_0^2\bfI$, where $\bfI\in\mathbb{R}^d$ is the identity matrix.
  $\bfW_Q^{(0)}=\bfW_Q^{(0)}=\delta^2\bfI/c_0^2$, $\bfW_V^{(0)}=\sigma^2\bfU/c_0^2$, and each entry of $\bfW_{O}^{(0)}$ follows $\mathcal{N}(0,\xi^2)$, where $\bfU$ is an $m_a\times m_a$ orthonormal matrix, and $\xi=0.01$.  The number of neurons $m$ of $\bfW_O$ is  $1000$. 
  We set the ratio of different patterns the same among all the data for simplicity. 

\begin{figure}[htbp]
  \centering
\begin{minipage}{0.65\textwidth}
  
    \subfigure[]{
        \begin{minipage}{0.32\textwidth}
        \centering
        \includegraphics[width=1.05\textwidth,height=1.05in]{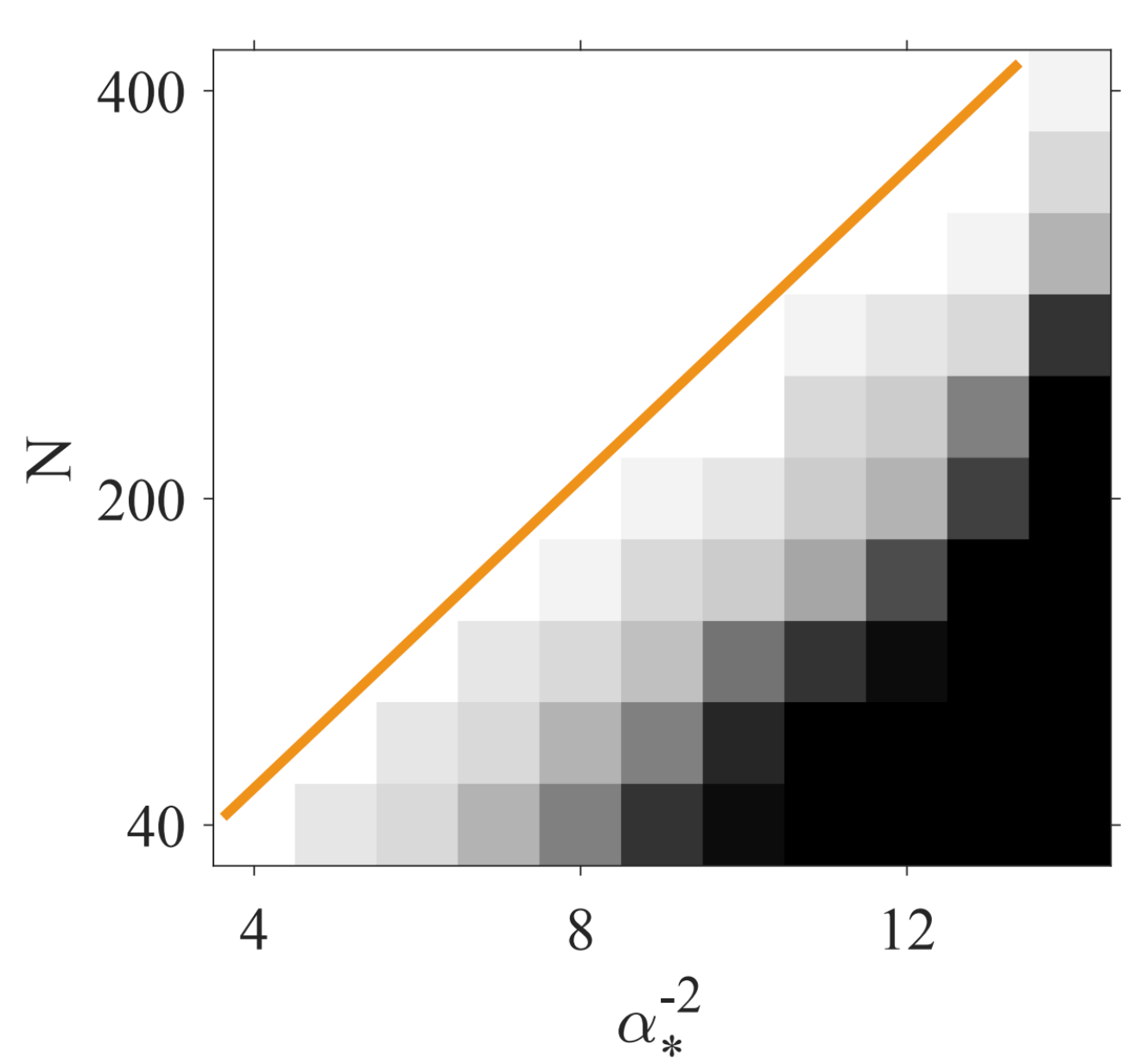}
        \end{minipage}
    }
    ~
        \subfigure[]{
        \begin{minipage}{0.32\textwidth}
        \centering
        \includegraphics[width=1.2\textwidth,height=1.05in]{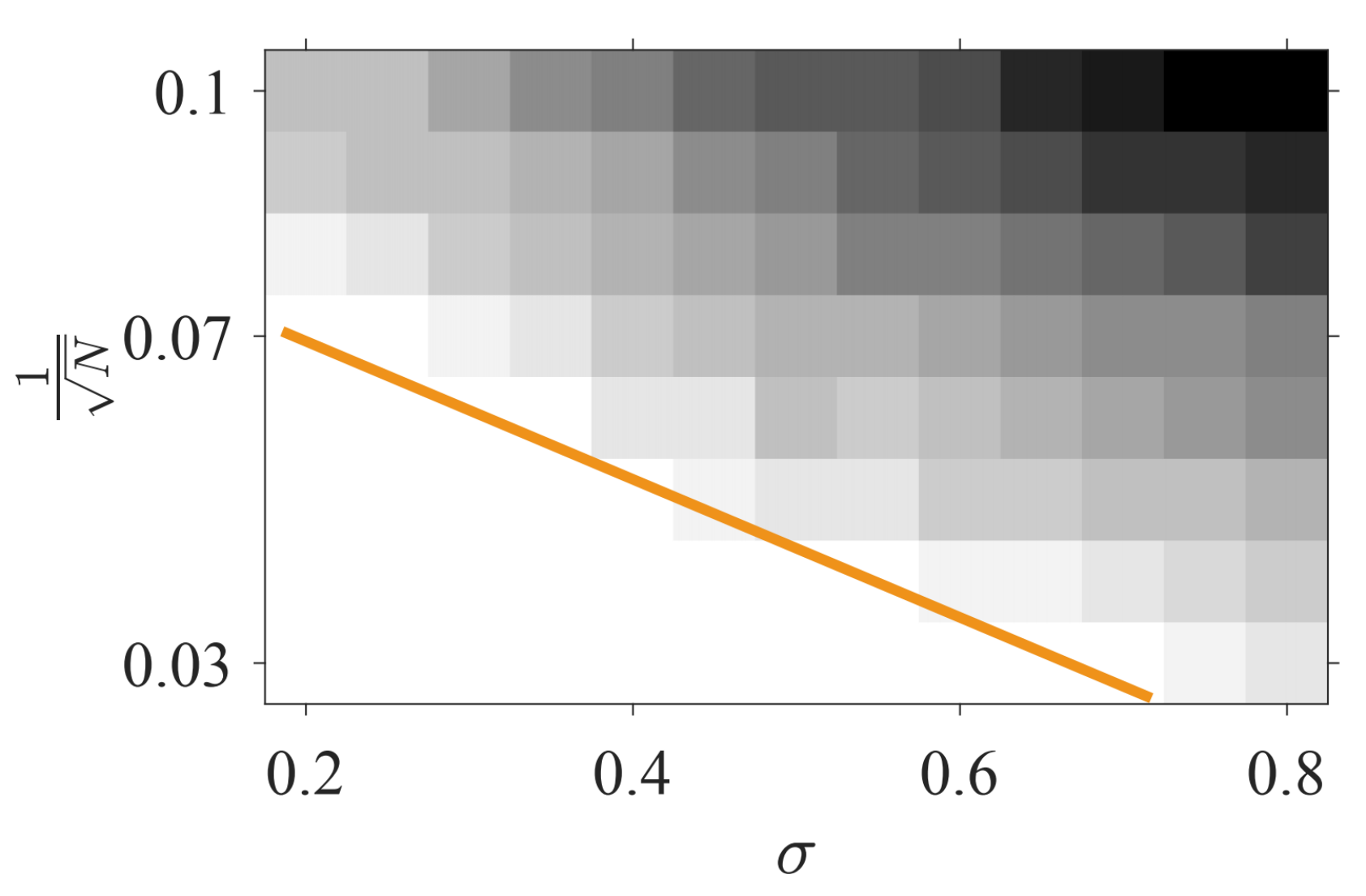}
        \end{minipage}
    }
     \caption{The impact of $\alpha_*$ and $\sigma$ on sample complexity.}     \label{figure: sample_complexity}
    \end{minipage}
    ~
        \begin{minipage}{0.26\textwidth}
        \centering
        \includegraphics[width=0.9\textwidth,height=1.05in]{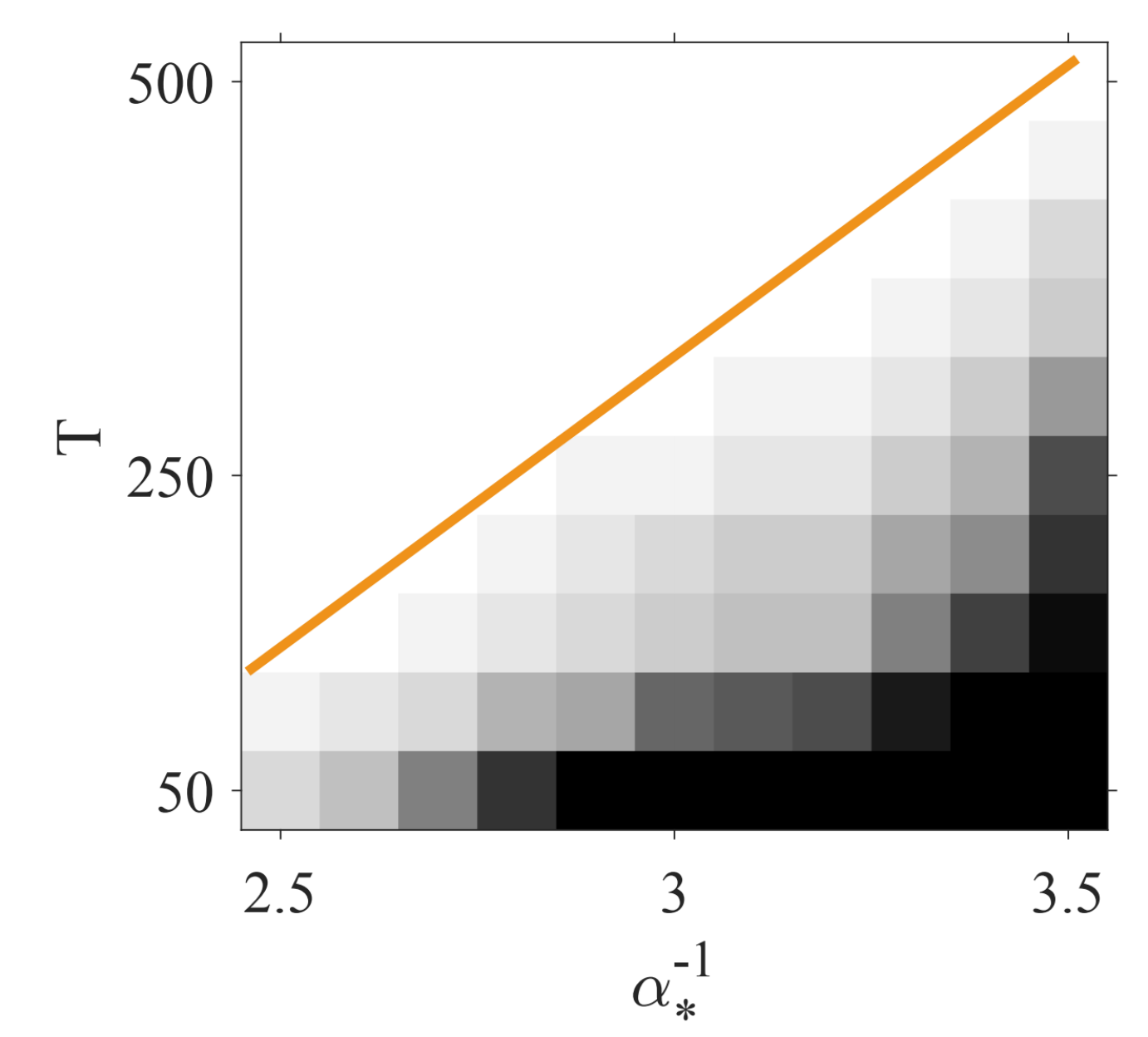}
        \caption{The number of iterations against $\alpha_*^{-1}$.}\label{figure: convergence}
        \end{minipage}
 \end{figure}

\textbf{Sample complexity and convergence rate}:
We first study the impact of the fraction of the label-relevant patterns $\alpha_*$  on the sample complexity. Let the number of tokens after sparsification be $|\mathcal{S}^n|=100$, the initialization error $\sigma=0.1$,  
and $\delta=0.2$. 
The fraction of non-discriminative patterns is fixed to be $0.5$. We implement $20$ independent experiments with the same $\alpha_*$ and $N$ and record the Hinge loss values of the testing data. An experiment is successful if the testing loss is smaller than $10^{-3}$. Figure \ref{figure: sample_complexity} (a) shows the success rate of these experiments. A black block means  that all the trials fail. A white block means  that they   all succeed. The sample complexity is indeed almost linear in $\alpha_*^{-2}$, as predicted in \ref{eqn: sample_complexity}.  We next explore the impact on 
$\sigma$. Set $\alpha_*=0.3$ and $\alpha_\#=0.2$. The number of tokens after sparsification is fixed at $50$ for all the data.   Figure \ref{figure: sample_complexity} (b) shows that $1/\sqrt{N}$ is linear in $\Theta(1)-\sigma$, matching our theoretical prediction 
in (\ref{eqn: sample_complexity}). The result on the noise level $\tau$ is similar to Figure \ref{figure: sample_complexity} (b), and we skip it here.  In Figure \ref{figure: convergence}, we verify the number of iterations $T$ against $\alpha_*^{-1}$ in (\ref{eqn: iterations}) where we set $\sigma=0.1$ and $\delta=0.4$. 

\begin{figure}[htbp]
    \centering
    \begin{minipage}{0.3\textwidth}
        \centering
        \includegraphics[width=0.8\textwidth, height=1.05in]{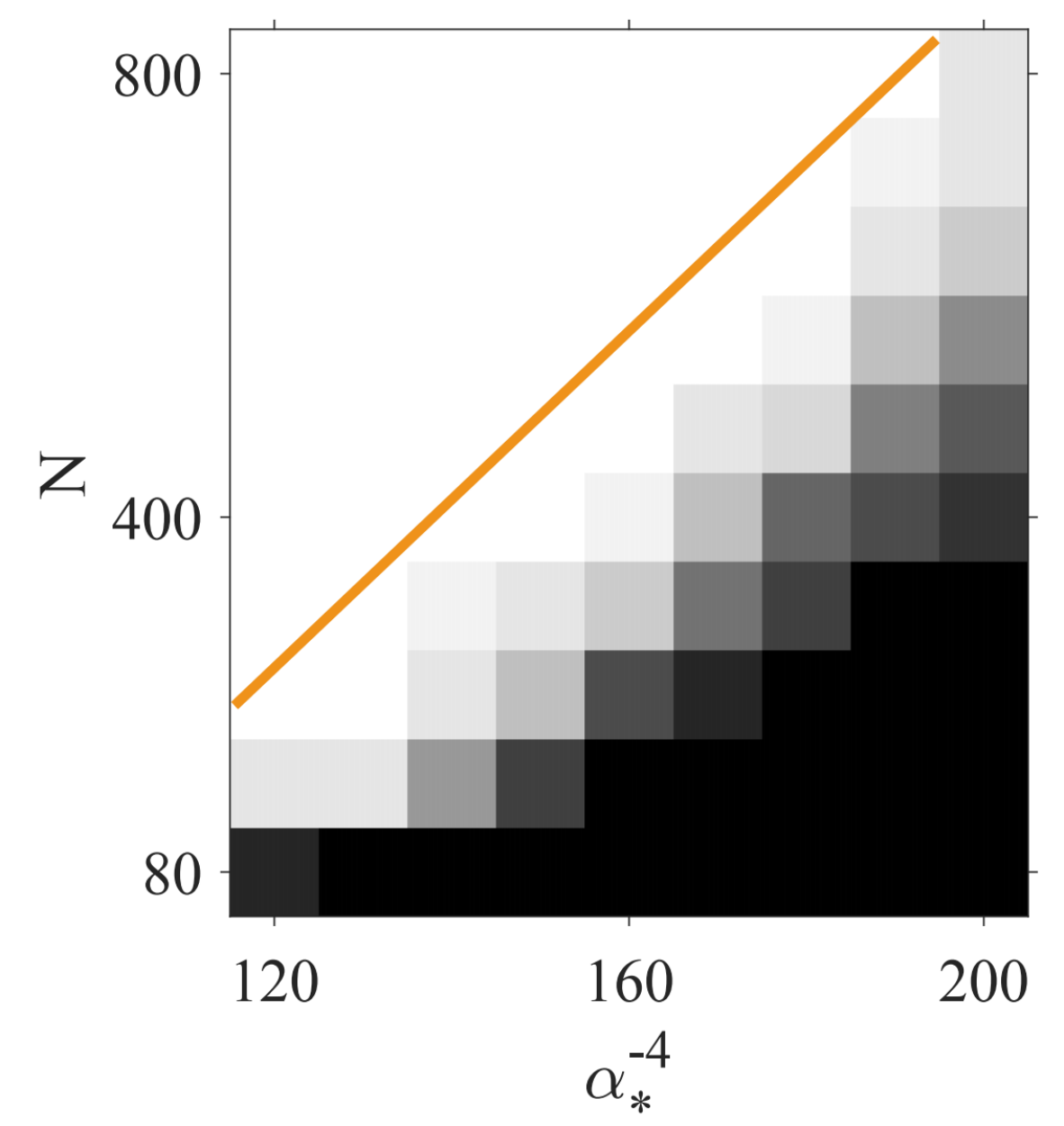}
        \caption{Comparison of ViT and CNN} \label{figure: CNN}
        \end{minipage}
    ~
        \begin{minipage}{0.3\textwidth}
        \centering
        \includegraphics[width=0.8\textwidth]{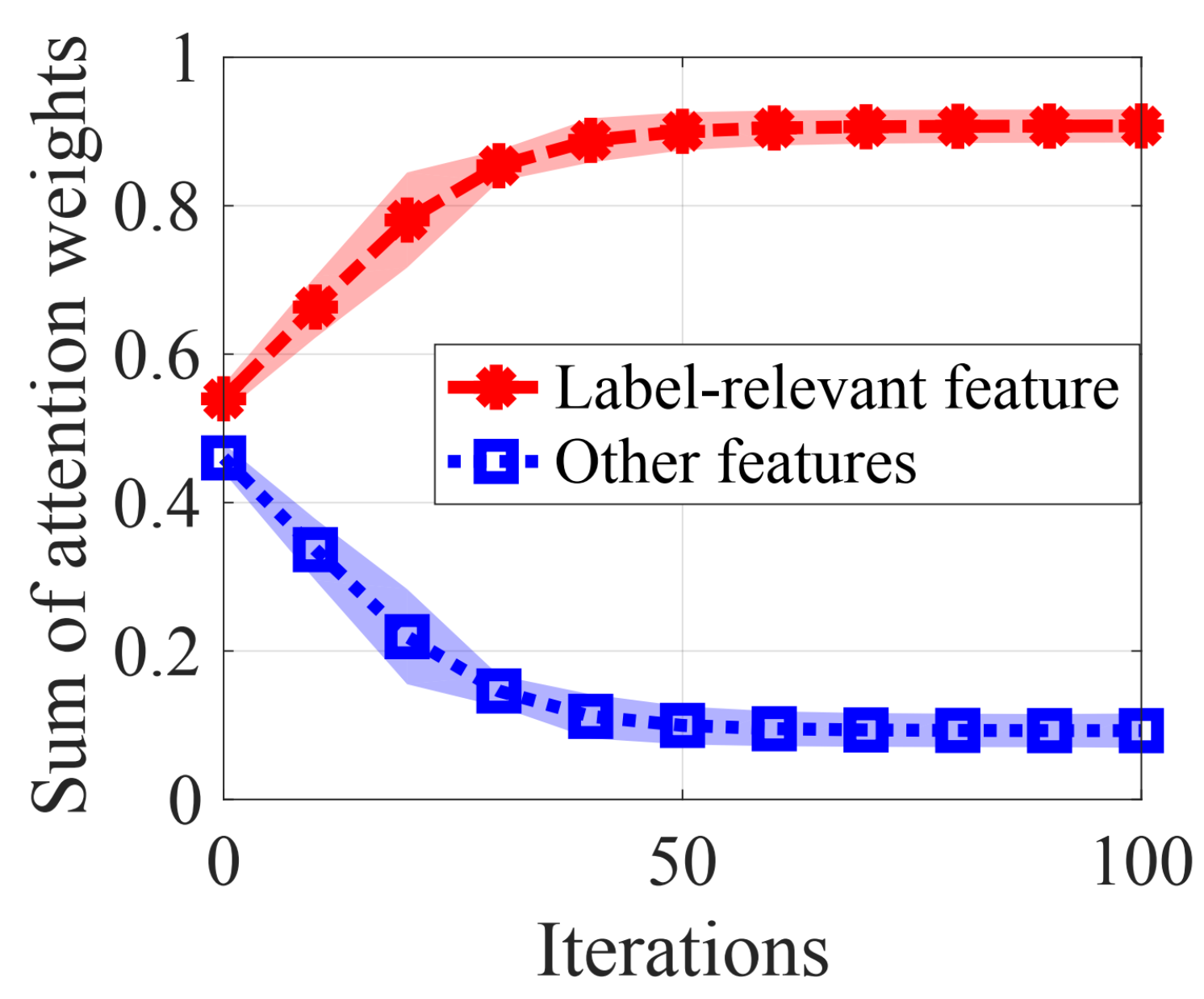}
        \caption{Concentration of attention weights} \label{figure: softmax}
        \end{minipage}
    ~
        \begin{minipage}{0.3\textwidth}
        \centering
        \includegraphics[width=0.92\textwidth,height=1.03in]{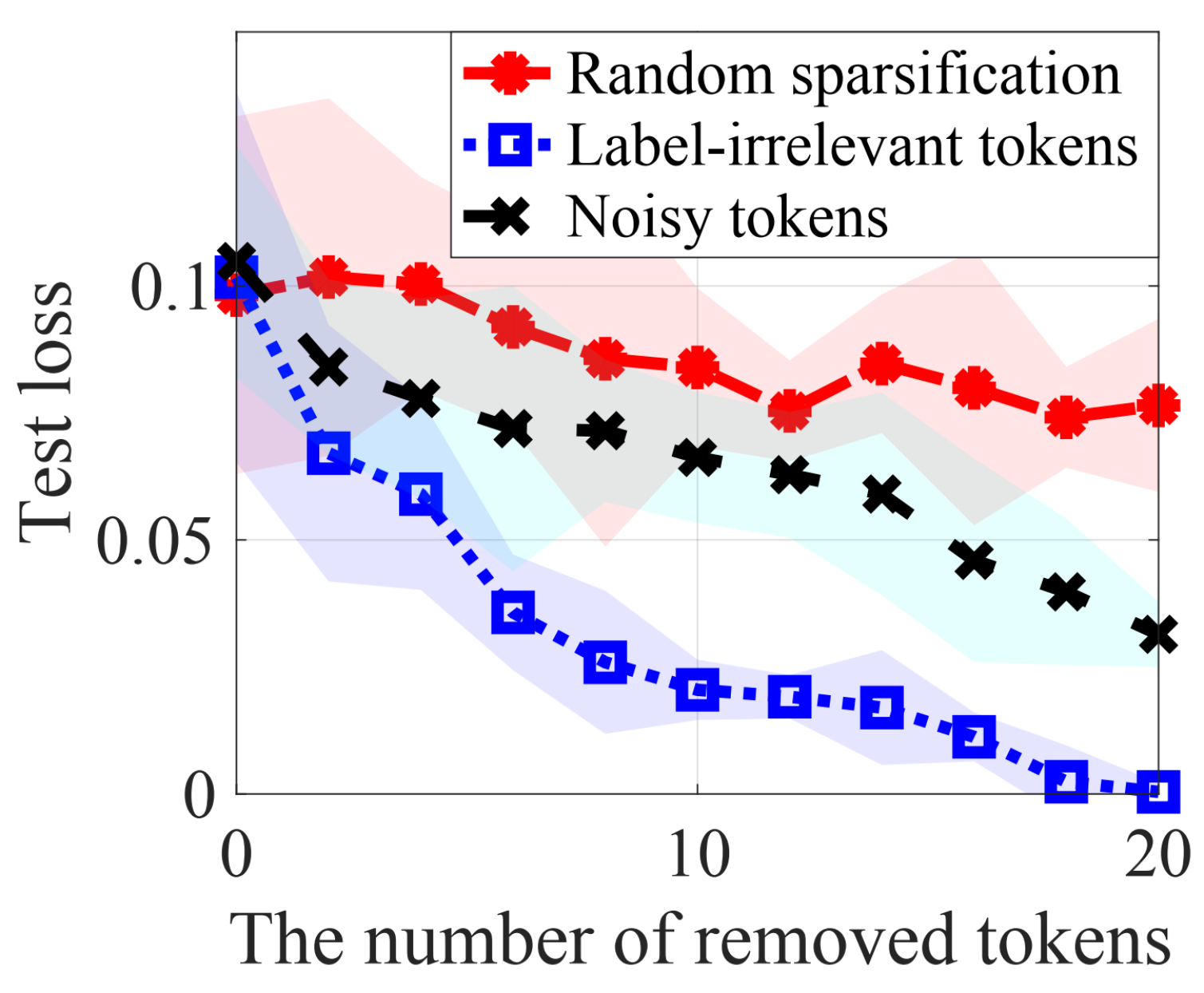}
        \caption{Impact of token sparsification on testing loss} \label{figure: sparse}
        \end{minipage}
   \end{figure}

\textbf{Advantage of self-attention}: To verify Proposition \ref{prpst: ViT_CNN}, we compare the performance on ViT in \ref{eqn: attention} and on the same network with $\bfW_K$ and $\bfW_Q$ fixed during the training, i.e., a three-layer CNN. Compared with ViT, the number of trainable parameters in CNN is reduced by only $1\%$. Figure \ref{figure: CNN} shows the sample complexity of CNN  is almost linear in $\alpha_*^{-4}$ as predicted in (\ref{eqn:no_attention}). Compared with Figure \ref{figure: convergence} (a), the sample complexity significantly increases for small $\alpha_*$, indicating a much worse generalization of CNN. 

\textbf{Attention map}:  
We then evaluate the evolution of the attention map during the training. Let  $|\mathcal{S}^n|=50$ for all $n\in[N]$. The number of training samples is $N=200$. 
$\sigma=0.1$, $\delta=0.2$, 
$\alpha_*=0.5$, $\alpha_\#=0.05$. 
In Figure \ref{figure: softmax}, the red   line with asterisks shows that the sum of attention weights on  label-relevant tokens, i.e., the left side of (\ref{eqn:attention}) averaged over all $l$, indeed increases to be close to 1 when the number of iterations increases. Correspondingly,  the sum of attention weights on  other tokens decreases to be close to $0$, as shown in the blue line with squares. 
This verifies Lemma \ref{lemma: concentration} on a sparse attention map.

\textbf{Token sparsification}: We verify the improvement by token sparsification in Figure \ref{figure: sparse}. The experiment is duplicated $20$ times. The number of training samples $N=80$. Let $|\mathcal{S}^n|=50$ for all $n\in[N]$. Set $\sigma=0.1$, $\delta=0.5$, 
$\alpha_*=0.6$, $\alpha_\#=0.05$. 
If we apply random sampling over all tokens, the performance cannot be improved as shown in the red curve because $\alpha_*$ and $\sigma$ do not change. If we remove either  label-irrelevant tokens or  tokens with significant noise,  
the testing loss decreases, as indicated in the blue and black curves. This justifies our insight \textbf{P3} on token sparsification. 

\subsection{Experiments on  Image Classification  Datasets}


\textbf{Dataset}: To characterize the effect of label-relevant and label-irrelevant tokens on generalization, following the setup of image integration in \citep{KWLS21},  we adopt an image from  CIFAR-10 dataset \citep{KNH10} as the   label-relevant image pattern and integrate it with a noisy background image from the IMAGENET Plants synset \citep{KWLS21, DDSL09}, which plays the role of label-irrelevant feature. Specifically, we randomly cut out a region with size  $26\times 26$ in the IMAGENET image and replace it with a resized CIFAR-10 image. 

\textbf{Architecture}: Experiments are implemented on a deep ViT model. Following \citep{DBKW20}, the network architecture contains $5$ blocks, where we have a $4$-head self-attention layer and a one-layer perceptron with skip connections and Layer-Normalization in each block.

\begin{figure}[htbp]
\centering
           \begin{minipage}{1\textwidth}
    \subfigure[]{
        \begin{minipage}{0.27\textwidth}
        \centering
        \includegraphics[width=0.85\textwidth, height=1in]{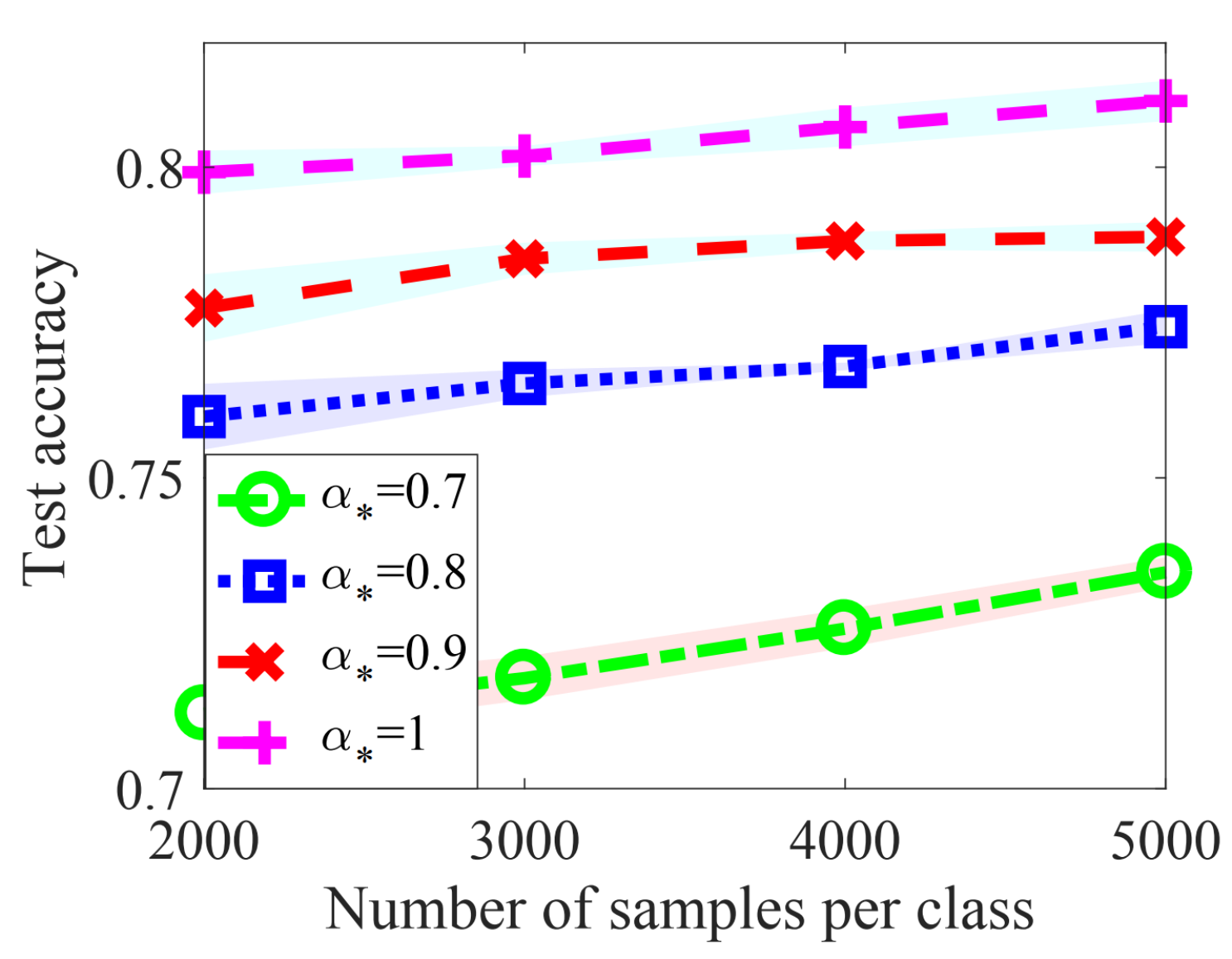}
        \end{minipage}
    }
     ~
        \subfigure[]{
        \begin{minipage}{0.27\textwidth}
        \centering
        \includegraphics[width=0.85\textwidth, height=1in]{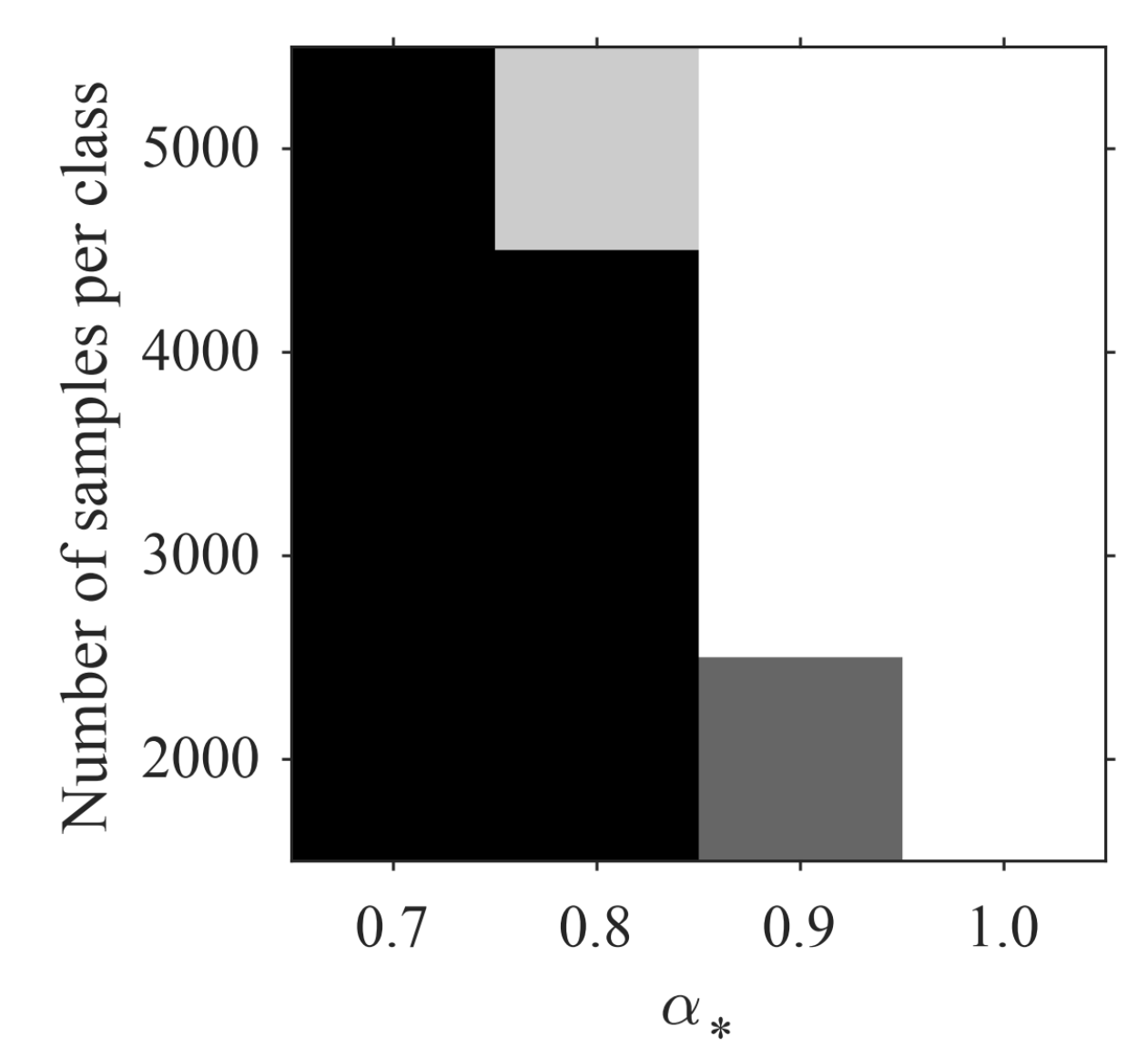}
        \end{minipage}
    }
     ~
    \subfigure[]{
        \begin{minipage}{0.27\textwidth}
        \centering
        \includegraphics[width=0.95\textwidth,height=1in]{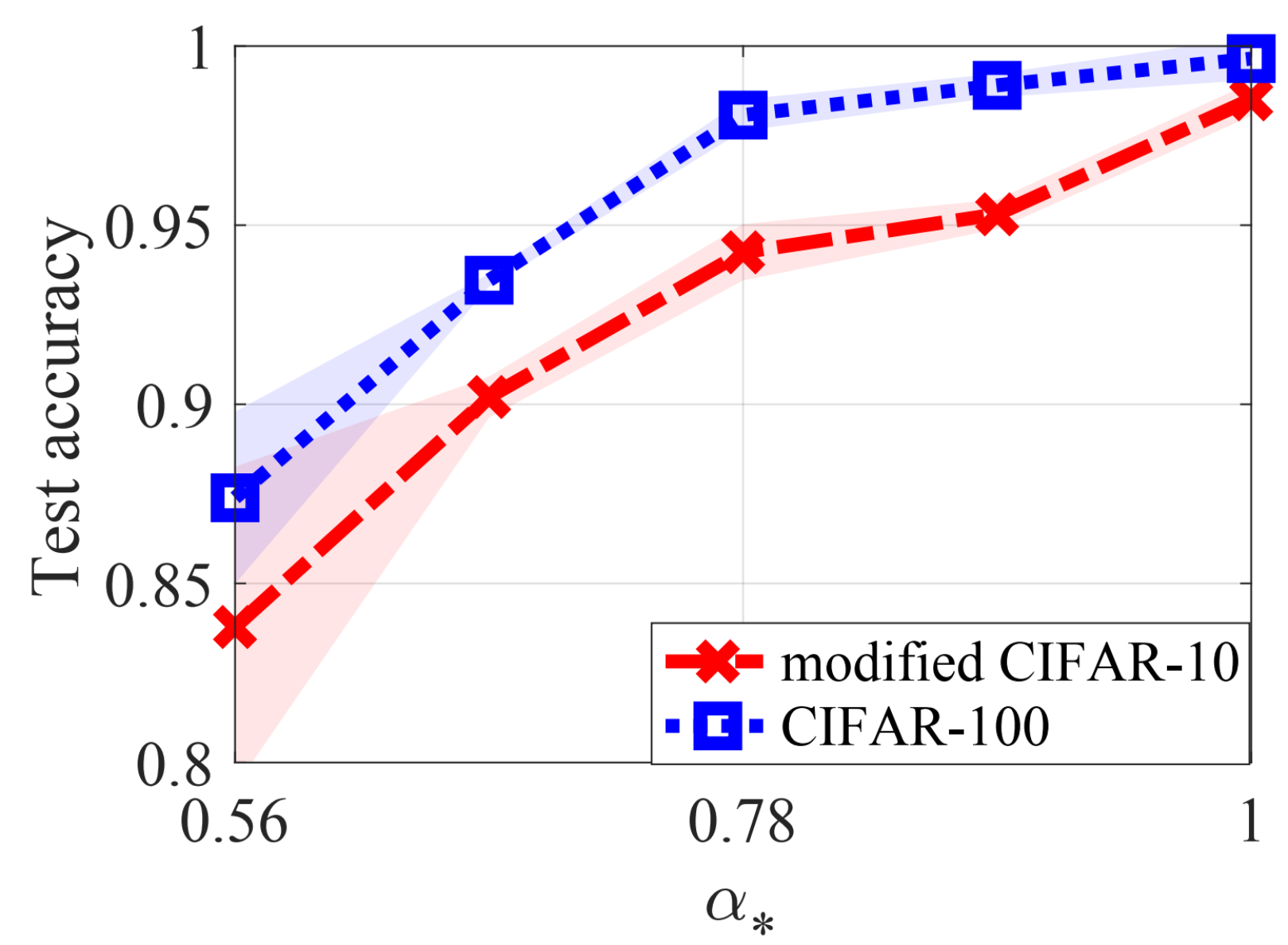}
        \end{minipage}
        }
    \caption{(a) Test accuracy when $N$ and $\alpha_*$ change. (b) Relationship of sample complexity against $\alpha^*$. (c) Test accuracy when token sparsification removes spurious correlations.}    \label{figure: cifar}
   \end{minipage}
\end{figure}

We first evaluate the impact on generalization of token sparsification that removes label-irrelevant patterns to increase $\alpha_*$. We consider a ten-classification problem where in both the training and testing datasets, the images used for integration are randomly selected from CIFAR-10 and IMAGENET. The number of samples for training and testing is $50K$ and $10K$, respectively. A pre-trained model from CIFAR-100  \citep{KNH10} is used as the initial model with the output layer randomly initialized. Without token sparsification, the fraction of class-relevant tokens is $\alpha_*\approx 0.66$.  $\alpha_*=1$ implies all background tokens are removed.  Figure \ref{figure: cifar} (a) indicates that a larger $\alpha_*$ by removing more label-irrelevant tokens leads to higher test accuracy. Moreover, the test performance  improves with more training samples. These are consistent with our sample complexity analysis in (\ref{eqn: sample_complexity}). Figure \ref{figure: cifar} (b) presents the required sample complexity to learn a model with desirable test accuracy. We run $10$ independent experiments for each pair of $\alpha_*$ and $N$, and the experiment is considered a success if the learned model achieves a test accuracy of at least $77.5\%$.


We then evaluate the impact of token sparsification on removing spurious correlations  \citep{SRKL20}, as well as the impact of the initial model. We consider a binary classification problem that differentiates ``bird'' and ``airplane'' images. To introduce spurious correlations in the training data,  $90\%$ of bird images in the training data are integrated  into the IMAGENET plant background, while only  $10\%$ of airplane images have the plant background. The remaining training data are integrated into a clean background by zero padding. Therefore, the label ``bird'' is spuriously correlated with the class-irrelevant plant background. 
The testing data contain 50\% birds and 50\% airplanes, and each class has 50\% plant background and 50\% clean background. 
The numbers of training and testing samples are $10K$ and $2K$, respectively. We initialize the ViT using two pre-trained models. The first one is pre-trained with CIFAR-100, which contains images of 100 classes not including birds and airplanes. The other initial model is trained with a modified CIFAR-10 with $500$ images per class for a total of eight classes,  excluding birds and airplanes. The pre-trained model on CIFAR-100 is a  better initial model because it is trained on a more diverse dataset with more samples. 

In
Figure \ref{figure: cifar} (c), the token sparsification method removes the tokens of the added background, and the corresponding $\alpha_*$ increases. Note that removing background in the training dataset also reduces the spurious correlations between birds and plants. Figure \ref{figure: cifar} (c)  shows that from both initial models, the testing accuracy increases when more background tokens are removed. Moreover, a better initial model leads to a better testing performance.  This is consistent with Remarks \ref{rmk: sparsification} and \ref{rmk: initial_model}. 

\section{Conclusion}\label{sec: conclusion}
This paper provides a novel theoretical 
generalization analysis of shallow  ViTs. 
Focusing on a data model with label-relevant and label-irrelevant tokens, this paper explicitly quantifies the sample complexity as a function of the fraction of label-relevant tokens and the token noise projected by the  initial model. It proves that the learned attention map becomes increasingly sparse during the training, 
where the attention weights are concentrated on those of  label-relevant tokens. Our theoretical results   also offer a guideline on designing 
proper token sparsification methods to improve the test performance. 

This paper considers a simplified but representative Transformer architecture to theoretically examine the role of self-attention layer as the first step. One future direction is to analyze more practical architectures such as those with skip connection, local attention layers, and Transformers in other areas.  We see no ethical or immediate negative societal consequence of our work.


\subsubsection*{Acknowledgments}
This work was supported by AFOSR FA9550-20-1-0122, ARO W911NF-21-1-0255, NSF 1932196 and the Rensselaer-IBM AI Research Collaboration (http://airc.rpi.edu), part of the IBM AI Horizons Network (http://ibm.biz/AIHorizons). We thank Dr. Shuai Zhang at Rensselaer Polytechnic Institute for useful discussions. We thank all
anonymous reviewers for their constructive comments.


\newpage
\appendix

\textbf{The appendix contains 6 sections. We first provides a brief discussion about comparisons between our works and other two related works in Section \ref{sec: compare}. In Section \ref{sec: appendix_experiment}, we add additional experiments for the verification of our theory. In Section \ref{sec: preliminaries}, we introduce some definitions and assumptions in accordance with the main paper for the ease of the proof in the following. Section \ref{sec: proof_thm} first states a core lemma for the proof, based on which we provide the proof of Theorem \ref{thm: main_theory} and Proposition \ref{prpst: ViT_CNN} and \ref{lemma: concentration}. Section \ref{sec: lemma 2} gives the proof of Lemma \ref{lm: training} with three subsections to prove its three main claims. Section \ref{sec: lemmas} shows key lemmas and the proof of lemmas for this paper. We finally discuss the extension of our analysis in Section \ref{sec: extension}, including extension to multi-classification cases, general data model cases, multi-head attention cases, and cases with skip connections in Section \ref{subsec: ext_multi}, \ref{subsec: general_data}, \ref{subsec: multi-head}, and \ref{subsec: skip_connection}, respectively. }

\section{Comparison with two related works}\label{sec: compare}
\subsection{Comparison with (ALLEN-ZHU \& Li, 2023)}
\citet{AL23} studies ensemble learning and knowledge distillation. Its main proof idea is that given large amounts of multi-view data, each single model learns one feature, and then ensemble learning integrates all learned features and, thus, improves over single models. Knowledge distillation applies softmax logits to make use of information learned from the ensemble model. It is analyzed in a similar approach to studying single models. The single models considered in \citep{AL23} is a two-layer Relu network.

In contrast, in this paper, we consider a two-layer Relu network with an additional self-attention layer. The network architecture and training algorithm for the self-attention layer is completely different from those for the softmax logit in the knowledge distillation function. In our proof, we analyze the impact of the gradient of $\bfW_Q$, $\bfW_K$, and $\bfW_V$ on different patterns (Claims 2 and 3 of Lemma 2), showing that the training process helps to enlarge the magnitude of label-relevant features. We also show that neurons in $\bfW_O$ mainly learn from discriminative patterns (Claim 1 of Lemma 2). Such a learning process is affected by the error in the initial model and the noise in tokens. Please see details in “Proof idea sketch” and “Technical novelty” in Section 3.3 on Page 7 of the paper. This technique we develop plays a critical role in our analysis of self-attention layers. This technique is novel and did not appear in any existing works.

\subsection{Comparison with (JELASSI ET AL., 2022)}
\citet{JSL22} is a concurrent work which theoretically studies Vision Transformers. The major difference between \citep{JSL22} and our work is that we consider different data models and network architectures. In \citep{JSL22}, the data model requires spatial association between tokens. The attention map is replaced with position encoding, and the training process of the attention map is simplified to train a linear layer. Our setup models the data mainly based on the category of patterns. We keep the classical structure and training process of self-attention, where $\bfW_Q$, $\bfW_K$, and $\bfW_V$ are trained separately. The required number of samples and iterations are derived as functions of the fraction of label-relevant patterns. In addition, the non-linear activation function they consider is polynomial activation, instead of \text{Relu} or \text{Gelu} as in practice. Based on these conditions, they are able to study a different and mroe general labelling function.

\section{More experiments}\label{sec: appendix_experiment}

\begin{figure}[htbp]
    \centering
    \begin{minipage}{0.3\textwidth}
        \centering
        \includegraphics[width=0.88\textwidth]{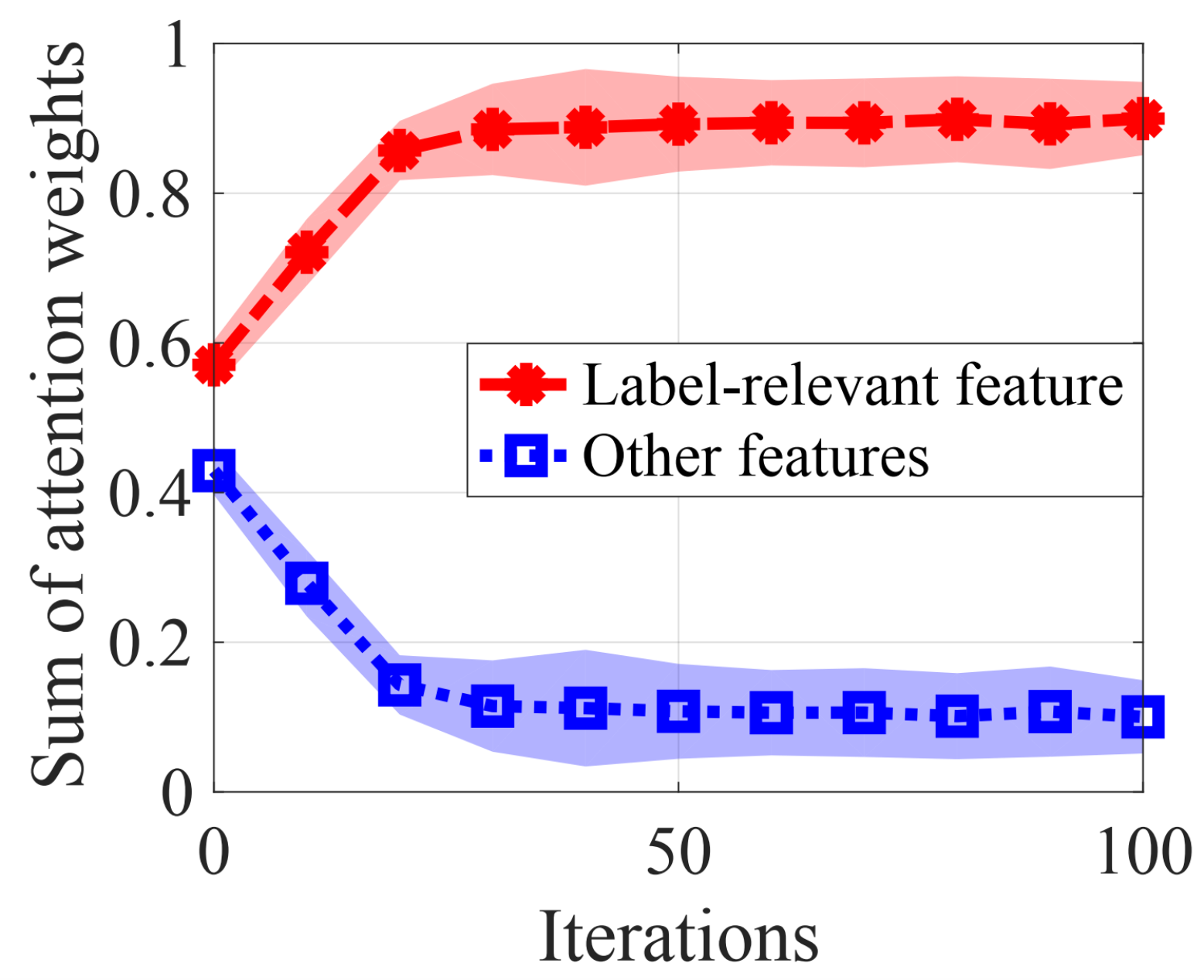}
        \end{minipage}
    ~
        \begin{minipage}{0.3\textwidth}
        \centering
        \includegraphics[width=0.92\textwidth]{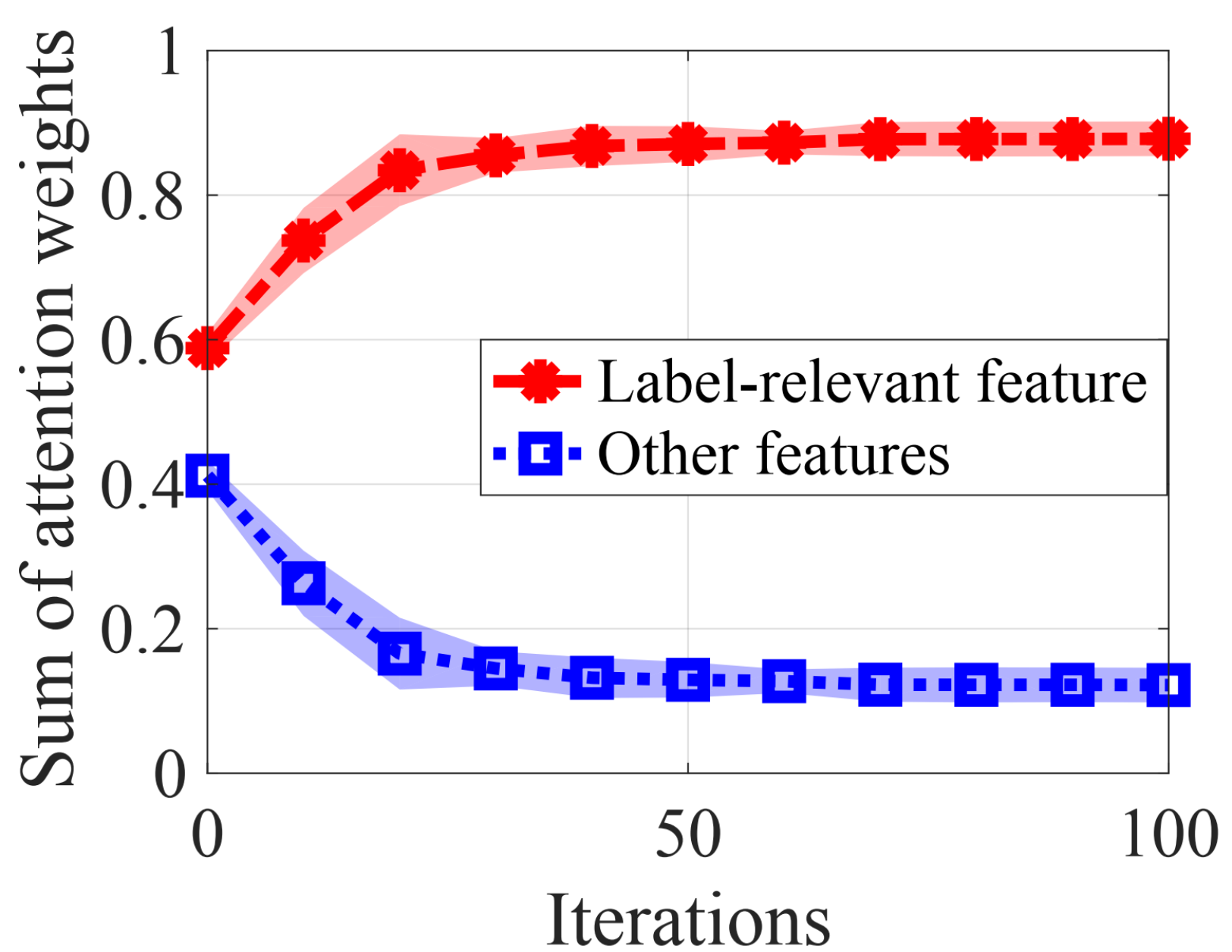}
        \end{minipage}
    ~
        \begin{minipage}{0.3\textwidth}
        \centering
        \includegraphics[width=0.92\textwidth]{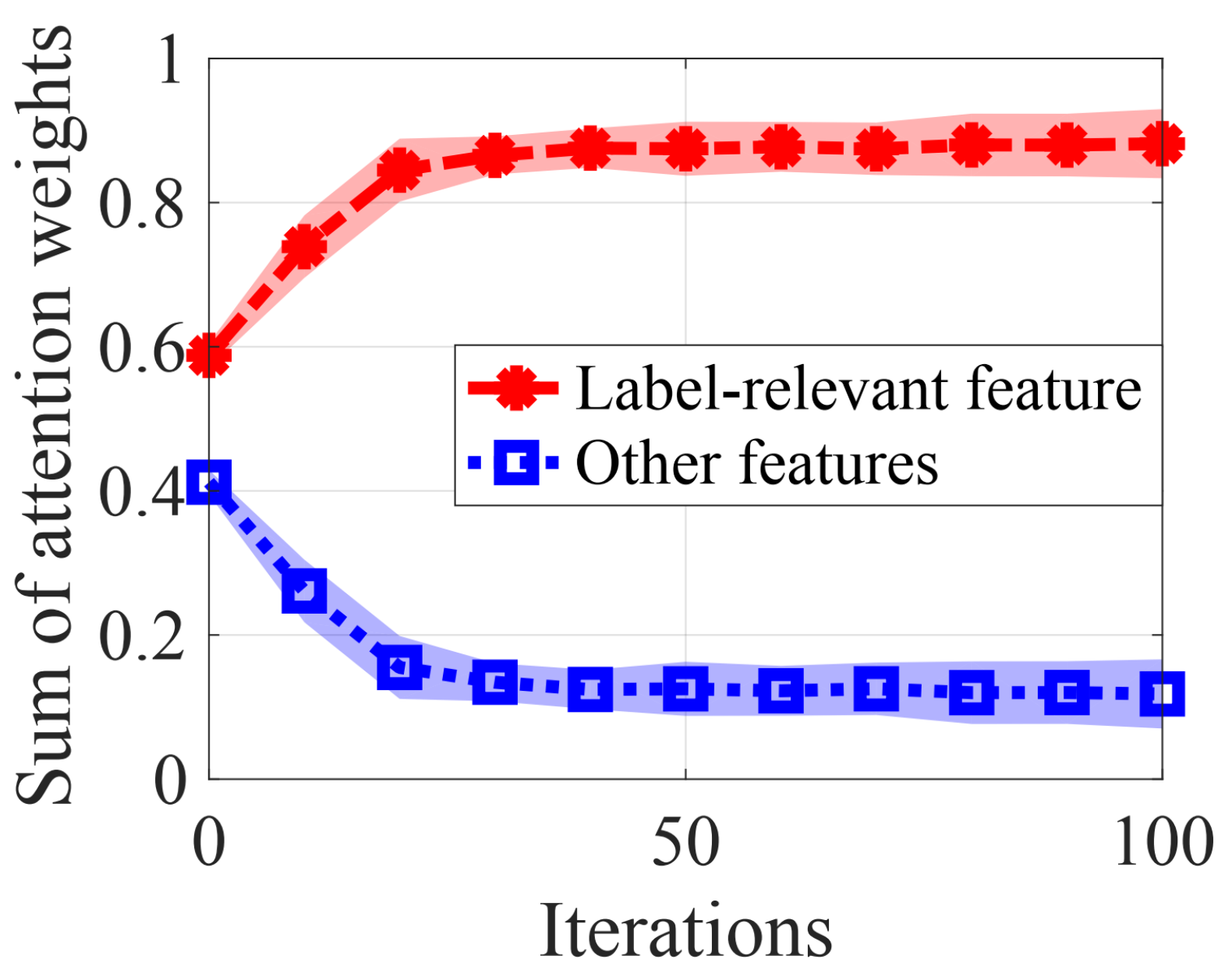}
        \end{minipage}
   \caption{Concentration of attention weights when (a) $M=10$ (b) $M=15$ (c) $M=20$.}\label{fig: attention_map_M}
   \end{figure}

\begin{figure}[htbp]
    \centering
    \begin{minipage}{0.3\textwidth}
        \centering
        \includegraphics[width=0.88\textwidth]{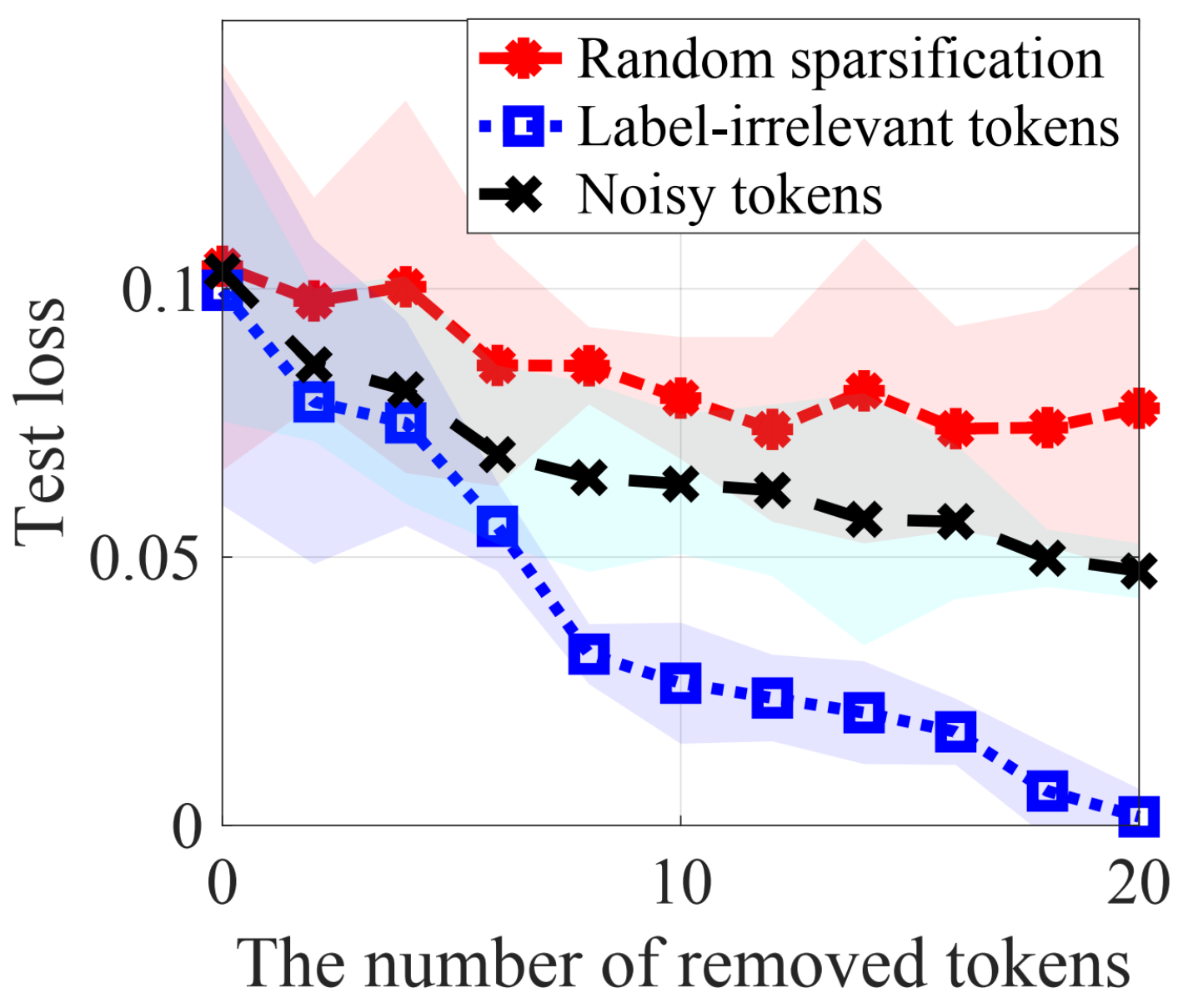}
        \end{minipage}
    ~
        \begin{minipage}{0.3\textwidth}
        \centering
        \includegraphics[width=0.9\textwidth]{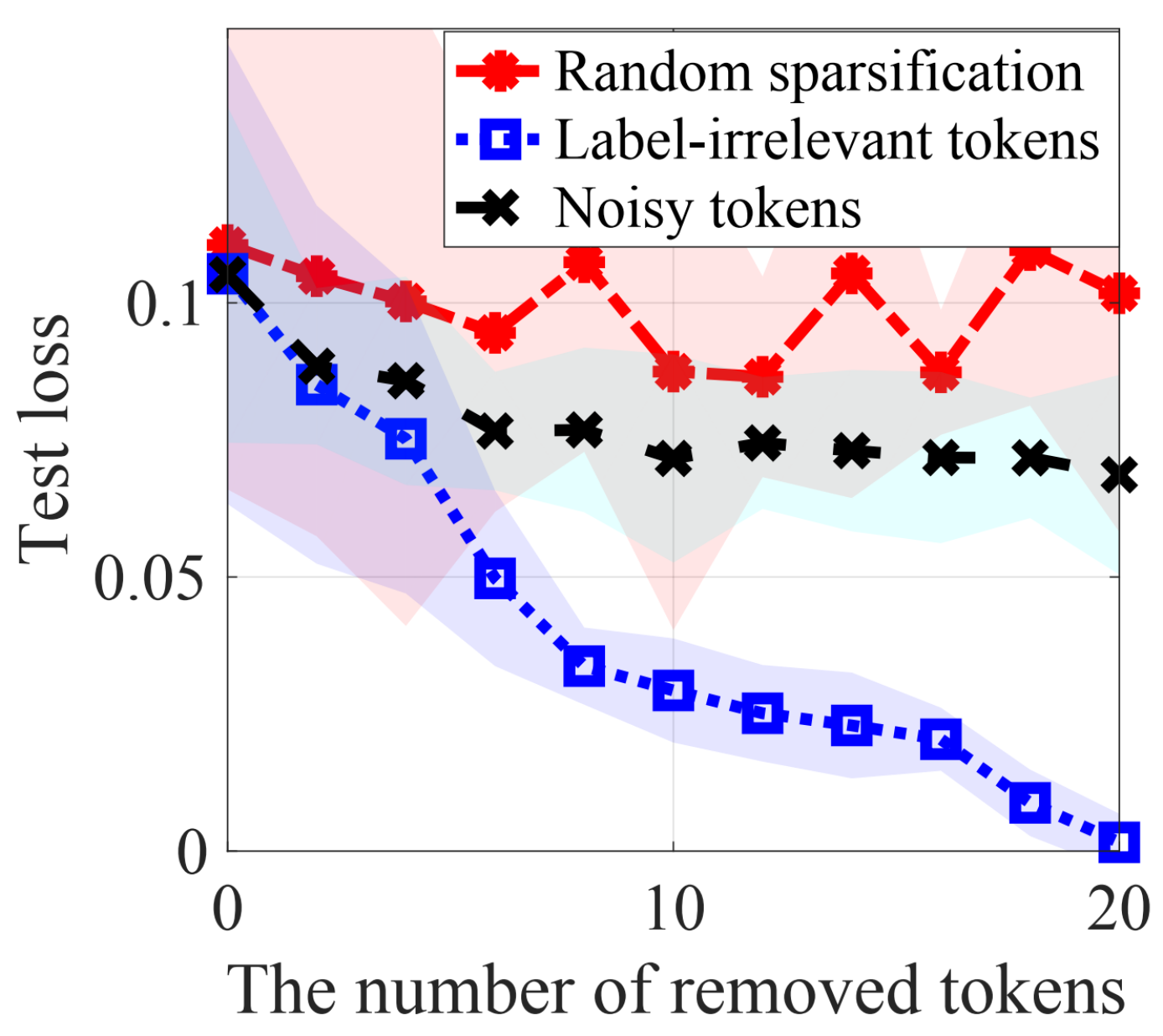}
        \end{minipage}
    ~
        \begin{minipage}{0.3\textwidth}
        \centering
        \includegraphics[width=0.92\textwidth]{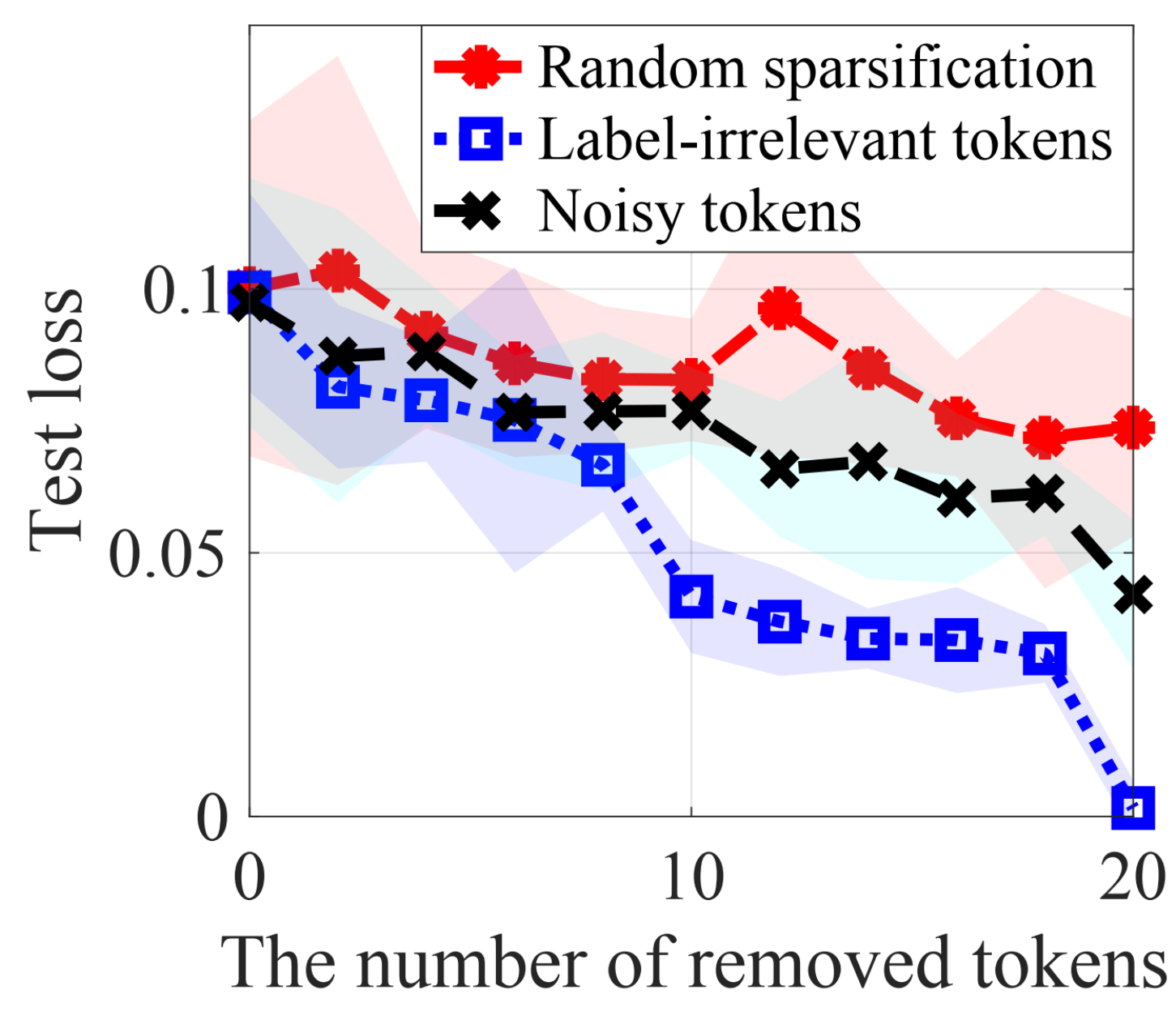}
        \end{minipage}
   \caption{Impact of token sparsification on testing data when (a) $M=10$ (b) $M=15$ (c) $M=20$.}\label{fig: sparsification_M}
   \end{figure}

Following a similar setup in Section \ref{sec: experiments}, we add more experiments.\\
For experiments on synthetic data, we set the dimension of data and attention embeddings to be $d=m_a=m_b=20$. We vary the number of patterns $M$ to be $10$, $15$, and $20$. Data generation and the network architecture follow the setup in Section \ref{sec: experiments}. One can observe the same trend in Figure \ref{fig: attention_map_M} and \ref{fig: sparsification_M} as in Figure \ref{figure: softmax} and \ref{figure: sparse}, respectively, indicating that our conclusion that the attention map becomes sparse during the training, and that pruning label-irrelevant tokens or noisy tokens improves the performance, both hold for different choices of $M$.

\section{Preliminaries}\label{sec: preliminaries}

We first formally restate the neural network with different notations of loss functions, and the Algorithm \ref{alg: sgd} of the training steps after token sparsification. The notations used in the Appendix is summarized in Table \ref{tbl:notations}.

\begin{table}[h!]

  \begin{center}
        \caption{Summary of notations}
        \label{tbl:notations}
\begin{tabularx}{\textwidth}{lX} 
 \hline
 \small $F(\bfX^n)$, $\text{Loss}(\bfX^n, y^n)$ &  The network output for $\bfX^n$ and the loss function of a single data.\\ 
 \hline
 \small $\overline{\text{Loss}}_b$,  $\overline{\text{Loss}}$, $\text{Loss}$ &  The loss function of a mini-batch, the empirical loss, and the population loss, respectively.\\ 
 \hline
 \small $\bfp_j(t)$, $\bfq_j(t)$, $\bfr_j(t)$ & The features in value, key, and query vectors at the iteration $t$ for pattern $j$, respectively. We have $\bfp_j(0)=\bfp_j$, $\bfq_j(0)=\bfq_j$, and $\bfr_j(0)=\bfr_j$. \\
 \hline
 \small $\bfz_j^n(t)$, $\bfn_j^n(t)$, $\bfo_j^n(t)$ & The error terms in the value, key, and query vectors of the $j$-th token and $n$-th data compared to their features at iteration $t$. \\
 \hline
  \small $\mathcal{W}_{l,n}(0)$, $\mathcal{U}_{l,n}(0)$ & The set of lucky neurons for the token $l$ of data $n$.\\
 \hline
  \small $\phi_n(t)$, $\nu_n(t)$, $p_n(t)$, $\lambda$ &  The bounds of value of some attention weights at iteration $t$. $\lambda$ is the threshold between inner products of tokens from the same pattern and different patterns.\\
 \hline
 \small $\mathcal{S}_j^n$, $\mathcal{S}_*^n$, $\mathcal{S}_\#^n$ &  $\mathcal{S}_j^n$ is the set of sampled tokens of pattern $j$ for the $n$-th data. $\mathcal{S}_*^n$, $\mathcal{S}_\#^n$ are sets of sampled tokens of the label-relevant pattern and the confusion pattern for the $n$-th data, respectively.\\
 \hline
 \small $\alpha_*$, $\alpha_\#$, $\alpha_{nd}$ &  The mean of fraction of label-relevant tokens, confusion tokens, and non-discriminative tokens, respectively.\\
 \hline

\end{tabularx}
\end{center}

\end{table}

For the network\footnote{Note that in our proof in the Appendix, we often use the notation $\text{softmax}({\bfx_i^n}^\top{\bfW_K^{(t)}}^\top\bfW_Q^{(t)}\bfx_l^n)$, which is the same meaning as $\text{softmax}({\bfX^n}^\top{\bfW_K^{(t)}}^\top\bfW_Q^{(t)}\bfx_l^n)_{i}$.}
\begin{equation}
    F(\bfX^n)=\frac{1}{|\mathcal{S}^n|}\sum_{l\in\mathcal{S}^n}\bfa_{(l)}^\top\text{Relu}(\bfW_O\bfW_V\bfX^n\text{softmax}({\bfX^n}^\top\bfW_K^\top\bfW_Q\bfx_l^n))
\end{equation}
The loss function of a single data, a mini-batch, the empirical loss, and the population loss is defined in the following.
\begin{equation}
    \text{Loss}(\bfX^n,y^n)=\max\{1-y^n\cdot F(\bfX^n),0\}\label{loss_data}
\end{equation}
\begin{equation}
    \overline{\text{Loss}}_b=\frac{1}{B}\sum_{n\in\mathcal{B}_b} \text{Loss}(\bfX^n,y^n)\label{loss_batch}
\end{equation}
\begin{equation}
    \overline{\text{Loss}}=\frac{1}{N}\sum_{n=1}^N \text{Loss}(\bfX^n,y^n)\label{loss_empirical}
\end{equation}
\begin{equation}
    \text{Loss}=\mathbb{E}_{(\bfX,y)\sim\mathcal{D}}[\overline{\text{Loss}}]
\end{equation}

\noindent The formal algorithm is as follows. We assume that each entry of $\bfW_O^{(0)}$ is randomly initialized from $\mathcal{N}(0, \xi^2)$ where $\xi=\frac{1}{\sqrt{M}}$. Define that $a_{(l)_i}^{(0)},\ i\in[m],\ l\in[L]$ is uniformly initialized from $+\{\frac{1}{\sqrt{m}},-\frac{1}{\sqrt{m}}\}$ and fixed during the training. $\bfW_V$, $\bfW_K$, and $\bfW_Q$ are initialized from a good pretrained model.

\begin{algorithm}
\begin{algorithmic}[1]
\caption{Training with SGD}\label{alg: sgd}
\STATE{\textbf{Input: }} 
Training data $\{(\bfX^n,y^n)\}_{n=1}^N$, the step size $\eta$, the total number of iterations $T$, batch size $B$.
\STATE{\textbf{Initialization:}} Every entry of $\bfW_O^{(0)}$ from $\mathcal{N}(0,\xi^2)$,  and every entry of $\bfa_{(l)}^{(0)}$ from  $\text{Uniform}(\{+\frac{1}{\sqrt{m}},-\frac{1}{\sqrt{m}}\})$. 
$\bfW_V^{(0)}$, $\bfW_K^{(0)}$ and $\bfW_Q^{(0)}$ 
from a pre-trained model. 
\STATE{\textbf{Stochastic Gradient Descent:}} for $t=0,1,\cdots,T-1$ and $\bfW^{(t)}\in\{\bfW_O^{(t)},\bfW_V^{(t)},\bfW_K^{(t)}, \bfW_Q^{(t)}\}$
\vspace{-0.1in}
\begin{equation}\label{eqn:gradient}
\begin{aligned}
\bfW^{(t+1)}&=\bfW^{(t)}-\eta \cdot \frac{1}{B} \sum_{n\in\mathcal{B}_t} \nabla_{\bfW^{(t)}}\ell(\bfX^n,y^n;\bfW_O^{(t)},\bfW_V^{(t)},\bfW_K^{(t)}, \bfW_Q^{(t)})
\end{aligned}
\end{equation}
\vspace{-0.1in}
\STATE{\textbf{Output: }} $\bfW_O^{(T)}$, $\bfW_V^{(T)}$, $\bfW_K^{(T)}$,  $\bfW_Q^{(T)}$.
\end{algorithmic}
\end{algorithm}

Assumption \ref{asmp: VQK_initialization} can be interpreted as that we initialize $\bfW_V$, $\bfW_K$, and $\bfW_Q$ to be the matrices that can map tokens to orthogonal features with added error terms. 
\begin{assumption}\label{asmp: apdx_VQK_initialization}
Define $\bfP=(\bfp_1,\bfp_2,\cdots,\bfp_M)\in\mathbb{R}^{m_a\times M}$, $\bfQ=(\bfq_1,\bfq_2,\cdots,\bfq_M)\in\mathbb{R}^{m_b\times M}$ and $\bfR=(\bfr_1,\bfr_2,\cdots,\bfr_M)\in\mathbb{R}^{m_b\times M}$ as three feature matrices, where $\mathcal{P}=\{\bfp_1,\bfp_2,\cdots,\bfp_M\}$,  $\mathcal{Q}=\{\bfq_1,\bfq_2,\cdots,\bfq_M\}$ and $\mathcal{R}=\{\bfr_1,\bfr_2,\cdots,\bfr_M\}$ are three sets of orthonormal bases. Define the noise terms $\bfz_j^n(t)$, $\bfn_j^n(t)$ and ${\bfo_j^n}(t)$ with $\|\bfz_j^n(0)\|\leq \sigma+\tau$ and $\|\bfn_j^n(0)\|, \|{\bfo_j^n}(0)\|\leq \delta+\tau$ for $j\in[L]$. $\bfq_1=\bfr_1$, $\bfq_2=\bfr_2$. Suppose $\|\bfW_V^{(0)}\|,\|\ \bfW_K^{(0)}\|,\ \|\bfW_Q^{(0)}\|\leq 1$, $\sigma,\tau<O(1/M)$ and $\delta<1/2$. Then, for $\bfx_l^n\in\mathcal{S}_j^n$
\begin{enumerate}
    \item $\bfW_V^{(0)}\bfx_l^n=\bfp_j+\bfz_j^n(0)$.
    \item $\bfW_K^{(0)}\bfx_l^n=\bfq_j+\bfn_j^n(0)$.
    \item $\bfW_Q^{(0)}\bfx_l^n=\bfr_j+{\bfo_j^n}(0)$.
\end{enumerate}
\end{assumption}
Assumption \ref{asmp: apdx_VQK_initialization} is a straightforward combination of Assumption \ref{asmp: VQK_initialization} and (\ref{eqn: mu_j}) by applying the triangle inequality to bound the error terms for tokens.
\begin{definition}\label{def: terms}
\begin{enumerate}
    \item $\phi_n(t)=\frac{1}{|\mathcal{S}_1^n|e^{\|\bfq_1(t)\|^2+(\delta+\tau)\|\bfq_1(t)\|}+|\mathcal{S}^n|-|\mathcal{S}_1^n|}$.
    \item $\nu_n(t)=\frac{1}{|\mathcal{S}_1^n|e^{\|\bfq_1(t)\|^2-(\delta+\tau)\|\bfq_1(t)\|}+|\mathcal{S}^n|-|\mathcal{S}_1^n|}$.
    \item $p_n(t)=|\mathcal{S}_1^n|e^{\|\bfq_1(t)\|^2-(\delta+\tau)\|\bfq_1(t)\|}\nu_n(t)$.
    \item $\mathcal{S}_*^n=\begin{cases}\mathcal{S}_1^n,\ &\text{if }y^n=1\\
    \mathcal{S}_2^n,\ &\text{if }y^n=-1\end{cases}$, $\mathcal{S}_\#^n=\begin{cases}\mathcal{S}_2^n,\ &\text{if }y^n=1\\
    \mathcal{S}_1^n,\ &\text{if }y^n=-1\end{cases}$
    \item $\alpha_*=\mathbb{E}\Big[\frac{|\mathcal{S}^n_*|}{|\mathcal{S}^n|}\Big]$, $\alpha_\#=\mathbb{E}\Big[\frac{|\mathcal{S}^n_\#|}{|\mathcal{S}^n|}\Big]$, $\alpha_{nd}=\sum_{l=3}^M\mathbb{E}\Big[\frac{|\mathcal{S}^n_l|}{|\mathcal{S}^n|}\Big]$.
\end{enumerate}
\end{definition}

\begin{definition}\label{def: WU}
Define 
\begin{equation}
    \bfV_l^n(t)=\bfW_V^{(t)}\bfX^n\text{softmax}({\bfX^n}^\top{\bfW_K^{(t)}}^\top\bfW_Q^{(t)}\bfx_l^n)
\end{equation}
for the token $l$ of data $n$. Define $\mathcal{W}_{l,n}(0)$, $\mathcal{U}_{l,n}(0)$ as the sets of lucky neurons such that
\begin{equation}
    \mathcal{W}_{l,n}(0)=\{i: \bfW_{O_{(i,\cdot)}}^{(0)}\bfV_{l,n}(0)>0, l\in\mathcal{S}_1^n\}
\end{equation}
\begin{equation}
    \mathcal{U}_{l,n}(0)=\{i: \bfW_{O_{(i,\cdot)}}^{(0)}\bfV_{l,n}(0)>0, l\in\mathcal{S}_2^n\}
\end{equation}
\end{definition}

\begin{assumption}\label{asmp: lambda}
For one data $\bfX^n$, if the patch $i$ and $j$ correspond to the same feature $k\in[M]$, i.e., $i\in\mathcal{S}_k^n$ and $j\in\mathcal{S}_k^n$, we have
\begin{equation}
    {{\bfx_i}^n}^\top\bfx_j^n\geq 1
\end{equation}
If the patch $i$ and $j$ correspond to the different feature $k,l\in[M]$, $k\neq l$ i.e., $i\in\mathcal{S}_k^n$ and $j\in\mathcal{S}_l^n$, $k\neq l$, we have
\begin{equation}
    {{\bfx_i}^n}^\top\bfx_j^n\leq \lambda<1
\end{equation}
\end{assumption}
This assumption is equivalent to the data model by (\ref{eqn: mu_j}) since $\tau<O(1/M)$. For the simplicity of presentation, we scale up all tokens a little bit to make the threshold of linear separability be $1$. We also take $1-\lambda$ and $\lambda$ as $\Theta(1)$ for the simplicity.

\begin{definition}\label{def: sub-Gaussian}\citep{V10}
We say $X$ is a sub-Gaussian random variable with sub-Gaussian norm $K>0$, if $(\mathbb{E}|X|^p)^{\frac{1}{p}}\leq K\sqrt{p}$ for all $p\geq 1$. In addition, the sub-Gaussian norm of X, denoted $\|X\|_{\psi_2}$, is defined as $\|X\|_{\psi_2}=\sup_{p\geq 1}p^{-\frac{1}{2}}(\mathbb{E}|X|^p)^{\frac{1}{p}}$.
\end{definition}

\begin{lemma}  (\cite{V10} 
Proposition 5.1,  Hoeffding's inequality)  Let $X_1, X_2, \cdots, X_N$ be independent centered sub-gaussian random variables, and let $K=\max_i\|\bfX_i\|_{\psi_2}$. Then for every $\bfa=(a_1,\cdots,a_N)\in\mathbb{R}^N$ and every $t\geq0$, we have
\begin{equation}
    \mathbb{P}\Big\{\Big|\sum_{i=1}^N a_i X_i\Big|\geq t\Big\}\leq e\cdot \exp(-\frac{ct^2}{K^2\|\bfa\|^2})\label{hoeffding}
\end{equation}
where $c>0$ is an absolute constant.
\end{lemma}

\section{Proof of the Main Theorem and Propositions}\label{sec: proof_thm}

We state Lemma \ref{lm: training} first before we introduce the proof of main theorems. Lemma \ref{lm: training} is the key lemma in our paper to show the training process of our ViT model using SGD. It has three major claims. Claim \ref{clm: W_O} involves the growth of $\bfW_O^{(t)}$ in terms of different directions of $\bfp_l$, $i\in[M]$. Claim \ref{clm: W_QK} describes the training dynamics of $\bfW_Q^{(t)}$ and $\bfW_K^{(t)}$ separately to show the tendency to a sparse attention map. Claim \ref{clm: W_V} studies the gradient update process of $\bfW_V^{(t)}$. \\
\begin{lemma}\label{lm: training}
For $l\in\mathcal{S}_1^n$ for the data with $y^n=1$, define
\begin{equation}
\begin{aligned}
    \bfV_l^n(t)=\bfW_V^{(t)}\bfX^n\text{softmax}({\bfX^n}^\top{\bfW_K^{(t)}}^\top\bfW_Q^{(t)}\bfx_l^n)
\end{aligned}
\end{equation}
We later show that
\begin{equation}
    \begin{aligned}
        \bfV_l^n(t)=&\sum_{s\in\mathcal{S}_1^n}\text{softmax}({\bfx_s^n}^\top{\bfW_K^{(t)}}^\top\bfW_Q^{(t)}\bfx_l^n)\bfp_1+\bfz(t)+\sum_{j\neq 1}W_j^n(t)\bfp_j\\
&-\eta \sum_{b=1}^t(\sum_{i\in \mathcal{W}_{l,n}(0)}V_i(b){\bfW_{O_{(i,\cdot)}}^{(b)}}^\top+\sum_{i\notin \mathcal{W}_{l,n}(0)}V_i(b)\lambda{\bfW_{O_{(i,\cdot)}}^{(b)}}^\top)\label{V_l0}
    \end{aligned}
\end{equation}
where
\begin{equation}
    W_l^n(t)\leq \nu_n(t)|\mathcal{S}_j^n|
\end{equation}
\begin{equation}
    V_i(t)
    \lesssim \frac{1}{2B}\sum_{n\in{\mathcal{B}_b}_+}-\frac{|\mathcal{S}_1^n|}{a|\mathcal{S}^n|}p_n(t),\ \ \ \ i\in\mathcal{W}_{l,n}(0)
\end{equation}
\begin{equation}
    V_i(t)\gtrsim \frac{1}{2B}\sum_{n\in{\mathcal{B}_b}_-}\frac{|\mathcal{S}_2^n|}{a|\mathcal{S}^n|}p_n(t),\ \ \ \ i\in\mathcal{U}_{l,n}(0)
\end{equation}
\begin{equation}
    |V_i(t)|\leq\frac{1}{\sqrt{B}aM},\ \ \ \ \text{if }i\text{ is an unlucky neuron.}
\end{equation}

We also have the following claims when $m\gtrsim M^2\log N$:

\begin{claim}\label{clm: W_O} For the lucky neuron $i\in\mathcal{W}_{l,n}(0)$ and $b\in[T]$, we have
    \begin{equation}
        \bfW_{O_{(i,\cdot)}}^{(t)}\bfp_1\gtrsim \frac{\xi}{aB}\sum_{n\in\mathcal{B}_b}\frac{\eta t^2 |\mathcal{S}_1^n|}{|\mathcal{S}^n|}\frac{1}{4B}\sum_{n\in\mathcal{B}_b}\frac{|\mathcal{S}_1^n|m}{|\mathcal{S}^n|a}p_n(t)  +\xi\label{W_O_p_1}
    \end{equation}
    \begin{equation}
        \bfW_{O_{(i,\cdot)}}^{(t)}\bfp\lesssim \frac{1}{\sqrt{B}}\bfW_{O_{(i,\cdot)}}^{(t)}\bfp_1,\ \ \ \ \text{for }\bfp\in\mathcal{P}\backslash\bfp_1,
    \end{equation}
    \begin{equation}
\|\bfW_{O_{(i,\cdot)}}^{(t)}\|^2\geq (\frac{\xi}{aB}\sum_{n\in\mathcal{B}_b}\frac{\eta t^2 |\mathcal{S}_1^n|}{|\mathcal{S}^n|}\frac{1}{4B}\sum_{n\in\mathcal{B}_b}\frac{|\mathcal{S}_1^n|m}{|\mathcal{S}^n|a}p_n(t)  +\xi)^2\label{W_O_two_side_bound}
    \end{equation}
    and for the noise $\bfz_l(t)$,
    \begin{equation}
        \|\bfW_{O_{(i, \cdot)}}^{(t)}\bfz_l(t)\|
    \leq (\sigma+\tau) \|\bfW_{O_{(i, \cdot)}}^{(t)}\|\label{W_O_noise}
    \end{equation}
    For $i\in\mathcal{U}_{l,n}(0)$, we also have equations as in (\ref{W_O_p_1}) to (\ref{W_O_noise}), including
        \begin{equation}
        \bfW_{O_{(i,\cdot)}}^{(t)}\bfp_2\gtrsim \frac{\xi}{aB}\sum_{n\in\mathcal{B}_b}\frac{\eta t^2 |\mathcal{S}_1^n|}{|\mathcal{S}^n|}\frac{1}{4B}\sum_{n\in\mathcal{B}_b}\frac{|\mathcal{S}_1^n|m}{|\mathcal{S}^n|a}p_n(t)  +\xi\label{W_O_p_2}
    \end{equation}
    \begin{equation}
        \bfW_{O_{(i,\cdot)}}^{(t)}\bfp\lesssim \frac{1}{\sqrt{B}}\bfW_{O_{(i,\cdot)}}^{(t)}\bfp_1,\ \ \ \ \text{for }\bfp\in\mathcal{P}\backslash\bfp_2,
    \end{equation}
    \begin{equation}
       \|\bfW_{O_{(i,\cdot)}}^{(t)}\|^2\geq (\frac{\xi}{aB}\sum_{n\in\mathcal{B}_b}\frac{\eta t^2 |\mathcal{S}_1^n|}{|\mathcal{S}^n|}\frac{1}{4B}\sum_{n\in\mathcal{B}_b}\frac{|\mathcal{S}_1^n|m}{|\mathcal{S}^n|a}p_n(t)  +\xi)^2\label{W_O_two_side_bound2}
    \end{equation}
    and for the noise $\bfz_l(t)$,
    \begin{equation}
        \|\bfW_{O_{(i, \cdot)}}^{(t)}\bfz_l(t)\|
    \leq (\sigma+\tau) \|\bfW_{O_{(i, \cdot)}}^{(t)}\|\label{W_O_noise2}
    \end{equation}
    For unlucky neurons $i$ and $j\in\mathcal{W}_{l,n}(0)$, $k\in\mathcal{U}_{l,n}(0)$, $p\in\mathcal{P}$, we have
    \begin{equation}
        \bfW_{O_{(i,\cdot)}}^{(t)}\bfp\leq \frac{1}{\sqrt{B}}\min\{\bfW_{O_{(j,\cdot)}}^{(t)}\bfp_1, \bfW_{O_{(k,\cdot)}}^{(t)}\bfp_2\},\label{W_O_p_3}
    \end{equation}
        \begin{equation}
        \|\bfW_{O_{(i, \cdot)}}^{(t)}\bfz_l(t)\|
    \leq (\sigma+\tau)\|\bfW_{O_{(j,\cdot)}}^{(t)}\|
    \end{equation}
    \begin{equation}
     \|\bfW_{O_{(i,\cdot)}}^{(t)}\|^2\leq \frac{1}{B}\min\{\|\bfW_{O_{(j,\cdot)}}^{(t)}\|^2, \|\bfW_{O_{(k,\cdot)}}^{(t)}\|^2\}\label{W_O_noise_3}
    \end{equation}
    \end{claim}
\begin{claim} \label{clm: W_QK} Given conditions in (\ref{eqn: alpha*}), there exists $K(t), Q(t)>0$, where $t$ is large enough before the end of training, such that for $j\in\mathcal{S}_*^n$,
    \begin{equation}
    \text{softmax}({\bfx_j^n}^\top\bfW_K^{(t+1)}\bfW_Q^{(t+1)}\bfx_l^n)\gtrsim \frac{e^{(1+K(t))\|\bfq_1(t)\|^2-(\delta+\tau)\|\bfq_1(t)\|}}{|\mathcal{S}_1^n|e^{(1+K(t))\|\bfq_1(t)\|^2-(\delta+\tau)\|\bfq_1(t)\|}+(|\mathcal{S}^n|-|\mathcal{S}_1^n|)}\label{softmax_t1_same}
    \end{equation}
    \begin{equation}
    \begin{aligned}
       &\text{softmax}({\bfx_j^n}^\top{\bfW_K^{(t+1)}}^\top\bfW_Q^{(t+1)}\bfx_j^n)-\text{softmax}({\bfx_j^n}^\top{\bfW_K^{(t)}}^\top\bfW_Q^{(t)}\bfx_l^n)\\
    \gtrsim & \frac{|\mathcal{S}^n|-|\mathcal{S}_1^n|}{(|\mathcal{S}_1^n|e^{(1+K(t))\|\bfq_1(t)\|^2-(\delta+\tau)\|\bfq_1(t)\|}+(|\mathcal{S}^n|-|\mathcal{S}_1^n|))^2}e^{\|\bfq_1(t)\|^2-(\delta+\tau)\|\bfq_1(t)\|}\cdot K(t),
    \end{aligned}
    \end{equation}
    and for $j\notin\mathcal{S}_*^n$, we have
    \begin{equation}
    \text{softmax}({\bfx_j^n}^\top{\bfW_K^{(t+1)}}^\top\bfW_Q^{(t+1)}\bfx_l^n)\lesssim \frac{1}{|\mathcal{S}_1^n|e^{(1+K(t))\|\bfq_1(t)\|^2-\delta\|\bfq_1(t)\|}+(|\mathcal{S}^n|-|\mathcal{S}_1^n|)}\label{softmax_t1_diff}
    \end{equation}
    \begin{equation}
    \begin{aligned}
       &\text{softmax}({\bfx_j^n}^\top{\bfW_K^{(t+1)}}\bfW_Q^{(t+1)}\bfx_l^n)-\text{softmax}({\bfx_j^n}^\top{\bfW_K^{(t)}}^\top\bfW_Q^{(t)}\bfx_l^n)\\
    \lesssim & -\frac{|\mathcal{S}_1^n|}{(|\mathcal{S}_1^n|e^{(1+K(t))\|\bfq_1(t)\|^2-\delta\|\bfq_1(t)\|}+(|\mathcal{S}^n|-|\mathcal{S}_1|))^2}e^{\|\bfq_1(t)\|^2-\delta\|\bfq_1(t)\|}\cdot K(t)
    \end{aligned}
    \end{equation}
    For $i=1,2$,
    \begin{equation}
        \bfq_i(t)=\sqrt{\prod_{l=0}^{t-1}(1+K(l))}\bfq_i
    \end{equation}
    \begin{equation}
        \bfr_i(t)=\sqrt{\prod_{l=0}^{t-1}(1+Q(l))}\bfr_i
    \end{equation}
    \end{claim}
\begin{claim}\label{clm: W_V}
    For the update of $\bfW_V^{(t)}$, there exists $\lambda\leq \Theta(1)$ such that
    \begin{equation}
        \bfW_V^{(t)}\bfx_j^n
    =\bfp_1-\eta\sum_{b=1}^t(\sum_{i\in\mathcal{W}_{l,n}(0)} V_i(b){\bfW_{O_{(i,\cdot)}}^{(b)}}^\top+\sum_{i\notin\mathcal{W}_{l,n}(0)} \lambda V_i(b){\bfW_{O_{(i,\cdot)}}^{(b)}}^\top)+\bfz_j(t),\ \ \ \ j\in\mathcal{S}_1^n
    \end{equation}
    \begin{equation}
        \bfW_V^{(t)}\bfx_j^n
    =\bfp_2-\eta\sum_{b=1}^t(\sum_{i\in\mathcal{U}_{l,n}(0)} V_i(b){\bfW_{O_{(i,\cdot)}}^{(b)}}^\top+\sum_{i\notin\mathcal{U}_{l,n}(0)} \lambda V_i(b){\bfW_{O_{(i,\cdot)}}^{(b)}}^\top)+\bfz_j(t),\ \ \ \ j\in\mathcal{S}_2^n
    \end{equation}
    \begin{equation}
        \bfW_V^{(t+1)}\bfx_j^n
    =\bfp_j-\eta \sum_{b=1}^t\sum_{i=1}^m \lambda V_i(b){\bfW_{O_{(i,\cdot)}}^{(b)}}^\top+\bfz_j(t),\ \ \ \ j\in[|\mathcal{S}^n|]/(\mathcal{S}_1^n\cup\mathcal{S}_2^n)
    \end{equation}
    \begin{equation}
        \|\bfz_j(t)\|\leq (\sigma+\tau)\label{noise_bound}
    \end{equation}
\end{claim}

\end{lemma}

To prove Theorem \ref{thm: main_theory}, we either show $F(\bfX^n)>1$ for $y^n=1$ or show $F(\bfX^n)<-1$ for $y^n=-1$. Take $y^n=1$ as an example, the basic idea of the proof is to make use of Lemma \ref{lm: training} to find a lower bound as a function of $\alpha_*$, $\sigma$, $\tau$, etc.. The remaining step is to derive conditions on the sample complexity and the required number of iterations in terms of $\alpha_*$, $\sigma$, and $\tau$ such that the lower bound is greater than $1$. Given a balanced dataset, these conditions also ensure that $F(\bfX^n)<-1$ for $y^n=-1$. During the proof, we may need to use some of equations as intermediate steps in the proof of Lemma \ref{lm: training}. Since that these equations are not concise for presentation, we prefer not to state them formally in Lemma \ref{lm: training}, but still refer to them as useful conclusions. The following is the details of the proof.\\
\noindent \textbf{Proof of Theorem \ref{thm: main_theory}: }\\
For $y^n=1$, define $\mathcal{K}_+=\{i\in[m]: a_i>0\}$ and $\mathcal{K}_-=\{i\in[m]: a_i<0\}$. We have
\begin{equation}
    \begin{aligned}
    F(\bfX^n)=&\frac{1}{|\mathcal{S}^n|}\sum_{l\in\mathcal{S}^n}\sum_{i\in\mathcal{W}_{l,n}(0)}\frac{1}{a}\text{Relu}(\bfW_{O_{(i, \cdot)}}^{(t)}\bfV_l^n(t))+\frac{1}{|\mathcal{S}^n|}\sum_{l\in\mathcal{S}^n}\sum_{i\in\mathcal{K}_+\backslash\mathcal{W}_{l,n}(0)}\frac{1}{a}\text{Relu}(\bfW_{O_{(i, \cdot)}}^{(t)}\bfV_l^n(t))\\
    &-\frac{1}{|\mathcal{S}^n|}\sum_{l\in\mathcal{S}^n}\sum_{i\in\mathcal{K}_-}\frac{1}{a}\text{Relu}(\bfW_{O_{(i, \cdot)}}^{(t)}\bfV_l^n(t))
    \end{aligned}
\end{equation}

\noindent By Lemma \ref{lm: training}, we have that when $m\gtrsim  M^2\log N$, 
\begin{equation}
    \begin{aligned}
    &\frac{1}{|\mathcal{S}^n|}\sum_{l\in\mathcal{S}^n}\sum_{i\in\mathcal{W}_{l,n}(0)}\frac{1}{a}\text{Relu}(\bfW_{O_{(i, \cdot)}}^{(t)}\bfV_l^n(t))\\=& \frac{1}{|\mathcal{S}^n|}\sum_{l\in\mathcal{S}_1^n}\sum_{i\in\mathcal{W}_{l,n}(0)}\frac{1}{a}\text{Relu}(\bfW_{O_{(i, \cdot)}}^{(t)}\bfV_l^n(t))+\sum_{l\notin\mathcal{S}_1^n}\sum_{i\in\mathcal{W}_{l,n}(0)}\frac{1}{m}\text{Relu}(\bfW_{O_{(i, \cdot)}}^{(t)}\bfV_l^n(t))\\
    \gtrsim &|\mathcal{S}_1^n|\frac{1}{a|\mathcal{S}^n|}\cdot \bfW_{O_{(i,\cdot)}}^{(t)}\Big(\sum_{s\in\mathcal{S}_1^n} \bfp_s\text{softmax}({\bfx_s^n}^\top{\bfW_K^{(t)}}^\top\bfW_Q^{(t)}{\bfx^n}_l)+\bfz(t)+\sum_{l\neq s}W_l(u)\bfp_l\\
    &-\eta t(\sum_{j\in\mathcal{W}_{l,n}(0)} V_j(t){\bfW_{O_{(j,\cdot)}}^{(t)}}^\top+\sum_{j\notin\mathcal{W}_{l,n}(0)} V_j(t)\lambda {\bfW_{O_{(j,\cdot)}}^{(t)}}^\top)\Big)|\mathcal{W}_{l,n}(0)|+0\\
    \gtrsim &\frac{|\mathcal{S}_1^n|m}{|\mathcal{S}^n|a}  \Big(\frac{1}{B}\sum_{n\in\mathcal{B}_b}\frac{\xi\eta (t+1)^2 |\mathcal{S}_1^n|m}{|\mathcal{S}^n|a^2}(\frac{1}{4B}\sum_{n\in\mathcal{B}_b}\frac{|\mathcal{S}_1^n|}{|\mathcal{S}^n|}p_n(b)-\sigma-\tau)\\
    &\cdot  p_n(t)+ \eta  m  \frac{1}{2B}\sum_{b=1}^t\sum_{n\in{\mathcal{B}_b}_+}\frac{|\mathcal{S}_1^n|}{a|\mathcal{S}^n|}p_n(b)\\
    &\cdot (\frac{\xi}{aB}\sum_{n\in\mathcal{B}_b}\frac{\eta (t+1)^2 |\mathcal{S}_1^n|}{|\mathcal{S}^n|}\frac{1}{4B}\sum_{n\in\mathcal{B}_b}\frac{|\mathcal{S}_1^n|m}{|\mathcal{S}^n|a}p_n(t)  )^2\Big)
    \end{aligned}
\end{equation}
where the second step comes from (\ref{V_l0}) and the last step is by (\ref{s_Omega_1}). By the definition of $\mathcal{K}_+^l$, we have
\begin{equation}
    \begin{aligned}
    \frac{1}{|\mathcal{S}^n|}\sum_{l\in\mathcal{S}^n}\sum_{i\in\mathcal{K}_+\backslash \mathcal{W}_{l,n}(0)}\frac{1}{a}\text{Relu}(\bfW_{O_{(i, \cdot)}}^{(t)}\bfV_l^n(t))\geq 0
\end{aligned}
\end{equation}
We can obtain that
\begin{equation}
    \frac{1}{|\mathcal{S}^n|}\sum_{l\in\mathcal{S}^n}\sum_{i\in\mathcal{U}_{l,n}(0)}\frac{1}{a}\text{Relu}(\bfW_{O_{(i, \cdot)}}^{(t)}\bfV_l^n(t))\leq \frac{|\mathcal{S}_2^n|}{|\mathcal{S}_1^n|}\frac{1}{|\mathcal{S}^n|}\sum_{l\in\mathcal{S}^n}\sum_{i\in\mathcal{W}_{l,n}(0)}\frac{1}{a}\text{Relu}(\bfW_{O_{(i, \cdot)}}^{(t)}\bfV_l^n(t))
\end{equation}
Combining (\ref{s_Omega_1}) and (\ref{s_Omega_other}), we can obtain
\begin{equation}
    \begin{aligned}
        &\frac{1}{|\mathcal{S}^n|}\sum_{l\in\mathcal{S}^n}\sum_{i\in\mathcal{K}_-\backslash \mathcal{U}_{l,n}(0)}\frac{1}{a}\text{Relu}(\bfW_{O_{(i, \cdot)}}^{(t)}\bfV_l^n(t))\\
        \lesssim  & \frac{|\mathcal{S}_1^n|m}{|\mathcal{S}^n|a}\cdot \frac{1}{\sqrt{B}}\frac{1}{|\mathcal{S}^n|}\sum_{l\in\mathcal{S}^n}\sum_{i\in \mathcal{W}_{l,n}(0)}\frac{1}{a}\text{Relu}(\bfW_{O_{(i, \cdot)}}^{(t)}\bfV_l^n(t))
    \end{aligned}
\end{equation}

\noindent Note that at the $T$-th iteration where $\eta T=\Theta(1)$,
\begin{equation}
\begin{aligned}
   &K(t)\\
   \gtrsim &\eta\frac{1}{B}\sum_{n\in\mathcal{B}_b}\frac{|\mathcal{S}_1^n|m}{|\mathcal{S}^n|a}p_n(t)\Big(\frac{1}{B}\sum_{n\in\mathcal{B}_b}\frac{\xi\eta (t+1)^2 |\mathcal{S}_1^n|m}{|\mathcal{S}^n|a^2}(\frac{1}{4B}\sum_{n\in\mathcal{B}_b}\frac{|\mathcal{S}_1^n|}{|\mathcal{S}^n|}p_n(b)-\sigma-\tau)\\
    &\cdot  + \eta  m  \frac{1}{2B}\sum_{b=1}^t\sum_{n\in{\mathcal{B}_b}_+}\frac{|\mathcal{S}_1^n|}{a|\mathcal{S}^n|}p_n(b)\\
    &\cdot (\frac{\xi}{aB}\sum_{n\in\mathcal{B}_b}\frac{\eta (t+1)^2 |\mathcal{S}_1^n|}{|\mathcal{S}^n|}\frac{1}{4B}\sum_{n\in\mathcal{B}_b}\frac{|\mathcal{S}_1^n|m}{|\mathcal{S}^n|a}p_n(t)  )^2\Big)\\
    &\cdot\phi_n(t)(|\mathcal{S}^n|-|\mathcal{S}_1^n|)\|\bfq_1(t)\|^2\\
    \gtrsim &\frac{\eta}{e^{\|\bfq_1(t)\|^2-(\delta+\tau)\|\bfq_1(t)\|}}
\end{aligned}
\end{equation}
Since that
\begin{equation}
\begin{aligned}
    \bfq_1(T)&\gtrsim(1+\min_{l=0,1,\cdots,T-1}\{K(l)\})^{T}\\
    &\gtrsim (1+\frac{\eta}{e^{\|\bfq_1(T)\|^2-(\delta+\tau)\|\bfq_1(T)\|}})^T
\end{aligned}
\end{equation}
To find the order-wise lower bound of $\bfq_1(T)$, we need to check the equation
\begin{equation}
\begin{aligned}
    \bfq_1(T)
    &\lesssim (1+\frac{1}{e^{\|\bfq_1(T)\|^2-(\delta+\tau)\|\bfq_1(T)\|}})^{T}
\end{aligned}
\end{equation}

One can obtain
\begin{equation}
\Theta(\log T)=\|\bfq_1(T)\|^2= o(T)\label{logT}
\end{equation}

Therefore,
\begin{equation}
    p_n(T)\gtrsim \frac{ {T}^{C}}{ {T}^{C}+\frac{1-\alpha}{\alpha}}\geq 1-\frac{1}{\frac{\alpha}{1-\alpha}(\eta^{-1})^{C}}\geq  1-\Theta(\eta^{C})\label{p_n(T)}
\end{equation}

\begin{equation}
    \phi_n(T)(|\mathcal{S}^n|-|\mathcal{S}_1^n|)\leq \eta^{C}
\end{equation}
for some large $C>0$.
\noindent We require that
\begin{equation}
\begin{aligned}
   &\frac{|\mathcal{S}_1^n|m}{|\mathcal{S}^n|a}  \Big(\frac{1}{B}\sum_{n\in\mathcal{B}_b}\frac{\xi\eta (t+1)^2 |\mathcal{S}_1^n|m}{|\mathcal{S}^n|a^2}(\frac{1}{4B}\sum_{n\in\mathcal{B}_b}\frac{|\mathcal{S}_1^n|}{|\mathcal{S}^n|}p_n(b)-\sigma-\tau)\\
    &\cdot  p_n(t)+ \eta  m  \frac{1}{2B}\sum_{b=1}^t\sum_{n\in{\mathcal{B}_b}_+}\frac{|\mathcal{S}_1^n|}{a|\mathcal{S}^n|}p_n(b)\\
    &\cdot(\frac{\xi}{aB}\sum_{n\in\mathcal{B}_b}\frac{\eta (t+1)^2 |\mathcal{S}_1^n|}{|\mathcal{S}^n|}\frac{1}{4B}\sum_{n\in\mathcal{B}_b}\frac{|\mathcal{S}_1^n|m}{|\mathcal{S}^n|a}p_n(t)  )^2\Big)\\
   :=&a_0\eta^3 T^5+a_1 \eta T^2\\
   >&1,\label{gt1}
  \end{aligned}
\end{equation}
where the first step is by letting $a=\sqrt{m}$ and $\xi=1/\sqrt{m}$. We replace $p_n(b)$ with $p_n(T)$ because when $b$ achieves the level of $T$, $b^{o_1}p_n(b)^{o_2}$ is close to $b^{o_1}$ for $o_1, o_2\geq0$ by (\ref{p_n(T)}). Thus,
\begin{equation}
    \sum_{b=1}^T b^{o_1}p_n(b)^{o_2}\gtrsim T^{o_1+1}p_n(\Theta(1)\cdot T)^{o_2}\gtrsim T^{o_1+1}p_n(T)^{o_2}
\end{equation}
We also require
\begin{equation}
    B\geq \Omega(1),
\end{equation}
for some $\epsilon_0>0$.

\noindent We know that
\begin{equation}
    \begin{aligned}
    &\Big|\frac{1}{N}\sum_{n=1}^N \frac{|\mathcal{S}_*^n|}{|\mathcal{S}^n|}p_n(T)(p_n(T)-(\sigma+\tau))-\mathbb{E}\Big[\frac{|\mathcal{S}_*^n|}{|\mathcal{S}^n|}\Big]\Big|\\
    \leq &\Big|\frac{1}{N}\sum_{n=1}^N \frac{|\mathcal{S}_*^n|}{|\mathcal{S}^n|}p_n(T)(p_n(T)-(\sigma+\tau))-\mathbb{E}\Big[\frac{|\mathcal{S}_*^n|}{|\mathcal{S}^n|}p_n(T)(p_n(T)-(\sigma+\tau))\Big]\Big|\\
    &+\Big|\mathbb{E}\Big[\frac{|\mathcal{S}_*^n|}{|\mathcal{S}^n|}\Big(p_n(T)(p_n(T)-(\sigma+\tau))-1\Big)\Big]\Big|\\
    \lesssim & \sqrt{\frac{\log N}{N}}+c'(1-\zeta)+c''((\sigma+\tau))
    \end{aligned}\label{N_bound_derivation}
\end{equation}
for $c'>0$ and $c''>0$. We can then have
\begin{equation}
\begin{aligned}
    t\geq T =&\frac{\eta^{-\frac{3}{5}}}{a_1}=\frac{\eta^{-\frac{3}{5}}}{\frac{|\mathcal{S}_1^n|}{|\mathcal{S}^n|}  \frac{1}{N}\sum_{n=1}^N(\frac{|\mathcal{S}_*^n|}{|\mathcal{S}^n|}\|\bfp_1\|^2p_n(t)-(\sigma+\tau))p_n(t)}\\
    =&\Theta(\frac{\eta^{-\frac{3}{5}}}{  (\mathbb{E}\Big[\frac{|\mathcal{S}_*^n|}{|\mathcal{S}^n|}\Big]-\sqrt{\frac{\log N}{N}}-c'(1-\zeta)-c''(\sigma+\tau))})\\
    =& \Theta(\frac{\eta^{-\frac{3}{5}}}{  \mathbb{E}\Big[\frac{|\mathcal{S}_*^n|}{|\mathcal{S}^n|}\Big]})
\end{aligned}
\end{equation}
where \begin{equation}
    \alpha\geq \frac{1-\alpha_{nd}}{1+ e^{-(\delta+\tau)}(1-(\sigma+\tau))}
\end{equation}
by (\ref{alpha_proof}), as long as 
\begin{equation}
    N\geq \Omega(\frac{1}{(\alpha-c'(1-\zeta)-c''((\sigma+\tau)))^{2}})\label{sp_thm1}
\end{equation}
where $\zeta\geq1-\eta^{10}$. If there is no mechanism like the self-attention to compute the weight using the correlations between tokens, we have
\begin{equation}
    c'(1-\zeta)=O(\alpha_*(1-\alpha_*)),
\end{equation}
which can scale up the sample complexity in (\ref{sp_thm1}) by $\alpha_*^{-2}$.\\
\noindent Therefore, we can obtain
\begin{equation}
    F(\bfX^n)> 1
\end{equation}

\noindent Similarly, we can derive that for $y=-1$,
\begin{equation}
    F(\bfX)<-1
\end{equation}
Hence, for all $n\in[N]$,
\begin{equation}
    \text{Loss}(\bfX^n, y^n)=0
\end{equation}
We also have 
\begin{equation}
    \text{Loss}=\mathbb{E}_{(\bfX^n,y^n)\sim\mathcal{D}}[\text{Loss}(\bfX^n,y^n)]=0
\end{equation}
with the conditions of sample complexity and the number of iterations. 

\noindent \textbf{Proof of Proposition \ref{prpst: ViT_CNN}: }\\
The main proof is the same as the proof of Theorem \ref{thm: main_theory}. The only difference is that we need to modify (\ref{N_bound_derivation}) as follows
\begin{equation}
    \begin{aligned}
    &\Big|\frac{1}{N}\sum_{n=1}^N \frac{|\mathcal{S}_*^n|}{|\mathcal{S}^n|}p_n(T)(p_n(T)-(\sigma+\tau))-\mathbb{E}\Big[\frac{|\mathcal{S}_*^n|}{|\mathcal{S}^n|}\Big]\Big|\\
    \leq &\Big|\frac{1}{N}\sum_{n=1}^N \frac{|\mathcal{S}_*^n|}{|\mathcal{S}^n|}p_n(0)(p_n(0)-(\sigma+\tau))-\mathbb{E}\Big[\frac{|\mathcal{S}_*^n|}{|\mathcal{S}^n|}p_n(0)(p_n(T)-(\sigma+\tau))\Big]\Big|\\
    &+\Big|\mathbb{E}\Big[\frac{|\mathcal{S}_*^n|}{|\mathcal{S}^n|}\Big(p_n(0)(p_n(0)-(\sigma+\tau))-1\Big)\Big]\Big|\\
    \lesssim & \sqrt{\frac{\log N}{N}}+|1-\Theta(\alpha_*^2)+\Theta(\alpha_*)(\sigma+\tau)|
    \end{aligned}
\end{equation}
where the first step is because $p_n(T)$ does not update since $\bfW_K^{(t)}$ and $\bfW_Q^{(t)}$ are fixed at initialization $\bfW_K^{(0)}$ and $\bfW_Q^{(0)}$, and the second step is by $p_n(0)=\Theta(\alpha_*)$. Since that 
\begin{equation}
    \sqrt{\frac{\log N}{N}}+|1-\Theta(\alpha_*^2)+\Theta(\alpha_*)(\sigma+\tau)|\leq \Theta(1)\cdot\alpha_*,
\end{equation}
we have
\begin{equation}
\begin{aligned}
    N\geq &\frac{1}{(\Theta(\alpha_*)-1+\Theta(\alpha_*^2)-\Theta(\alpha_*)(\sigma+\tau))^2}\\
    &=\Omega(\frac{1}{(\alpha_*(\alpha_*-\sigma-\tau))^2})
\end{aligned}
\end{equation}

\noindent \textbf{Proof of Proposition \ref{lemma: concentration}: }\\
It can be easily derived from Claim \ref{clm: W_QK} of Lemma \ref{lm: training}, (\ref{logT}), and (\ref{p_n(T)}).

\section{Proof of Lemma \ref{lm: training}}\label{sec: lemma 2}

We prove the whole lemma by a long induction, which is the reason why we prefer to wrap three claims into one lemma. To make it easier to follow, however, we break this Section into three parts to introduce the proof of three claims of Lemma \ref{lm: training} separately.\\
\subsection{\textbf{Proof of Claim \ref{clm: W_O} of Lemma \ref{lm: training}}}
\noindent Although it looks cumbersome, the key idea of Claim \ref{clm: W_O} is to characterize the growth of $\bfW_O^{(t)}$ in terms of $\bfp_l$, $l\in[M]$. We compare $\bfW_{O_{(i,\cdot)}}^{(t+1)}\bfp_l$ and $\bfW_{O_{(i,\cdot)}}^{(t)}\bfp_l$ to see the direction of growth by computing the gradient. One can eventually find that lucky neurons grow the most in directions of $\bfp_1$ and $\bfp_2$, i.e., the feature of label-relevant patterns, while unlucky neurons do not change much in magnitude. \\
\noindent We start our poof. At the $t$-th iteration ($t>1$), if $l\in\mathcal{S}_1^n$, let 
\begin{equation}
\begin{aligned}
\bfV_l^n(t)&=\bfW_V^{(t)}\bfX^n\text{softmax}({\bfX^n}^\top{\bfW_K^{(t)}}^\top\bfW_Q^{(t)}\bfx_l^n)\\
&=\sum_{s\in\mathcal{S}_1}\text{softmax}({\bfx_s^n}^\top{\bfW_K^{(t)}}^\top\bfW_Q^{(t)}\bfx_l^n)\bfp_1+\bfz(t)+\sum_{j\neq 1}W_j^n(t)\bfp_j\\
&-\eta\sum_{b=0}^{t-1}(\sum_{i\in \mathcal{W}_{l,n}(0)}V_i(b){\bfW_{O_{(i,\cdot)}}^{(b)}}^\top+\sum_{i\notin \mathcal{W}_{l,n}(0)}V_i(b)\lambda{\bfW_{O_{(i,\cdot)}}^{(b)}}^\top)\label{V_l(t)_first}
\end{aligned}
\end{equation}, $l\in[M]$, where the second step comes from Claim \ref{clm: W_V} of Lemma \ref{lm: training}. We can derive
\begin{equation}
    0< W_l^n(t)\leq \frac{|\mathcal{S}_j^n|e^{\delta\|\bfq_1(t)\|}}{(|\mathcal{S}^n|-|\mathcal{S}_1^n|)e^{\delta\|\bfq_1(t)\|}+|\mathcal{S}_1^n|e^{\|\bfq_1(t)\|^2-\delta\|\bfq_1(t)\|}}=\nu_n(t)|\mathcal{S}_j^n|
\end{equation}
which is much smaller than $\Theta(1)$ when $t$ is large. This is the  reason why we ignore the impact of $W_l^n(t)$ on $\eta\sum_{b=0}^{t-1}(\sum_{i\in \mathcal{W}_{l,n}(0)}V_i(b){\bfW_{O_{(i,\cdot)}}^{(b)}}^\top+\sum_{i\notin \mathcal{W}_{l,n}(0)}V_i(b)\lambda{\bfW_{O_{(i,\cdot)}}^{(b)}}^\top)$ in (\ref{V_l(t)_first}). 
Hence, by (\ref{loss_batch}),
\begin{equation}
    \frac{\partial \overline{\textbf{Loss}}_b}{\partial {\bfW_{O_{(i, \cdot)}}}^\top}=-\frac{1}{B}\sum_{n\in\mathcal{B}_b} y^n\frac{1}{|\mathcal{S}^n|}\sum_{l\in\mathcal{S}^n} a_{(l)_i} \mathbbm{1}[\bfW_{O_{(i, \cdot)}}\bfV_l^n(t)\geq0]{\bfV_l^n(t)}^\top 
\end{equation}
Define that for $j\in[M]$,
\begin{equation}
    \begin{aligned}
        I_4=&\frac{1}{B}\sum_{n\in\mathcal{B}_b}\eta y^n\frac{1}{|\mathcal{S}^n|}\sum_{l\in\mathcal{S}^n} a_{(l)_i} \mathbbm{1}[\bfW_{O_{(i, \cdot)}}^{(t)}\bfV_l^n(t)\geq0]\sum_{k\in \mathcal{W}_{l,n}(0)}V_k(t)\bfW_{O_{(k,\cdot)}}^{(t)}\bfp_j
    \end{aligned}
\end{equation}
\begin{equation}
    \begin{aligned}
        I_5=&\frac{1}{B}\sum_{n\in\mathcal{B}_b}\eta y^n\frac{1}{|\mathcal{S}^n|}\sum_{l\in\mathcal{S}^n} a_{(l)_i} \mathbbm{1}[\bfW_{O_{(i, \cdot)}}^{(t)}\bfV_l^n(t)\geq0]\sum_{k\notin \mathcal{W}_{l,n}(0)}V_k(t)\bfW_{O_{(k,\cdot)}}^{(t)}\bfp_j,
    \end{aligned}
\end{equation}
and we can then obtain
\begin{equation}
    \begin{aligned}
    &\left\langle{\bfW_{O_{(i, \cdot)}}^{(t+1)}}^\top, \bfp_j\right\rangle-\left\langle{\bfW_{O_{(i, \cdot)}}^{(t)}}^\top, \bfp_j\right\rangle\\
    =& \frac{1}{B}\sum_{n\in\mathcal{B}_b}\eta y^n\frac{1}{|\mathcal{S}^n|}\sum_{l\in\mathcal{S}^n} a_{(l)_i} \mathbbm{1}[\bfW_{O_{(i, \cdot)}}^{(t)}\bfV_l^n(t)\geq0]{\bfV_l^n(t)}^\top\bfp_j\\
    =& \frac{1}{B}\sum_{n\in\mathcal{B}_b}\eta y^n\frac{1}{|\mathcal{S}^n|}\sum_{l\in\mathcal{S}^n} a_{(l)_i} \mathbbm{1}[\bfW_{O_{(i, \cdot)}}^{(t)}\bfV_l^n(t)\geq0]\bfz_l(t)^\top\bfp_j\\
    &+\frac{1}{B}\sum_{n\in\mathcal{B}_b}\eta y^n\frac{1}{|\mathcal{S}^n|}\sum_{l\in\mathcal{S}^n} a_{(l)_i} \mathbbm{1}[\bfW_{O_{(i, \cdot)}}^{(t)}\bfV_l^n(t)\geq0]\sum_{s\in\mathcal{S}_l}\text{softmax}(\bfx_s^\top{\bfW_K^{(t)}}^\top\bfW_Q^{(t)}\bfx_l)\bfp_l^\top\bfp_j\\
    &+ \frac{1}{B}\sum_{n\in\mathcal{B}_b}\eta y^n\frac{1}{|\mathcal{S}^n|}\sum_{l\in\mathcal{S}^n} a_{(l)_i} \mathbbm{1}[\bfW_{O_{(i, \cdot)}}^{(t)}\bfV_l^n(t)\geq0]\sum_{k\neq l}W_l(t)\bfp_k^\top\bfp_j+I_4+I_5\\
    :=& I_1+I_2+I_3+I_4+I_5,
    \end{aligned}
\end{equation}
where
\begin{equation}
    I_1=\frac{1}{B}\sum_{n\in\mathcal{B}_b}\eta y^n\frac{1}{|\mathcal{S}^n|}\sum_{l\in\mathcal{S}^n} a_{(l)_i} \mathbbm{1}[\bfW_{O_{(i, \cdot)}}^{(t)}\bfV_l^n(t)\geq0]\bfz_l(t)^\top\bfp_j
\end{equation}
\begin{equation}
    I_2=\frac{1}{B}\sum_{n\in\mathcal{B}_b}\eta y^n\frac{1}{|\mathcal{S}^n|}\sum_{l\in\mathcal{S}^n} a_{(l)_i} \mathbbm{1}[\bfW_{O_{(i, \cdot)}}^{(t)}\bfV_l^n(t)\geq0]\sum_{s\in\mathcal{S}_l}\text{softmax}(\bfx_s^\top{\bfW_K^{(t)}}^\top\bfW_Q^{(t)}\bfx_l)\bfp_l^\top\bfp_j
\end{equation}
\begin{equation}
    I_3=\frac{1}{B}\sum_{n\in\mathcal{B}_b}\eta y^n\frac{1}{|\mathcal{S}^n|}\sum_{l\in\mathcal{S}^n} a_{(l)_i} \mathbbm{1}[\bfW_{O_{(i, \cdot)}}^{(t)}\bfV_l^n(t)\geq0]\sum_{k\neq l}W_l(t)\bfp_k^\top\bfp_j
\end{equation}
We then show the statements in different cases.\\
\noindent (1) When $j=1$, since that $\Pr(y^n=1)=\Pr(y^n=-1)=1/2$, by Hoeffding's inequality in (\ref{hoeffding}), we can derive
\begin{equation}
    \Pr\Big(\Big|\frac{1}{B}\sum_{n\in\mathcal{B}_b} y^n\Big|\geq \sqrt{\frac{\log B}{B}}\Big)\leq B^{-c}
\end{equation}
\begin{equation}
    \Pr\Big(\Big|\bfz_l(t)^\top\bfp_1\Big|\geq \sqrt{((\sigma+\tau))^2\log m}\Big)\leq m^{-c}
\end{equation}
Hence, with probability at least $1-(mB)^{-c}$, we have
\begin{equation}
\begin{aligned}
    |I_1|\leq &\frac{\eta((\sigma+\tau))}{a}\sqrt{\frac{\log m \log B}{B}}\label{I1_p1}
\end{aligned}
\end{equation}
For $i\in\mathcal{W}_{l,n}(0)$, by Lemma \ref{lemma: update_WU}, we have
\begin{equation}
\begin{aligned}
    &\bfW_{O_{(i,\cdot)}}^{(t)}\sum_{s=1}^L \bfW_V^{(t)}\bfx_s^n\text{softmax}({\bfx_s^n}^\top{\bfW_K^{(t)}}^\top\bfW_Q^{(t)}\bfx_l^n)>0
\end{aligned}
\end{equation}
Denote $p_n(t)=|\mathcal{S}_1^n|\nu_n(t)e^{\|\bfq_1(t)\|^2-2\delta\|\bfq_1(t)\|}$. Hence, for $k\notin\mathcal{W}_{l,n}(0)$,
\begin{equation}
\begin{aligned}
    I_2\gtrsim \eta \frac{1}{B}\sum_{n\in\mathcal{B}_b}\frac{|\mathcal{S}_1^n|}{|\mathcal{S}^n|}\cdot\frac{1}{a}\|\bfp_1\|^2\cdot p_n(t)\label{I2_p1}
\end{aligned}
\end{equation}
\begin{equation}
    I_3=0\label{I3_p1}
\end{equation}
\begin{equation}
\begin{aligned}
    I_4\gtrsim &\frac{1}{B}\sum_{b=1}^t\sum_{n\in\mathcal{B}_b}\frac{\eta^2  |\mathcal{S}_1^n|}{|\mathcal{S}^n| a}\frac{1}{2B}\sum_{n\in\mathcal{B}_b} \frac{|\mathcal{S}_1^n|m}{|\mathcal{S}^n|a}p_n(t)\|\bfp_1\|^2  \bfW_{O_{(i,\cdot)}}\bfp_1\\\label{I4_p1}
\end{aligned}
\end{equation}
\begin{equation}
\begin{aligned}
    |I_5|\lesssim &\frac{1}{B}\sum_{b=1}^t\sum_{n\in\mathcal{B}_b}\frac{\eta^2  |\mathcal{S}_1^n|}{|\mathcal{S}^n| a}  \frac{1}{2B}\sum_{n\in\mathcal{B}_b} \frac{|\mathcal{S}_2^n|m}{|\mathcal{S}^n|a}p_n(t)\|\bfp_1\|^2\bfW_{O_{(i,\cdot)}}\bfp_2\\
    &+\frac{\eta^2 tm}{\sqrt{B} a^2}\bfW_{O_{(k,\cdot)}}\bfp_1
    \label{I5_p1}
\end{aligned}
\end{equation}
Hence, combining (\ref{I1_p1}), (\ref{I2_p1}), (\ref{I3_p1}), (\ref{I4_p1}), and (\ref{I5_p1}), we can obtain
\begin{equation}
\begin{aligned}
    &\left\langle{\bfW_{O_{(i, \cdot)}}^{(t+1)}}^\top, \bfp_1\right\rangle-\left\langle{\bfW_{O_{(i, \cdot)}}^{(t)}}^\top, \bfp_1\right\rangle\\
    \gtrsim &\frac{\eta  }{a}\cdot\frac{1}{B}\sum_{n\in\mathcal{B}_b}(\frac{|\mathcal{S}_1^n|}{|\mathcal{S}^n|}p_n(t)-((\sigma+\tau))+\frac{\eta t |\mathcal{S}_1^n|}{|\mathcal{S}^n|}\frac{1}{2B}\sum_{n\in\mathcal{B}_b}\frac{|\mathcal{S}_1^n|m}{|\mathcal{S}^n|a}p_n(t)  \\
    &\cdot \bfW_{O_{(i,\cdot)}}\bfp_1(1-(\sigma+\tau))- \frac{\eta t |\mathcal{S}_1^n|}{|\mathcal{S}^n|}\frac{1}{2B}\sum_{n\in\mathcal{B}_b}\frac{|\mathcal{S}_2^n|m}{|\mathcal{S}^n|a}p_n(t)  \\
    &\cdot\bfW_{O_{(i,\cdot)}}\bfp_2(1+(\sigma+\tau))-\frac{\eta t m\bfW_{O_{(k,\cdot)}}\bfp_1}{\sqrt{B}a^2})\|\bfp_1\|^2\\
    \gtrsim &\frac{\eta  }{aB}\sum_{n\in\mathcal{B}_b}(\frac{|\mathcal{S}_1^n|}{|\mathcal{S}^n|}p_n(t)-((\sigma+\tau))+\frac{\eta t |\mathcal{S}_1^n|}{|\mathcal{S}^n|}\frac{1}{2B}\sum_{n\in\mathcal{B}_b}\frac{|\mathcal{S}_1^n|m}{|\mathcal{S}^n|a}p_n(t)\\
    &\cdot  \bfW_{O_{(i,\cdot)}}\bfp_1)\|\bfp_1\|^2\label{WO_update}
\end{aligned}
\end{equation}

where the last step holds when $B\geq \Omega(1)$. Since that $\bfW_{O_{(i,\cdot)}}^{(0)}\sim\mathcal{N}(0,\frac{\xi^2\bfI}{m_a})$, 
by the standard property of Gaussian distribution, we have
\begin{equation}
    \Pr(\|\bfW_{O_{(i,\cdot)}}^{(0)}\|\lesssim \xi)\lesssim\xi
\end{equation}
Therefore, with high probability for all $i\in[m]$, we have
\begin{equation}
    \|\bfW_{O_{(i,\cdot)}}^{(0)}\|\gtrsim \xi
\end{equation}
\begin{equation}
    \|\bfW_{O_{(i,\cdot)}}^{(0)}\bfp_1\|\gtrsim \xi
\end{equation}
When $\eta$ is very small, given $p_n(t)$ as the order of a constant,  (\ref{WO_update}) leads to a PDE on the lower bound of $\bfW_{O_{(i,\cdot)}}\bfp_1$ since the last step of (\ref{WO_update}) is always positive. Denote $y(t)$ as a lower bound of $\bfW_{O_{(i,\cdot)}}\bfp_1$, we have
\begin{equation}
\begin{aligned}
        &\frac{\partial y(t)}{\partial t}\\
        =&\Theta(\frac{1}{aB}\sum_{n\in\mathcal{B}_b}(\frac{|\mathcal{S}_1^n|}{|\mathcal{S}^n|}p_n(t)-(\sigma+\tau))+\frac{\eta t |\mathcal{S}_1^n|}{|\mathcal{S}^n|}\frac{1}{2B}\sum_{n\in\mathcal{B}_b}\frac{|\mathcal{S}_1^n|m}{|\mathcal{S}^n|a}p_n(t)   y(t))
\end{aligned}
\end{equation}
Therefore, we can derive
\begin{equation}
\begin{aligned}
   y(t)=&e^{\frac{1}{aB}\sum_{n\in\mathcal{B}_b}\frac{\eta t^2 |\mathcal{S}_1^n|}{|\mathcal{S}^n|}\frac{1}{4B}\sum_{n\in\mathcal{B}_b}\frac{|\mathcal{S}_1^n|m}{|\mathcal{S}^n|a}p_n(t)})(\int_{-\infty}^t\frac{1}{aB}\sum_{n\in\mathcal{B}_b}(\frac{|\mathcal{S}_1^n|}{|\mathcal{S}^n|}p_n(t)-(\sigma+\tau))\\
\cdot &e^{-\frac{1}{aB}\sum_{n\in\mathcal{B}_b}\frac{\eta u^2 |\mathcal{S}_1^n|}{|\mathcal{S}^n|}\frac{1}{4B}\sum_{n\in\mathcal{B}_b}\frac{|\mathcal{S}_1^n|m}{|\mathcal{S}^n|a}p_n(t)  }du+C_0)
\end{aligned}
\end{equation}
Note that
\begin{equation}
\begin{aligned}
    &\int_{-\infty}^t e^{-\frac{1}{aB}\sum_{n\in\mathcal{B}_b}\frac{\eta u^2 |\mathcal{S}_1^n|}{|\mathcal{S}^n|}\frac{1}{4B}\sum_{n\in\mathcal{B}_b}\frac{|\mathcal{S}_1^n|m}{|\mathcal{S}^n|a}p_n(t)})du\\
    \leq & \int_{-\infty}^\infty e^{-\frac{1}{aB}\sum_{n\in\mathcal{B}_b}\frac{\eta u^2 |\mathcal{S}_1^n|}{|\mathcal{S}^n|}\frac{1}{4B}\sum_{n\in\mathcal{B}_b}\frac{|\mathcal{S}_1^n|m}{|\mathcal{S}^n|a}p_n(t)})du\\
    = &\sqrt{2\pi}\cdot (\frac{1}{aB}\sum_{n\in\mathcal{B}_b}\frac{\eta |\mathcal{S}_1^n|}{|\mathcal{S}^n|}\frac{1}{4B}\sum_{n\in\mathcal{B}_b}\frac{|\mathcal{S}_1^n|m}{|\mathcal{S}^n|a}p_n(t)  )^{-1}\\
    = & \Theta(\eta^{-1})
\end{aligned}
\end{equation}
\begin{equation}
\begin{aligned}
    &\int_{-\infty}^t e^{-\frac{1}{aB}\sum_{n\in\mathcal{B}_b}\frac{\eta u^2 |\mathcal{S}_1^n|}{|\mathcal{S}^n|}\frac{1}{4B}\sum_{n\in\mathcal{B}_b}\frac{|\mathcal{S}_1^n|m}{|\mathcal{S}^n|a}p_n(t)})du\\
    \geq & \int_{-\infty}^0 e^{-\frac{1}{aB}\sum_{n\in\mathcal{B}_b}\frac{\eta u^2 |\mathcal{S}_1^n|}{|\mathcal{S}^n|}\frac{1}{4B}\sum_{n\in\mathcal{B}_b}\frac{|\mathcal{S}_1^n|m}{|\mathcal{S}^n|a}p_n(t)})du\\
    = & \Theta(\eta^{-1})
\end{aligned}
\end{equation}
Hence,
\begin{equation}
    y(0)=\frac{\eta^{-1}}{aB}\sum_{n\in\mathcal{B}_b}(\frac{|\mathcal{S}_1^n|}{|\mathcal{S}^n|}p_n(t)-(\sigma+\tau))+C_0=\Theta( \eta^{-1}\xi)+C_0=\xi
\end{equation}
\begin{equation}
    C_0=\xi(1-\Theta(\eta^{-1}))
\end{equation}
\begin{equation}
    \begin{aligned}
        \bfW_{O_{(i,\cdot)}}^{(t+1)}\bfp_1\gtrsim &y(t)\\
        \gtrsim & e^{\frac{1}{aB}\sum_{n\in\mathcal{B}_b}\frac{\eta (t+1)^2 |\mathcal{S}_1^n|}{|\mathcal{S}^n|}\frac{1}{4B}\sum_{n\in\mathcal{B}_b}\frac{|\mathcal{S}_1^n|m}{|\mathcal{S}^n|a}p_n(t)  }\xi\\
        \gtrsim &\frac{\xi}{aB}\sum_{n\in\mathcal{B}_b}\frac{\eta (t+1)^2 |\mathcal{S}_1^n|}{|\mathcal{S}^n|}\frac{1}{4B}\sum_{n\in\mathcal{B}_b}\frac{|\mathcal{S}_1^n|m}{|\mathcal{S}^n|a}p_n(t)  +\xi\label{Wp_p1_0}
    \end{aligned}
\end{equation}
\noindent (2) When $\bfp_j\in\mathcal{P}\backslash p^+$, we have
\begin{equation}
    I_2=0\label{I2_pj}
\end{equation}
\begin{equation}
    |I_3|\leq \frac{1}{B}\sum_{n\in\mathcal{B}_b} \nu_n(t) \frac{\eta|\mathcal{S}_l^n|}{a}\sqrt{\frac{\log m\log B}{B}}\|\bfp\|^2\label{I3_pj}
\end{equation}
\begin{equation}
    |I_4|\leq \frac{\eta^2  }{a}\sum_{b=1}^t\sqrt{\frac{\log m\log B}{B}}\frac{1}{2B}\sum_{n\in\mathcal{B}_b} \frac{|\mathcal{S}_1^n|m}{|\mathcal{S}^n|a}p_n(b)\bfW_{O_{(i,\cdot)}}^\top\bfp_j\|\bfp\|\label{I4_pj}
\end{equation}
\begin{equation}
    |I_5|\lesssim \frac{\eta^2 tm}{\sqrt{B}a^2}\xi\|\bfp\|^2+\frac{\eta^2}{a}\sum_{b=1}^t\sqrt{\frac{\log m\log B}{B}}\frac{1}{2B}\sum_{n\in\mathcal{B}_b} \frac{|\mathcal{S}_2^n|m}{|\mathcal{S}^n|a}p_n(t)\xi\|\bfp\|\label{I5_pj}
\end{equation}
with probability at least $1-(mB)^{-c}$. (\ref{I4_pj}) comes from (\ref{W_O_two_side_bound}). Then, combining (\ref{I1_p1}), (\ref{I2_pj}), (\ref{I3_pj}), (\ref{I4_pj}) and (\ref{I5_pj}), we can obtain
\begin{equation}
\begin{aligned}
    &\Big|\left\langle{\bfW_{O_{(i, \cdot)}}^{(t+1)}}^\top, \bfp_j\right\rangle-\left\langle{\bfW_{O_{(i, \cdot)}}^{(t)}}^\top, \bfp_j\right\rangle\Big|\\
    \lesssim &\frac{\eta  }{a}\cdot\frac{1}{B}\sum_{n\in\mathcal{B}_b} (\frac{|\mathcal{S}_l^n|}{|\mathcal{S}^n|}|\mathcal{S}_l^n|\nu_n(t)+((\sigma+\tau))\\
    &+\sum_{b=1}^t\frac{|\mathcal{S}_1^n|p_n(b) \eta m}{|\mathcal{S}^n|a}\bfW_{O_{(i,\cdot)}}^\top\bfp_j)\sqrt{\frac{\log m\log B}{B}}\|\bfp\|^2+\frac{\eta^2 t m}{\sqrt{B}a^2}\xi\|\bfp\|\\
    \lesssim &\frac{\eta  }{a}\cdot\frac{1}{B}\sum_{n\in\mathcal{B}_b} (\frac{|\mathcal{S}_l^n|}{|\mathcal{S}^n|}|\mathcal{S}_l^n|\nu_n(t)+((\sigma+\tau))\\
    &+\sum_{b=1}^t\frac{|\mathcal{S}_1^n|p_n(b) \eta m}{|\mathcal{S}^n|a}\bfW_{O_{(i,\cdot)}}^\top\bfp_j)\sqrt{\frac{\log m\log B}{B}}\|\bfp\|^2\label{WO_update_case2}
\end{aligned}
\end{equation}
Comparing (\ref{WO_update}) and (\ref{WO_update_case2}), we have
\begin{equation}
    \bfW_{O_{(i,\cdot)}}^{(t+1)}\bfp_j\lesssim \frac{1}{\sqrt{B}}\bfW_{O_{(i,\cdot)}}^{(t+1)}\bfp_1\label{Wp_pj}
\end{equation}
(3) If $i\in\mathcal{U}_{l,n}(0)$, following the derivation of (\ref{Wp_p1_0}) and (\ref{Wp_pj}), we can conclude that
\begin{equation}
        \bfW_{O_{(i,\cdot)}}^{(t+1)}\bfp_2\gtrsim \frac{\xi}{aB}\sum_{n\in\mathcal{B}_b}\frac{\eta (t+1)^2 |\mathcal{S}_1^n|}{|\mathcal{S}^n|}\frac{1}{4B}\sum_{n\in\mathcal{B}_b}\frac{|\mathcal{S}_1^n|m}{|\mathcal{S}^n|a}p_n(t)  +\xi
    \end{equation}
    \begin{equation}
        \bfW_{O_{(i,\cdot)}}^{(t)}\bfp\lesssim \frac{1}{\sqrt{B}}\bfW_{O_{(i,\cdot)}}^{(t)}\bfp_2,\ \ \ \ \text{for }\bfp\in\mathcal{P}\backslash\bfp_2,
    \end{equation}
(4) If $i\notin (\mathcal{W}_{l,n}(0)\cup\mathcal{U}_{l,n}(0))$, 
\begin{equation}
    |I_2+I_3|\leq \frac{\eta}{a}\sqrt{\frac{\log m\log B}{B}}\|\bfp\|^2\label{I23_pj_i}
\end{equation}
Following (\ref{I4_pj}) and (\ref{I5_pj}), we have
\begin{equation}
    |I_4|\leq \sum_{b=1}^t\frac{\eta^2  }{a}\sqrt{\frac{\log m\log B}{B}}\frac{1}{2B}\sum_{n\in\mathcal{B}_b} \frac{|\mathcal{S}_1^n|m}{|\mathcal{S}^n|a}p_n(b)\bfW_{O_{(i,\cdot)}}^\top\bfp_j\|\bfp\|\label{I4_pj_i}
\end{equation}
\begin{equation}
    |I_5|\lesssim \frac{\eta^2 tm}{\sqrt{B}a^2}\xi\|\bfp\|^2+\sum_{b=1}^t\frac{\eta^2  }{a}\sqrt{\frac{\log m\log B}{B}}\frac{1}{2B}\sum_{n\in\mathcal{B}_b} \frac{|\mathcal{S}_2^n|m}{|\mathcal{S}^n|a}p_n(b)\xi\|\bfp\|\label{I5_pj_i}
\end{equation}
Hence, combining (\ref{I23_pj_i}), (\ref{I4_pj_i}), and (\ref{I5_pj_i}), we can obtain
\begin{equation}
\begin{aligned}
    &\Big|\left\langle{\bfW_{O_{(i, \cdot)}}^{(t+1)}}^\top, \bfp\right\rangle-\left\langle{\bfW_{O_{(i, \cdot)}}^{(t)}}^\top, \bfp\right\rangle\Big|\\
    \lesssim& \frac{\eta  }{a}\cdot(\|\bfp\|+(\sigma+\tau)+\sum_{b=1}^t\frac{1}{2B}\sum_{n\in\mathcal{B}_b}\frac{|\mathcal{S}_1^n|p_n(b)\eta m}{|\mathcal{S}^n|a}\bfW_{O_{(i,\cdot)}}^\top\bfp_j)\sqrt{\frac{\log m\log B}{B}}\|\bfp\|\\
    &+\frac{\eta^2 t m }{\sqrt{B}a^2}\xi\|\bfp\|^2\\
    \lesssim & \frac{\eta  }{a}\cdot(\|\bfp\|+(\sigma+\tau)+\sum_{b=1}^t\frac{1}{2B}\sum_{n\in\mathcal{B}_b}\frac{|\mathcal{S}_1^n|p_n(b)\eta m}{|\mathcal{S}^n|a}\bfW_{O_{(i,\cdot)}}^\top\bfp_j)\sqrt{\frac{\log m\log B}{B}}\|\bfp\|,\label{WO_update_case4}
\end{aligned}
\end{equation}
Comparing (\ref{WO_update}) and (\ref{WO_update_case4}), we have
\begin{equation}
    \bfW_{O_{(i,\cdot)}}^{(t+1)}\bfp_j\lesssim \frac{1}{\sqrt{B}}\bfW_{O_{(j,\cdot)}}^{(t+1)}\bfp_1
\end{equation}
where $j\in\mathcal{W}_{l,n}(0)$.\\
(5) We finally study the bound of $\bfW_{O_{(i,\cdot)}}^{(t)}$ and the product with the noise term according to the analysis above. \\
\noindent By (\ref{noise_bound}), for the lucky neuron $i$, since that the update of $\bfW_{O_{(i,\cdot)}}^{(t)}$ lies in the subspace spanned by $\mathcal{P}$ and $\bfp_1,\bfp_2,\cdots,\bfp_M$ all have a unit norm, we can derive
\begin{equation}
\begin{aligned}
    \|\bfW_{O_{(i, \cdot)}}^{(t+1)}\|^2=&\sum_{l=1}^M (\bfW_{O_{(i, \cdot)}}^{(t+1)}\bfp_l)^2\geq  (\bfW_{O_{(i, \cdot)}}^{(t+1)}\bfp_1)^2\\
    \gtrsim  &(\frac{\xi}{aB}\sum_{n\in\mathcal{B}_b}\frac{\eta (t+1)^2 |\mathcal{S}_1^n|}{|\mathcal{S}^n|}\frac{1}{4B}\sum_{n\in\mathcal{B}_b}\frac{|\mathcal{S}_1^n|m}{|\mathcal{S}^n|a}p_n(t)  )^2
\end{aligned}
\end{equation}
\begin{equation}
\begin{aligned}
    \|\bfW_{O_{(i, \cdot)}}^{(t+1)}\bfz_l(t)\|\leq& \Big|(\sigma+\tau)\|\bfW_{O_{(i, \cdot)}}^{(t+1)}\|\Big|\\
\end{aligned}
\end{equation}
For the unlucky neuron $i$, we can similarly obtain
\begin{equation}
    \begin{aligned}
        \|\bfW_{O_{(i, \cdot)}}^{(t+1)}\|^2\leq& \frac{1}{B}\|\bfW_{O_{(j, \cdot)}}^{(t+1)}\|^2
    \end{aligned}
\end{equation}
where $j$ is a lucky neuron. The proof of Claim \ref{clm: W_O} ends here.

\subsection{\textbf{Proof of Claim \ref{clm: W_QK} of Lemma \ref{lm: training}}}
The proof of Claim \ref{clm: W_QK} is one of the most challenging parts in our paper, since that we need to deal with the complicated softmax function. The core idea of proof is that we pay more attention on the changes of label-relevant features in the gradient update, which should be the most crucial factor based on our data model. We then show the attention map converges to be sparse as long as the data model satisfies (\ref{eqn: alpha*}).

\noindent We first study the gradient of $\bfW_Q^{(t+1)}$ in part (a) and the gradient of $\bfW_K^{(t+1)}$ in part (b).\\
\noindent (a) By (\ref{loss_data}), we have
\begin{equation}
\begin{aligned}
    &\eta\frac{1}{B}\sum_{n\in\mathcal{B}_b}\frac{\partial \textbf{Loss}(\bfX^n,y^n)}{\partial \bfW_Q}\\
    =&\eta\frac{1}{B}\sum_{n\in\mathcal{B}_b} \frac{\partial \textbf{Loss}(\bfX^n,y^n)}{\partial F(\bfX^n)}\frac{F(\bfX^n)}{\partial \bfW_Q}\\
    =&\eta\frac{1}{B}\sum_{n\in\mathcal{B}_b}(-y^n)\frac{1}{|\mathcal{S}^n|}\sum_{l\in\mathcal{S}^n} \sum_{i=1}^m a_{(l)_i}\mathbbm{1}[\bfW_{O_{(i,\cdot)}}\bfW_V\bfX\text{softmax}({\bfX^n}^\top\bfW_K^\top\bfW_Q\bfx_l^n)\geq0]\\
    &\cdot\Big(\bfW_{O_{(i,\cdot)}}\sum_{s\in\mathcal{S}^n} \bfW_V\bfx_s^n\text{softmax}({\bfx_s^n}^\top\bfW_K^\top\bfW_Q\bfx_l^n)\\
    &\cdot\sum_{r\in\mathcal{S}^n} \text{softmax}({\bfx_r^n}^\top\bfW_K^\top\bfW_Q\bfx_l^n)\bfW_K(\bfx_s^n-\bfx_r^n){\bfx_l^n}^\top\Big)\\
    =&\eta\frac{1}{B}\sum_{n\in\mathcal{B}_b}(-y^n)\frac{1}{|\mathcal{S}^n|}\sum_{l\in\mathcal{S}^n} \sum_{i=1}^m a_{(l)_i}\mathbbm{1}[\bfW_{O_{(i,\cdot)}}\bfW_V\bfX^n\text{softmax}({\bfX^n}^\top\bfW_K^\top\bfW_Q\bfx_l^n)\geq0]\\
    &\cdot\Big(\bfW_{O_{(i,\cdot)}}\sum_{s\in\mathcal{S}^n} \bfW_V\bfx_s^n\text{softmax}({\bfx_s^n}^\top\bfW_K^\top\bfW_Q\bfx_l^n)\\
    &\cdot(\bfW_K\bfx_s^n-\sum_{r\in\mathcal{S}^n} \text{softmax}({\bfx_r^n}^\top\bfW_K^\top\bfW_Q\bfx_l^n)\bfW_K\bfx_r^n){\bfx_l^n}^\top\Big)\label{grad_Wk}
\end{aligned}
\end{equation}
For $r,l\in\mathcal{S}_1^n$, by (\ref{softmax_t1_same}) we have
    \begin{equation}
    \text{softmax}({\bfx_j^n}^\top\bfW_K^{(t)}\bfW_Q^{(t)}\bfx_l^n)\gtrsim \frac{e^{\|\bfq_1(t)\|^2-(\delta+\tau)\|\bfq_1(t)\|}}{|\mathcal{S}_1^n|e^{\|\bfq_1(t)\|^2-(\delta+\tau)\|\bfq_1(t)\|}+(|\mathcal{S}^n|-|\mathcal{S}_1^n|)}
    \end{equation}
For $r\notin\mathcal{S}_1^n$ and $l\in\mathcal{S}_1^n$, we have
    \begin{equation}
    \text{softmax}({\bfx_j^n}^\top{\bfW_K^{(t+1)}}^\top\bfW_Q^{(t+1)}\bfx_l^n)\lesssim \frac{1}{|\mathcal{S}_1^n|e^{(1+K(t))\|\bfq_1(t)\|^2-(\delta+\tau)\|\bfq_1(t)\|}+(|\mathcal{S}^n|-|\mathcal{S}_1^n|)}
    \end{equation}
Therefore, for $s,r,l\in\mathcal{S}_1^n$, let
\begin{equation}
    \bfW_K^{(t)}\bfx_s^n-\sum_{r\in\mathcal{S}^n} \text{softmax}({\bfx_r^n}^\top{\bfW_K^{(t)}}^\top\bfW_Q^{(t)}\bfx_l^n)\bfW_K^{(t)}\bfx_r^n:=\beta_1^n(t)\bfq_1(t)+\boldsymbol{\beta}_2^n(t),
\end{equation}
where
\begin{equation}
\begin{aligned}
    \beta_1^n(t)&\gtrsim \frac{|\mathcal{S}^n|-|\mathcal{S}_1^n|}{|\mathcal{S}_1^n|e^{\|\bfq_1(t)\|^2+(\delta+\tau)\|\bfq_1(t)\|}+|\mathcal{S}^n|-|\mathcal{S}_1^n|}\\
    &:=\phi_n(t)(|\mathcal{S}^n|-|\mathcal{S}_1^n|).\label{beta1}
\end{aligned}
\end{equation}
\begin{equation}
    \beta_1^n(t)\lesssim \nu_n(t)(|\mathcal{S}^n|-|\mathcal{S}_1^n|)\lesssim e^{2(\tau+\delta)\|\bfq_1(t)\|}\phi_n(t)(|\mathcal{S}^n|-|\mathcal{S}_1^n|)\leq \phi_n(t)(|\mathcal{S}^n|-|\mathcal{S}_1^n|)\label{beta_1_upper}
\end{equation}
where the last inequality holds when the final iteration $\log T\leq \Theta(1)$. 
\begin{equation}
    \begin{aligned}
    \boldsymbol{\beta}_2^n(t)\approx &\Theta(1)\cdot \bfo_j^n(t)+Q_e(t)\bfr_2(t)+\sum_{l=3}^M  \gamma_l' \bfr_l(t)-\sum_{a=1}^M\sum_{r\in \mathcal{S}_l^n}\text{softmax}(\bfx_r^\top{\bfW_K^{(t)}}^\top\bfW_Q^{(t)}\bfx_l)\bfr_{a}(t)\\
    =& \Theta(1)\cdot \bfo_j^n(t)+\sum_{l=1}^M \zeta_l'\bfr_l(t)
    \end{aligned}
\end{equation}
for some $Q_e(t)>0$ and $\gamma'_l>0$. 
Here
\begin{equation}
    |\zeta_l'|\leq \beta_1^n(t)\frac{|\mathcal{S}_l^n|}{|\mathcal{S}^n|-|\mathcal{S}_1^n|}
\end{equation}
for $l\geq 2$. Note that $|\zeta_l'|=0$ if $|\mathcal{S}^n|=|\mathcal{S}_1^n|$, $l\geq2$.\\
For $i\in\mathcal{W}_{l,n}(0)$, by Lemma \ref{lemma: update_WU},
\noindent Then we study how large the coefficient of $\bfq_1(t)$ in (\ref{grad_Wk}).\\
\noindent If $s\in\mathcal{S}_1^n$, by basic computation given (\ref{W_O_p_1}) to (\ref{W_O_noise}),
\begin{equation}
\begin{aligned}
    &\bfW_{O_{(i,\cdot)}}^{(t)} \bfW_V^{(t)}\bfx_s^n\text{softmax}({\bfx_s^n}^\top{\bfW_K^{(t)}}^\top\bfW_Q^{(t)}\bfx_l^n)\\
    \gtrsim &\frac{p_n(t)}{|\mathcal{S}_1^n|}\Big(\frac{1}{B}\sum_{n\in\mathcal{B}_b}\frac{\xi\eta (t+1)^2 |\mathcal{S}_1^n|m}{|\mathcal{S}^n|a^2}(\frac{1}{4B}\sum_{n\in\mathcal{B}_b}\frac{|\mathcal{S}_1^n|}{|\mathcal{S}^n|}p_n(b)-\sigma-\tau)\\
    &\cdot  + \eta  m  \frac{1}{2B}\sum_{b=1}^t\sum_{n\in{\mathcal{B}_b}_+}\frac{|\mathcal{S}_1^n|}{a|\mathcal{S}^n|}p_n(b)\\
    &\cdot (1-(\sigma+\tau))\cdot(\frac{\xi}{aB}\sum_{n\in\mathcal{B}_b}\frac{\eta (t+1)^2 |\mathcal{S}_1^n|}{|\mathcal{S}^n|}\frac{1}{4B}\sum_{n\in\mathcal{B}_b}\frac{|\mathcal{S}_1^n|m}{|\mathcal{S}^n|a}p_n(t)  )^2\Big)\label{s_Omega_1}
\end{aligned}
\end{equation}
\noindent If $s\in\mathcal{S}_2^n$ and $j\in\mathcal{S}_1^n$, from (\ref{W_O_p_2}) to (\ref{W_O_noise2}), we have
\begin{equation}
    \begin{aligned}
    &\bfW_{O_{(i,\cdot)}}^{(t)} \bfW_V^{(t)}{\bfx_s^n}\text{softmax}({\bfx_s^n}^\top{{\bfW_K}^{(t)}}^\top\bfW_Q^{(t)}\bfx_l^n)\\
    \lesssim&\bfW_{O_{(i,\cdot)}}^{(t)} \bfW_V^{(t)}\bfx_j^n\text{softmax}({\bfx_j^n}^\top{\bfW_K^{(t)}}^\top\bfW_Q^{(t)}\bfx_l^n)\phi_n(t)\cdot \frac{|\mathcal{S}_1^n|}{p_n(t)}\label{s_Omega_2}
    \end{aligned}
\end{equation}
If $i\in\mathcal{W}_{l,n}(0)$, $s\notin(\mathcal{S}_1^n\cup\mathcal{S}_2^n)$ and $j\in\mathcal{S}_1^n$, 
\begin{equation}
    \begin{aligned}
    &\bfW_{O_{(i,\cdot)}}^{(t)} \bfW_V^{(t)}\bfx_s^n\text{softmax}({\bfx_s^n}^\top{{\bfW_K}^{(t)}}^\top\bfW_Q^{(t)}\bfx_l^n)\\
    \lesssim&\bfW_{O_{(i,\cdot)}}^{(t)} \bfW_V^{(t)}\bfx_j^n\text{softmax}({\bfx_j^n}^\top{\bfW_K^{(t)}}^\top\bfW_Q^{(t)}\bfx_l^n)\phi_n(t)\cdot \frac{|\mathcal{S}_1^n|}{\sqrt{B}p_n(t)}\label{s_Omega_other}
    \end{aligned}
\end{equation}
by (\ref{W_O_p_3}) to (\ref{W_O_noise_3}).\\
\noindent Hence, for $i\in\mathcal{W}_{l,n}(0)$, $j\in\mathcal{S}_1^g$, combining (\ref{beta1}) and (\ref{s_Omega_1}), we have
\begin{equation}
    \begin{aligned}
    &\bfW_{O_{(i,\cdot)}}^{(t)} \sum_{s\in\mathcal{S}^n}\bfW_V^{(t)}\bfx_s^n\text{softmax}({\bfx_s^n}^\top{{\bfW_K}^{(t)}}^\top\bfW_Q^{(t)}\bfx_l^n)\bfq_1(t)^\top\\
    \cdot&(\bfW_K^{(t)}\bfx_s^n-\sum_{r=1}^L \text{softmax}({\bfx_r^n}^\top{\bfW_K^{(t)}}^\top\bfW_Q^{(t)}\bfx_l^n)\bfW_K^{(t)}\bfx_r^n){\bfx_l^n}^\top\bfx_j^g\\
    \gtrsim& p_n(t)\Big(\frac{1}{B}\sum_{n\in\mathcal{B}_b}\frac{\xi\eta (t+1)^2 |\mathcal{S}_1^n|m}{|\mathcal{S}^n|a^2}(\frac{1}{4B}\sum_{n\in\mathcal{B}_b}\frac{|\mathcal{S}_1^n|}{|\mathcal{S}^n|}p_n(b)-\sigma-\tau)\\
    &\cdot  + \eta  m  \frac{1}{2B}\sum_{b=1}^t\sum_{n\in{\mathcal{B}_b}_+}\frac{|\mathcal{S}_1^n|}{a|\mathcal{S}^n|}p_n(b)\\
    &\cdot (\frac{\xi}{aB}\sum_{n\in\mathcal{B}_b}\frac{\eta (t+1)^2 |\mathcal{S}_1^n|}{|\mathcal{S}^n|}\frac{1}{4B}\sum_{n\in\mathcal{B}_b}\frac{|\mathcal{S}_1^n|m}{|\mathcal{S}^n|a}p_n(t)  )^2\Big)\\
    &\cdot\phi_n(t)(|\mathcal{S}^n|-|\mathcal{S}_1^n|)\|\bfq_1(t)\|^2\label{i_U(t)_Omega1}
    \end{aligned}
\end{equation}
The following upper bounds use the lower bound of (\ref{i_U(t)_Omega1}) since further results will be the gap of these terms. For $i\in\mathcal{U}_{l,n}(0)$ and $l\in\mathcal{S}_1^n$, $j\in\mathcal{S}_1^g$, and $k\in\mathcal{W}_{l,n}(0)$,
\begin{equation}
    \begin{aligned}
    &\bfW_{O_{(i,\cdot)}}^{(t)} \sum_{s\in\mathcal{S}^n}\bfW_V^{(t)}\bfx_s^n\text{softmax}({\bfx_s^n}^\top{{\bfW_K^{(t)}}^\top}\bfW_Q^{(t)}\bfx_l^n)\bfq_1(t)^\top\\
    \cdot&(\bfW_K^{(t)}\bfx_s^n-\sum_{r\in\mathcal{S}^n} \text{softmax}({\bfx_r^n}^\top{\bfW_K^{(t)}}^\top\bfW_Q^{(t)}\bfx_l^n)\bfW_K^{(t)}\bfx_r^n){\bfx_l^n}^\top\bfx_j^g\\
    \lesssim& p_n(t)\Big(\frac{1}{B}\sum_{n\in\mathcal{B}_b}\frac{\xi\eta (t+1)^2 |\mathcal{S}_2^n|m}{|\mathcal{S}^n|a^2}(\frac{1}{4B}\sum_{n\in\mathcal{B}_b}\frac{|\mathcal{S}_2^n|}{|\mathcal{S}^n|}p_n(b)-\sigma-\tau)\\
    &\cdot  + \eta  m  \frac{1}{2B}\sum_{b=1}^t\sum_{n\in{\mathcal{B}_b}_+}\frac{|\mathcal{S}_2^n|}{a|\mathcal{S}^n|}p_n(b)\\
    &\cdot \cdot(\frac{\xi}{aB}\sum_{n\in\mathcal{B}_b}\frac{\eta (t+1)^2 |\mathcal{S}_2^n|}{|\mathcal{S}^n|}\frac{1}{4B}\sum_{n\in\mathcal{B}_b}\frac{|\mathcal{S}_2^n|m}{|\mathcal{S}^n|a}p_n(t)  )^2\Big)\\
    &\cdot\phi_n(t)|\mathcal{S}_2^n|\beta_1(t)\|\bfq_1(t)\|^2
    \end{aligned}
\end{equation}
For $i\notin(\mathcal{W}_{l,n}(0)\cup\mathcal{U}_{l,n}(0))$ and $l\in\mathcal{S}_1^n$, $j\in\mathcal{S}_1^g$, and $k\in\mathcal{W}_{l,n}(0)$,
\begin{equation}
    \begin{aligned}
    &\bfW_{O_{(i,\cdot)}}^{(t)} \sum_{s\in\mathcal{S}^n}\bfW_V^{(t)}\bfx_s^n\text{softmax}({\bfx_s^n}^\top{{\bfW_K}^{(t)}}^\top\bfW_Q^{(t)}\bfx_l^n){\bfq_1(t)}^\top\\
    \cdot&(\bfW_K^{(t)}\bfx_s^n-\sum_{r\in\mathcal{S}^n} \text{softmax}({\bfx_r^n}^\top{\bfW_K^{(t)}}^\top\bfW_Q^{(t)}\bfx_l^n)\bfx_r^n){\bfx_l^n}^\top\bfx_j^g\\
    \lesssim& \bfW_{O_{(k,\cdot)}}^{(t)} \sum_{s\in\mathcal{S}^n}\bfW_V^{(t)}\bfx_s^n\text{softmax}({\bfx_s^n}^\top{{\bfW_K}^{(t)}}^\top\bfW_Q^{(t)}\bfx_l^n){\bfq_1(t)}^\top\\
    &\cdot(\bfW_K^{(t)}\bfx_s^n-\sum_{r\in\mathcal{S}^n} \text{softmax}({\bfx_r^n}^\top{\bfW_K^{(t)}}^\top\bfW_Q^{(t)}\bfx_l^n)\bfW_K^{(t)}\bfx_r^n){\bfx_l^n}^\top\bfx_j^g\cdot \frac{1}{\sqrt{B}}
    \end{aligned}
\end{equation}
To study the case when $l\notin\mathcal{S}_1^n$ for all $n\in[N]$, we need to check all other $l$'s. Recall that we focus on the coefficient of $\bfq_1(t)$ in this part. Based on the computation in (\ref{s_Omega_2}) and (\ref{s_Omega_other}), we know that the contribution of coefficient from non-discriminative patches is no more than that from discriminative patches, i.e., for $l\notin(\mathcal{S}_1^n\cup\mathcal{S}_2^n),\ n\in[N]$ and $k\in\mathcal{S}_1^n$,
\begin{equation}
    \begin{aligned}
        &\Big|\bfW_{O_{(i,\cdot)}}^{(t)} \sum_{s\in\mathcal{S}^n}\bfW_V^{(t)}\bfx_s^n\text{softmax}({\bfx_s^n}^\top{\bfW_K^{(t)}}^\top\bfW_Q^{(t)}\bfx_l^n)\bfq_1(t)^\top\\
        \cdot&(\bfW_K^{(t)}\bfx_s^n-\sum_{r\in\mathcal{S}^n} \text{softmax}({\bfW_K^{(t)}\bfx_r^n}^\top{\bfW_K^{(t)}}^\top\bfW_Q^{(t)}\bfx_l^n)\bfW_K^{(t)}\bfx_r^n){\bfx_l^n}^\top\bfx_j^g|\Big|\\
        \lesssim &\Big|\bfW_{O_{(i,\cdot)}}^{(t)} \sum_{s\in\mathcal{S}^n}\bfW_V^{(t)}\bfx_s^n\text{softmax}({\bfx_s^n}^\top{\bfW_K^{(t)}}^\top\bfW_Q^{(t)}\bfx_k^n)\bfq_1(t)^\top\\
        \cdot&(\bfW_K^{(t)}\bfx_s^n-\sum_{r\in\mathcal{S}^n} \text{softmax}({\bfW_K^{(t)}\bfx_r^n}^\top{\bfW_K^{(t)}}^\top\bfW_Q^{(t)}\bfx_l^n)\bfW_K^{(t)}\bfx_r^n){\bfx_k^n}^\top\bfx_j^g\Big|
    \end{aligned}
\end{equation}
Similar to (\ref{i_U(t)_Omega1}), we have that for $l\in\mathcal{S}_2^n$, $j\in\mathcal{S}_1^g$, and $i\in\mathcal{U}_{l,n}(0)$,
\begin{equation}
    \begin{aligned}
        &\bfW_{O_{(i,\cdot)}}^{(t)} \sum_{s\in\mathcal{S}^n}\bfW_V^{(t)}\bfx_s^n\text{softmax}({\bfx_s^n}^\top{\bfW_K^{(t)}}^\top\bfW_Q^{(t)}\bfx_l^n)\bfq_1(t)^\top\\
        \cdot&(\bfW_K^{(t)}\bfx_s^n-\sum_{r\in\mathcal{S}^n} \text{softmax}({\bfW_K^{(t)}\bfx_r^n}^\top{\bfW_K^{(t)}}^\top\bfW_Q^{(t)}\bfx_l^n)\bfW_K^{(t)}\bfx_r^n){\bfx_l^n}^\top\bfx_j^g\\
    \lesssim& \Big(\frac{1}{B}\sum_{n\in\mathcal{B}_b}\frac{\xi\eta (t+1)^2 |\mathcal{S}_2^n|m}{|\mathcal{S}^n|a^2}(\frac{1}{4B}\sum_{n\in\mathcal{B}_b}\frac{|\mathcal{S}_2^n|}{|\mathcal{S}^n|}p_n(b))\\
    &\cdot  + \eta  m  \frac{1}{2B}\sum_{b=1}^t\sum_{n\in{\mathcal{B}_b}_+}\frac{|\mathcal{S}_2^n|}{a|\mathcal{S}^n|}p_n(b)\\
    &\cdot (\frac{\xi}{aB}\sum_{n\in\mathcal{B}_b}\frac{\eta (t+1)^2 |\mathcal{S}_2^n|}{|\mathcal{S}^n|}\frac{1}{4B}\sum_{n\in\mathcal{B}_b}\frac{|\mathcal{S}_2^n|m}{|\mathcal{S}^n|a}p_n(t)  )^2\Big)\\
    &\cdot\beta_1(t)\|\bfq_1(t)\|^2\lambda\frac{|\mathcal{S}_\#^n|}{|\mathcal{S}^n|-|\mathcal{S}_*^n|}\\
    \end{aligned}
\end{equation}
Therefore, by the update rule, 
\begin{equation}
\begin{aligned}
    \bfW_Q^{(t+1)}\bfx_j^n&=\bfW_Q^{(t)}\bfx_j^n-\eta\Big(\frac{\partial L}{\partial \bfW_Q}\Big|\bfW_Q^{(t)}\Big)\bfx_j^n\\
    &=\bfr_1(t)+K(t)\bfq_1(t)+\Theta(1)\cdot \bfn_j(t)+\sum_{b=0}^{t-1}|K_e(b)|\bfq_2(b)+\sum_{l=3}^M \gamma_l'\bfq_l(t)\\
    &= (1+K(t))\bfq_1(t)+\Theta(1)\cdot \bfn_j(t)+\sum_{b=0}^{t-1}|K_e(b)|\bfq_2(b)+\sum_{l=3}^M \gamma_l'\bfq_l(t)\label{W_K_t1}
\end{aligned}
\end{equation}
where the last step is by the condition that 
\begin{equation}
    \bfq_1(t)=k_1(t)\cdot\bfr_1(t),
\end{equation}
and
\begin{equation}
    \bfq_2(t)=k_2(t)\cdot\bfr_2(t)
\end{equation}
for $k_1(t)>0$ and $k_2(t)>0$ from induction, i.e., $\bfq_1(t)$ and $\bfr_1(t)$, $\bfq_1(t)$ and $\bfr_1(t)$ are in the same direction, respectively. 
Define $qc_t(\bfx)=\bfx^\top\bfq_1(t)/\|\bfq_1(t)\|$ and denote
\begin{equation}
\begin{aligned}
    \Delta(l,i)=&a_{(l)_i}\mathbbm{1}[\bfW_{O_{(i,\cdot)}}\bfW_V\bfX^n\text{softmax}({\bfX^n}^\top\bfW_K^\top\bfW_Q\bfx_l^n)\geq0]\\
    &\cdot\Big(\bfW_{O_{(i,\cdot)}}\sum_{s\in\mathcal{S}^n} \bfW_V\bfx_s^n\text{softmax}({\bfx_s^n}^\top\bfW_K^\top\bfW_Q\bfx_l^n)\\
    &\cdot(\bfW_K\bfx_s^n-\sum_{r\in\mathcal{S}^n} \text{softmax}({\bfx_r^n}^\top\bfW_K^\top\bfW_Q\bfx_l^n)\bfW_K\bfx_r^n){\bfx_l^n}^\top\Big)
\end{aligned}
\end{equation}
We also have
\begin{equation}
    \begin{aligned}
    &K(t)\\
    \gtrsim & \eta\frac{1}{B}\Big(\Big|\sum_{n\in\mathcal{B}_b}(-y^n)\frac{1}{|\mathcal{S}^n|}\sum_{l\in\mathcal{S}_1^n}\sum_{i\in\mathcal{W}_{l,n}(0)}qc_t(\Delta(l,i))\Big|-\Big|\sum_{n\in\mathcal{B}_b}(-y^n)\frac{1}{|\mathcal{S}^n|}\sum_{l\in\mathcal{S}_1^n}\sum_{i\in\mathcal{U}_{l,n}(0)}qc_t(\Delta(l,i))\Big|\\
    &-\Big|\sum_{n\in\mathcal{B}_b}(-y^n)\frac{1}{|\mathcal{S}^n|}\sum_{l\in\mathcal{S}_1^n}\sum_{i\notin\mathcal{W}_{l,n}(0)\cup\mathcal{U}_{l,n}(0)}qc_t(\Delta(l,i))\Big|-\Big|\sum_{n\in\mathcal{B}_b}(-y^n)\frac{1}{|\mathcal{S}^n|}\sum_{l\in\mathcal{S}_2^n}\sum_{i=1}^m qc_t(\Delta(l,i))\Big|\\
    &-\Big|\sum_{n\in\mathcal{B}_b}(-y^n)\frac{1}{|\mathcal{S}^n|}\sum_{l\in\mathcal{S}^n-\mathcal{S}_1^n-\mathcal{S}_2^n}\sum_{i=1}^m qc_t(\Delta(l,i))\Big|\Big)\\
        \gtrsim & \eta\frac{1}{B}\sum_{n\in\mathcal{B}_b}\frac{|\mathcal{S}_1^n|m}{|\mathcal{S}^n|a}p_n(t)\Big(\frac{1}{B}\sum_{n\in\mathcal{B}_b}\frac{\xi\eta (t+1)^2 |\mathcal{S}_1^n|m}{|\mathcal{S}^n|a^2}(\frac{1}{4B}\sum_{n\in\mathcal{B}_b}\frac{|\mathcal{S}_1^n|}{|\mathcal{S}^n|}p_n(b)-\sigma-\tau)\\
    &\cdot  + \eta  m  \frac{1}{2B}\sum_{b=1}^t\sum_{n\in{\mathcal{B}_b}_+}\frac{|\mathcal{S}_1^n|}{a|\mathcal{S}^n|}p_n(b)\\
    &\cdot (\frac{\xi}{aB}\sum_{n\in\mathcal{B}_b}\frac{\eta (t+1)^2 |\mathcal{S}_1^n|}{|\mathcal{S}^n|}\frac{1}{4B}\sum_{n\in\mathcal{B}_b}\frac{|\mathcal{S}_1^n|m}{|\mathcal{S}^n|a}p_n(t)  )^2\Big)\\
    &\cdot\phi_n(t)(|\mathcal{S}^n|-|\mathcal{S}_1^n|)\|\bfq_1(t)\|^2
    \end{aligned}
\end{equation}

\begin{equation}
    |\gamma_l'|\lesssim \frac{1}{B}\sum_{n\in\mathcal{B}_b} K(t)\cdot\frac{|\mathcal{S}_l^n|}{|\mathcal{S}^n|-|\mathcal{S}_1^n|}\label{gamma_2}
\end{equation}
\begin{equation}
    |K_e(t)|\lesssim \frac{1}{B}\sum_{n\in\mathcal{B}_b} K(t)\cdot\frac{|\mathcal{S}_2^n|}{|\mathcal{S}^n|-|\mathcal{S}_1^n|}\label{K_e}
\end{equation}
as long as 
\begin{equation}
    \begin{aligned}
        &\eta\frac{1}{B}\sum_{n\in\mathcal{B}_b}\frac{|\mathcal{S}_1^n|m}{|\mathcal{S}^n|a}p_n(t)\Big(\frac{1}{B}\sum_{n\in\mathcal{B}_b}\frac{\xi\eta (t+1)^2 |\mathcal{S}_1^n|m}{|\mathcal{S}^n|a^2}(\frac{1}{4B}\sum_{n\in\mathcal{B}_b}\frac{|\mathcal{S}_1^n|}{|\mathcal{S}^n|}p_n(b)-\sigma-\tau)\\
    &\cdot  + \eta  m  \frac{1}{2B}\sum_{b=1}^t\sum_{n\in{\mathcal{B}_b}_+}\frac{|\mathcal{S}_1^n|}{a|\mathcal{S}^n|}p_n(b)\\
    &\cdot(\frac{\xi}{aB}\sum_{n\in\mathcal{B}_b}\frac{\eta (t+1)^2 |\mathcal{S}_1^n|}{|\mathcal{S}^n|}\frac{1}{4B}\sum_{n\in\mathcal{B}_b}\frac{|\mathcal{S}_1^n|m}{|\mathcal{S}^n|a}p_n(t)  )^2\Big)\\
    &\cdot\phi_n(t)(|\mathcal{S}^n|-|\mathcal{S}_1^n|)\|\bfq_1(t)\|^2\\
    \gtrsim & \eta\frac{1}{B}\sum_{n\in\mathcal{B}_b}\frac{|\mathcal{S}_2^n|m}{|\mathcal{S}^n|a}\Big(\frac{1}{B}\sum_{n\in\mathcal{B}_b}\frac{\xi\eta (t+1)^2 |\mathcal{S}_2^n|m}{|\mathcal{S}^n|a^2}(\frac{1}{4B}\sum_{n\in\mathcal{B}_b}\frac{|\mathcal{S}_2^n|}{|\mathcal{S}^n|}p_n(b))\\
    &\cdot  + \eta  m  \frac{1}{2B}\sum_{b=1}^t\sum_{n\in{\mathcal{B}_b}_+}\frac{|\mathcal{S}_2^n|}{a|\mathcal{S}^n|}p_n(b)\\
    &\cdot(\frac{\xi}{aB}\sum_{n\in\mathcal{B}_b}\frac{\eta (t+1)^2 |\mathcal{S}_2^n|}{|\mathcal{S}^n|}\frac{1}{4B}\sum_{n\in\mathcal{B}_b}\frac{|\mathcal{S}_2^n|m}{|\mathcal{S}^n|a}p_n(t)  )^2\Big)\\
    &\cdot\beta_1(t)\|\bfq_1(t)\|^2\lambda\frac{|\mathcal{S}_\#^n|}{|\mathcal{S}^n|-|\mathcal{S}_*^n|}\label{alpha_condition_1_ini}
    \end{aligned}
\end{equation}
To find the sufficient condition for (\ref{alpha_condition_1_ini}), we mainly need to compare $\frac{1}{B}\sum_{n\in\mathcal{B}_b}\frac{|\mathcal{S}_1^n|}{|\mathcal{S}^n|}p_n(t)\phi_n(t)(|\mathcal{S}^n|-|\mathcal{S}_1^n|)$ and $\frac{1}{B}\sum_{n\in\mathcal{B}_b}\frac{|\mathcal{S}_2^n|}{|\mathcal{S}^n|}\beta_1(t)\lambda\frac{|\mathcal{S}_\#^n|}{|\mathcal{S}^n|-|\mathcal{S}_*^n|}$.\\
When $|\mathcal{S}^n|>|\mathcal{S}_1^n|$, by (\ref{beta_1_upper}), 
\begin{equation}
    \phi_n(t)(|\mathcal{S}^n|-|\mathcal{S}_1^n|)\gtrsim \beta_1^n(t)
\end{equation}
From Definition \ref{def: terms}, we know
\begin{equation} 
\begin{aligned}
    1&\geq p_n(t)\geq p_n(0)=\Theta(\frac{|\mathcal{S}_1^n|e^{-(\delta+\tau)}}{|\mathcal{S}_1^n| e^{-(\delta+\tau)}+|\mathcal{S}^n|-|\mathcal{S}_1^n|})\geq \Theta(e^{-(\delta+\tau)})
\end{aligned}
\end{equation}
Meanwhile, by Hoeffding's inequality (\ref{hoeffding}),
\begin{equation}
\begin{aligned}
    &\Big|\frac{1}{B}\sum_{n\in\mathcal{B}_b}(\frac{|\mathcal{S}_1^n|}{|\mathcal{S}^n|}-\sigma-\tau)-(1-\sigma-\tau)\mathbb{E}[\frac{|\mathcal{S}_1^n|}{|\mathcal{S}^n|}]\Big|\\
    \leq & \Big|\frac{1}{B}\sum_{n\in\mathcal{B}_b}(\frac{|\mathcal{S}_1^n|}{|\mathcal{S}^n|}-\sigma-\tau)-\mathbb{E}[\frac{|\mathcal{S}_1^n|}{|\mathcal{S}^n|}]+\sigma+\tau\Big|+\Big|(\sigma+\tau)(1-\mathbb{E}[\frac{|\mathcal{S}_1^n|}{|\mathcal{S}^n|}])\Big|\\
    \leq & \sqrt{\frac{\log B}{B}}+\sigma+\tau
\end{aligned}
\end{equation}
Therefore, a sufficient function for (\ref{alpha_condition_1_ini}) is that
\begin{equation}
    e^{-(\delta+\tau)}(1-\sigma-\tau)(\alpha_*-\sqrt{\frac{\log B}{B}}-\sigma-\tau)\geq \sqrt{\frac{\log B}{B}}+\alpha_\#
\end{equation}
Hence, 
\begin{equation}
    \alpha_*\geq \frac{1-\alpha_{nd}}{1+(1-(\tau+\sigma))e^{-(\delta+\tau)}}\label{alpha_proof}
\end{equation}
if
\begin{equation}
    B\geq \Omega(1)
\end{equation}

Then we give a brief derivation of $\bfW_Q^{(t+1)}\bfx_j^n$ for $j\notin\mathcal{S}_1^n$ in the following.\\
To be specific, for $j\in\mathcal{S}_n/(\mathcal{S}_1^n\cup\mathcal{S}_2^n)$,
\begin{equation}
    \begin{aligned}
    &\left\langle\eta\frac{1}{B}\sum_{n\in\mathcal{B}_b}\frac{\partial \textbf{Loss}(\bfX^n,y^n)}{\partial \bfW_Q^{(t)}}\bfx_j^n, \bfq_1(t)\right\rangle\\
    \gtrsim &\eta\frac{1}{B}\sum_{n\in\mathcal{B}_b}\frac{|\mathcal{S}_1^n|m}{|\mathcal{S}^n|a}p_n'(t)\Big(\frac{1}{B}\sum_{n\in\mathcal{B}_b}\frac{\xi\eta (t+1)^2 |\mathcal{S}_1^n|m}{|\mathcal{S}^n|a^2}(\frac{1}{4B}\sum_{n\in\mathcal{B}_b}\frac{|\mathcal{S}_1^n|}{|\mathcal{S}^n|}p_n(b)-\sigma-\tau)\\
    &\cdot  + \eta  m  \frac{1}{2B}\sum_{b=1}^t\sum_{n\in{\mathcal{B}_b}_+}\frac{|\mathcal{S}_1^n|}{a|\mathcal{S}^n|}p_n(b)\\
    &\cdot (\frac{\xi}{aB}\sum_{n\in\mathcal{B}_b}\frac{\eta (t+1)^2 |\mathcal{S}_1^n|}{|\mathcal{S}^n|}\frac{1}{4B}\sum_{n\in\mathcal{B}_b}\frac{|\mathcal{S}_1^n|m}{|\mathcal{S}^n|a}p_n(t)  )^2\Big)\\
    &\cdot\phi_n(t)(|\mathcal{S}^n|-|\mathcal{S}_1^n|)\|\bfq_1(t)\|^2\label{other_q_q1}
    \end{aligned}
\end{equation}
where 
\begin{equation}
    p_n'(t)=\frac{|\mathcal{S}_1^n|e^{\bfq_1(t)^\top\sum_{b=1}^t K(b)\bfq_1(0)-(\delta+\tau)]\|\bfq_1(t)\|}}{|\mathcal{S}_1^n|e^{\bfq_1(t)^\top\sum_{b=1}^t K(b)\bfq_1(b)-(\delta+\tau)]\|\bfq_1(t)\|}+|\mathcal{S}^n-|\mathcal{S}_1^n|}
\end{equation}
When $K(b)$ is close to $0^+$, we have
\begin{equation}
    \prod_{b=1}^t \sqrt{1+K(b)}\|\bfq(0)\|^2\gtrsim e^{\sum_{b=1}^t K(b)\|\bfq_1(0)\|^2}\geq \sum_{b=1}^t K(b)\|\bfq_1(0)\|^2
\end{equation}
where the first step is by $\log(1+x)\approx x$ when $x\rightarrow 0^+$. Therefore, one can derive that 
\begin{equation}
    \left\langle\eta\frac{1}{B}\sum_{n\in\mathcal{B}_b}\frac{\partial \textbf{Loss}(\bfX^n,y^n)}{\partial \bfW_Q^{(t)}}\bfx_j^n, \bfq_1(t)\right\rangle\gtrsim \Theta(1)\cdot K(t)
\end{equation}
Meanwhile, the value of $p_n'(t)$ will increase to $1$ during training, making the component of $\bfq_1(t)$ the major part in $\eta\frac{1}{B}\sum_{n\in\mathcal{B}_b}\frac{\partial \textbf{Loss}(\bfX^n,y^n)}{\partial \bfW_Q^{(t)}}\bfx_j^n$.\\
Hence, if $j\in\mathcal{S}_l^n$ for $l\geq3$,
\begin{equation}
\begin{aligned}
    \bfW_Q^{(t+1)}\bfx_j
    &= \bfq_l(t)+\Theta(1)\cdot \bfn_j(t)+\Theta(1)\cdot \sum_{b=0}^{t-1}K(b)\bfq_1(b)+\sum_{l=2}^M \gamma_l'\bfq_l(t)
\end{aligned}
\end{equation}
Similarly, for $j\in\mathcal{S}_2^n$,
\begin{equation}
\begin{aligned}
    \bfW_Q^{(t+1)}\bfx_j
    &= (1+K(t))\bfq_2(t)+\Theta(1)\cdot \bfn_j(t)+ \sum_{b=0}^{t-1}|K_e(b)|\bfq_1(b)+\sum_{l=2}^M \gamma_l'\bfq_l(t)
\end{aligned}
\end{equation}
\noindent (b) For the gradient of $\bfW_K$, we have
\begin{equation}
\begin{aligned}
    \frac{\partial \overline{\textbf{Loss}}_b}{\partial \bfW_K}=&\frac{1}{B}\sum_{n\in\mathcal{B}_b} \frac{\partial \textbf{Loss}(\bfX^n,y^n)}{\partial F(\bfX)}\frac{F(\bfX)}{\partial \bfW_K}\\
    =&\frac{1}{B}\sum_{n\in\mathcal{B}_b}(-y^n)\sum_{l\in\mathcal{S}^n} \sum_{i=1}^m a_{(l)_i}\mathbbm{1}[\bfW_{O_{(i,\cdot)}}\bfW_V\bfX\text{softmax}({\bfX^n}^\top\bfW_K^\top\bfW_Q\bfx_l^n)\geq0]\\
    &\cdot\Big(\bfW_{O_{(i,\cdot)}}\sum_{s\in\mathcal{S}^n} \bfW_V\bfx_s^n\text{softmax}({\bfx_s^n}^\top\bfW_K^\top\bfW_Q\bfx_l^n)\bfW_Q^\top\bfx_l^n\\
    &\cdot(\bfx_s^n-\sum_{r\in\mathcal{S}^n} \text{softmax}({\bfx_r^n}^\top\bfW_K^\top\bfW_Q\bfx_l^n)\bfx_r^n)^\top\Big)\label{grad_W_Q}
\end{aligned}
\end{equation}
Hence, for $j\in\mathcal{S}_1^n$, we can follow the derivation of (\ref{W_K_t1}) to obtain
\begin{equation}
    \bfW_K^{(t+1)}\bfx_j= (1+Q(t))\bfq_1(t)+\Theta(1)\cdot \bfo_j^n(t)\pm \sum_{b=0}^{t-1}|Q_e(b)|(1-\lambda)\bfr_2(b)+\sum_{l=3}^M  \gamma_l' \bfr_l(t),
\end{equation}
where
\begin{equation}
    \begin{aligned}
        Q(t)\geq &K(t)(1-\lambda)>0
    \end{aligned}
\end{equation}
for $\lambda<1$ introduced in Assumption \ref{asmp: lambda}, and
\begin{equation}
    |\gamma_l|\lesssim \frac{1}{B}\sum_{n\in\mathcal{B}_b} Q(t)\cdot\frac{|\mathcal{S}_l^n|}{|\mathcal{S}^n|-|\mathcal{S}_*^n|}\label{gamma}
\end{equation}
\begin{equation}
    |Q_e(t)|\lesssim \frac{1}{B}\sum_{n\in\mathcal{B}_b} Q(t)\cdot\frac{|\mathcal{S}_\#^n|}{|\mathcal{S}^n|-|\mathcal{S}_*^n|}\label{Q_e}
\end{equation}
Similarly, for $j\in\mathcal{S}_2^n$, we have
\begin{equation}
    \bfW_K^{(t+1)}\bfx_j\approx (1+Q(t))\bfq_2(t)+\Theta(1)\cdot \bfo_j^n(t)\pm \sum_{b=0}^{t-1}|Q_e(b)|(1-\lambda)\bfr_1(b)+\sum_{l=3}^M  \gamma_l' \bfr_l(t),
\end{equation}
For $j\in\mathcal{S}_z^n,\ z=3,4,\cdots,M$, we have
\begin{equation}
    \begin{aligned}
    &\Big|\left\langle \frac{1}{B}\sum_{n\in\mathcal{B}_b} \frac{\partial \textbf{Loss}(\bfX^n,y^n)}{\partial F(\bfX)}\frac{F(\bfX)}{\partial \bfW_K}\bfx_j^n, \bfq_1(t) \right\rangle\Big|\\
    \lesssim & \Big|\frac{1}{B}\sum_{n\in\mathcal{B}_b}(-y^n)\sum_{l\in\mathcal{S}_1^n} \sum_{i=1}^m a_{(l)_i}\mathbbm{1}[\bfW_{O_{(i,\cdot)}}\bfW_V\bfX\text{softmax}({\bfX^n}^\top\bfW_K^\top\bfW_Q\bfx_l^n)\geq0]\\
    &\cdot\Big(\bfW_{O_{(i,\cdot)}}(\sum_{s\in\mathcal{S}_z^n}+\lambda\sum_{s\in\mathcal{S}_1^n}) \bfW_V\bfx_s^n\text{softmax}({\bfx_s^n}^\top\bfW_K^\top\bfW_Q\bfx_l^n)\Big)\|\bfq_1(t)\|^2\Big|\\
    \leq & \lambda|Q_f(t)|\|\bfq_1(t)\|^2
    \end{aligned}
\end{equation}
\begin{equation}
    \begin{aligned}
    &\Big|\left\langle \frac{1}{B}\sum_{n\in\mathcal{B}_b} \frac{\partial \textbf{Loss}(\bfX^n,y^n)}{\partial F(\bfX)}\frac{F(\bfX)}{\partial \bfW_K}\bfx_j^n, \bfq_z(t) \right\rangle\Big|\\
    \lesssim & \Big|\frac{1}{B}\sum_{n\in\mathcal{B}_b}(-y^n)\sum_{l\in\mathcal{S}_z^n} \sum_{i=1}^m a_{(l)_i}\mathbbm{1}[\bfW_{O_{(i,\cdot)}}\bfW_V\bfX\text{softmax}({\bfX^n}^\top\bfW_K^\top\bfW_Q\bfx_l^n)\geq0]\\
    &\cdot\Big(\bfW_{O_{(i,\cdot)}}(\sum_{s\in\mathcal{S}_z^n}+\lambda\sum_{s\in\mathcal{S}_1^n}) \bfW_V\bfx_s^n\text{softmax}({\bfx_s^n}^\top\bfW_K^\top\bfW_Q\bfx_l^n)\Big)\|\bfq_z(t)\|^2\Big|\\
    \leq & \lambda|Q_f(t)|\|\bfq_z(t)\|^2
    \end{aligned}
\end{equation}
\begin{equation}
\begin{aligned}
    \bfW_K^{(t+1)}\bfx_j\approx &(1\pm c_{k_1}\lambda|Q_f(t)|)\bfq_l(t)+\Theta(1)\cdot \bfo_j^n(t)\\
    &\pm c_{k_2}\lambda\cdot \sum_{b=0}^{t-1}|Q_f(b)|\bfr_1(b)\pm c_{k_3}\lambda\cdot \sum_{b=0}^{t-1}|Q_f(b)|\bfr_2(b)+\sum_{i=3}^M  \gamma_i' \bfr_i(t),
\end{aligned}
\end{equation}
where $0<c_{k_1},c_{k_2},c_{k_3}<1$, and
\begin{equation}
    |Q_f(t)|\lesssim Q(t)
\end{equation}
Therefore, for $l\in\mathcal{S}_1^n$, if $j\in\mathcal{S}_1^n$,
\begin{equation}
    \begin{aligned}
        &{\bfx_j^n}^\top{\bfW_K^{(t+1)}}^\top\bfW_Q^{(t+1)}\bfx_l^n\\\gtrsim & (1+K(t))(1+Q(t))\|\bfq_1(t)\|^2-(\delta+\tau) \|\bfq_1(t)\|+\sum_{b=0}^{t-1}K_e(b)\sum_{b=0}^{t-1}Q_e(b)\|\bfq_2(b)\|\|\bfr_2(b)\|\\
        &+\sum_{l=3}^M \gamma_l\gamma_l'\|\bfq_l(t)\|\|\bfr_l(t)\|\\
        \gtrsim& (1+K(t))(1+Q(t))\|\bfq_1(t)\|^2-(\delta+\tau) \|\bfq_1(t)\|\\
        &-\sqrt{\sum_{l=3}^M (\frac{1}{B}\sum_{n\in\mathcal{B}_b} Q(t)\frac{|\mathcal{S}_l^n|}{|\mathcal{S}^n|-|\mathcal{S}_*^n|})^2\|\bfr_l(t)\|^2}\cdot\sqrt{\sum_{l=3}^M (\frac{1}{B}\sum_{n\in\mathcal{B}_b} K(t)\frac{|\mathcal{S}_l^n|}{|\mathcal{S}^n|-|\mathcal{S}_*^n|})^2\|\bfq_l(t)\|^2}\\
        \gtrsim & (1+K(t)+Q(t))\|\bfq_1(t)\|^2-(\delta+\tau)\|\bfq_1(t)\|
    \end{aligned}
\end{equation}
where the second step is by Cauchy-Schwarz inequality.\\
\noindent If $j\notin\mathcal{S}_1^n$,
\begin{equation}
    \begin{aligned}
        &{\bfx_j^n}^\top{\bfW_K^{(t+1)}}^\top\bfW_Q^{(t+1)}\bfx_l^n\\
        \lesssim & Q_f(t)\|\bfq_1(t)\|^2+(\delta+\tau)\|\bfq_1(t)\|
    \end{aligned}
\end{equation}
Hence, for $j,l\in\mathcal{S}_1^n$,
\begin{equation}
    \text{softmax}({\bfx_j^n}^\top\bfW_K^{(t+1)}\bfW_Q^{(t+1)}\bfx_l^n)\gtrsim \frac{e^{(1+K(t))\|\bfq_1(t)\|^2-(\delta+\tau)\|\bfq_1(t)\|}}{|\mathcal{S}_1^n|e^{(1+K(t))\|\bfq_1(t)\|^2-(\delta+\tau)\|\bfq_1(t)\|}+(|\mathcal{S}^n|-|\mathcal{S}_1^n|)}
\end{equation}
\begin{equation}
    \begin{aligned}
    &\text{softmax}({\bfx_j^n}^\top{\bfW_K^{(t+1)}}^\top\bfW_Q^{(t+1)}\bfx_l^n)-\text{softmax}({\bfx_j^n}^\top{\bfW_K^{(t)}}^\top\bfW_Q^{(t)}\bfx_l^n)\\
    \gtrsim& \frac{e^{(1+K(t))\|\bfq_1(t)\|^2-(\delta+\tau)\|\bfq_1(t)\|}}{|\mathcal{S}_1^n|e^{(1+K(t))\|\bfq_1(t)\|^2-(\delta+\tau)\|\bfq_1(t)\|}+(|\mathcal{S}^n|-|\mathcal{S}_1^n|)}
    -\frac{e^{\|\bfq_1(t)\|^2-(\delta+\tau)\|\bfq_1(t)\|}}{|\mathcal{S}_1^n|e^{\|\bfq_1(t)\|^2-(\delta+\tau)\|\bfq_1(t)\|}+(|\mathcal{S}^n|-|\mathcal{S}_1^n|)}\\
    =& \frac{|\mathcal{S}^n|-|\mathcal{S}_1^n|}{(|\mathcal{S}_1^n|e^x+(|\mathcal{S}^n|-|\mathcal{S}_1^n|))^2}e^{\|\bfq_1(t)\|^2-(\delta+\tau)\|\bfq_1(t)\|}(e^{K(t)}-1)\\
    \geq & \frac{|\mathcal{S}^n|-|\mathcal{S}_1^n|}{(|\mathcal{S}_1^n|e^{(1+K(t))\|\bfq_1(t)\|^2-(\delta+\tau)\|\bfq_1(t)\|}+(|\mathcal{S}^n|-|\mathcal{S}_1|))^2}e^{\|\bfq_1(t)\|^2-(\delta+\tau)\|\bfq_1(t)\|}\cdot K(t)
    \end{aligned}
\end{equation}
where the second to last step is by the Mean Value Theorem with \begin{equation}
    x\in[\|\bfq_1(t)\|^2-(\delta+\tau)\|\bfq_1(t)\|, (1+K(t))\|\bfq_1(t)\|^2-(\delta+\tau)\|\bfq_1(t)\|]
\end{equation}
We then need to study if $l\notin(\mathcal{S}_1^n\cup\mathcal{S}_2^n)$ and $j\in\mathcal{S}_1^n$, i.e., 
\begin{equation}
    \begin{aligned}
        {\bfx_j^n}^\top{\bfW_K^{(t+1)}}^\top\bfW_Q^{(t+1)}\bfx_l^n
        \gtrsim  (1+Q(t))\sum_{b=0}^{t-1}|K(b)|\|\bfq_1(t)\|\|\bfq_1(b)\|-(\delta+\tau)\|\bfq_1(t)\|
    \end{aligned}
\end{equation}
For $j,l\notin(\mathcal{S}_1^n\cup\mathcal{S}_2^n)$,
\begin{equation}
    \begin{aligned}
        {\bfx_j^n}^\top{\bfW_K^{(t+1)}}^\top\bfW_Q^{(t+1)}\bfx_l^n\lesssim & \pm c_{k_2}\lambda\cdot \sum_{b=0}^{t-1}|Q_f(b)|\bfr_1(b)\pm c_{k_3}\lambda\cdot \sum_{b=0}^{t-1}|Q_f(b)|\bfr_2(b)\\
          &+(1\pm c_{k_1}\lambda|Q_f(t)|)\|\bfq_l(t)\|^2
    \end{aligned}
\end{equation}
We know that the magnitude of $\|\bfq_1(t)\|$ increases along the training and finally reaches no larger than $\Theta(\sqrt{\log T})$. At the final step, we have 
\begin{equation}
    \sum_{b=0}^{t-1}K(b)\|\bfq_1(b)\|\geq \frac{T}{e^{\|\bfq_1(T)\|^2-(\delta+\tau)\|\bfq_1(T)\|}}\geq \Theta(\sqrt{\log T})
\end{equation}
Therefore, when $t$ is large enough during the training but before the final step of convergence, we have
if $j', l\notin(\mathcal{S}_1^n\cup\mathcal{S}_2^n)$ and $j\in\mathcal{S}_1^n$, we can obtain
\begin{equation}
    \begin{aligned}
        ({{\bfx_j^n}-{\bfx_{j'}^n}})^\top{\bfW_K^{(t+1)}}^\top\bfW_Q^{(t+1)}\bfx_l^n
        \gtrsim  \Theta(1)\cdot((1+K(t))\|\bfq_1(t)\|^2-(\delta+\tau)\|\bfq_1(t)\|)\label{compare_jj}
    \end{aligned}
\end{equation}
We can derive the same conclusion for $j\in\mathcal{S}_2^n$ in (\ref{compare_jj}). Therefore, by $|\mathcal{S}_2|\leq e^{-(\delta+\tau)}(1-(\sigma-\tau))|\mathcal{S}_1^n|$ in (\ref{alpha_proof}), we can obtain
\begin{equation}
\begin{aligned}
    &\text{softmax}({\bfx_j^n}^\top\bfW_K^{(t+1)}\bfW_Q^{(t+1)}\bfx_l^n)\\
    \gtrsim &\frac{e^{(1+K(t))\|\bfq_1(t)\|^2-(\delta+\tau)\|\bfq_1(t)\|}}{(|\mathcal{S}_1^n|+|\mathcal{S}_2^n|)e^{(1+K(t))\|\bfq_1(t)\|^2-(\delta+\tau)\|\bfq_1(t)\|}+(|\mathcal{S}^n|-|\mathcal{S}_1^n|-|\mathcal{S}_2^n|)}\\
    \gtrsim & \frac{e^{(1+K(t))\|\bfq_1(t)\|^2-(\delta+\tau)\|\bfq_1(t)\|}}{|\mathcal{S}_1^n|e^{(1+K(t))\|\bfq_1(t)\|^2-(\delta+\tau)\|\bfq_1(t)\|}+(|\mathcal{S}^n|-|\mathcal{S}_1^n|)}
\end{aligned}
\end{equation}
\begin{equation}
    \begin{aligned}
    &\text{softmax}({\bfx_j^n}^\top{\bfW_K^{(t+1)}}^\top\bfW_Q^{(t+1)}\bfx_l^n)-\text{softmax}({\bfx_j^n}^\top{\bfW_K^{(t)}}^\top\bfW_Q^{(t)}\bfx_l^n)\\
    \gtrsim & \frac{|\mathcal{S}^n|-|\mathcal{S}_1^n|}{(|\mathcal{S}_1^n|e^{\Theta(1)\cdot((1+K(t))\|\bfq_1(t)\|^2-(\delta+\tau)\|\bfq_1(t))\|}+(|\mathcal{S}^n|-|\mathcal{S}_1|))^2}e^{\Theta(1)(\|\bfq_1(t)\|^2-(\delta+\tau)\|\bfq_1(t)\|)}\cdot K(t)\\
    \gtrsim & \frac{|\mathcal{S}^n|-|\mathcal{S}_1^n|}{(|\mathcal{S}_1^n|e^{(1+K(t))\|\bfq_1(t)\|^2-(\delta+\tau)\|\bfq_1(t)\|}+(|\mathcal{S}^n|-|\mathcal{S}_1|))^2}e^{\|\bfq_1(t)\|^2-(\delta+\tau)\|\bfq_1(t)\|}\cdot K(t)
    \end{aligned}
\end{equation}
Meanwhile, for $l\in\mathcal{S}_1^n$ and $j\notin\mathcal{S}_1^n$,
\begin{equation}
    \text{softmax}({\bfx_j^n}^\top{\bfW_K^{(t+1)}}^\top\bfW_Q^{(t+1)}\bfx_l^n)\lesssim \frac{1}{|\mathcal{S}_1^n|e^{(1+K(t))\|\bfq_1(t)\|^2-(\delta+\tau)\|\bfq_1(t)\|}+(|\mathcal{S}^n|-|\mathcal{S}_1^n|)}
\end{equation}
\begin{equation}
    \begin{aligned}
    &\text{softmax}({\bfx_j^n}^\top{\bfW_K^{(t+1)}}\bfW_Q^{(t+1)}\bfx_l^n)-\text{softmax}({\bfx_j^n}^\top{\bfW_K^{(t)}}^\top\bfW_Q^{(t)}\bfx_l^n)\\
    \lesssim& \frac{1}{|\mathcal{S}_1^n|e^{(1+K(t))\|\bfq_1(t)\|^2-(\delta+\tau)\|\bfq_1(t)\|}+(|\mathcal{S}^n|-|\mathcal{S}_1^n|)}
    -\frac{1}{|\mathcal{S}_1^n|e^{\|\bfq_1(t)\|^2-(\delta+\tau)\|\bfq_1(t)\|}+(|\mathcal{S}^n|-|\mathcal{S}_1^n|)}\\
    =& -\frac{|\mathcal{S}_1^n|}{(|\mathcal{S}_1^n|e^x+(|\mathcal{S}^n|-|\mathcal{S}_1^n|))^2}e^{\|\bfq_1(t)\|^2-(\delta+\tau)\|\bfq_1(t)\|}(e^{K(t)}-1)\\
    \leq & -\frac{|\mathcal{S}_1^n|}{(|\mathcal{S}_1|e^{(1+K(t))\|\bfq_1(t)\|^2-(\delta+\tau)\|\bfq_1(t)\|}+(|\mathcal{S}^n|-|\mathcal{S}_1^n|))^2}e^{\|\bfq_1(t)\|^2-(\delta+\tau)\|\bfq_1(t)\|}\cdot K(t)
    \end{aligned}
\end{equation}
where the second to last step is by the Mean Value Theorem with \begin{equation}
    x\in[\|\bfq_1(t)\|^2-(\delta+\tau)\|\bfq_1(t)\|, (1+K(t))\|\bfq_1(t)\|^2-(\delta+\tau)\|\bfq_1(t)\|]
\end{equation}
The same conclusion holds if $l\notin(\mathcal{S}_1^n\cup\mathcal{S}_2^n)$ and $j\notin\mathcal{S}_1^n$.\\
\noindent Note that
\begin{equation}
    \bfq_1(t+1)=\sqrt{(1+K(t))}\bfq_1(t)
\end{equation}
\begin{equation}
    \bfq_2(t+1)=\sqrt{(1+K(t))}\bfq_2(t)
\end{equation}
\begin{equation}
    \bfr_1(t+1)=\sqrt{(1+Q(t))}\bfr_1(t)
\end{equation}
\begin{equation}
    \bfr_2(t+1)=\sqrt{(1+Q(t))}\bfr_2(t)
\end{equation}

\noindent It can also be verified that this claim holds when $t=1$.

\subsection{\textbf{Proof of Claim \ref{clm: W_V} of Lemma \ref{lm: training}}}
The computation of the gradient of $\bfW_V$ is straightforward. The gradient would be related to $\bfW_O$ by their connections. One still need to study the influence of the gradient on different patterns, where we introduce the discussion for the term $V_i(t)$'s. 

\noindent For the gradient of $\bfW_V$, by (\ref{loss_batch}) we have
\begin{equation}
\begin{aligned}
    \frac{\partial \overline{\textbf{Loss}}_b}{\partial \bfW_V}=&\frac{1}{B}\sum_{n\in\mathcal{B}_b} \frac{\partial \textbf{Loss}(\bfX^n,y^n)}{\partial F(\bfX^n)}\frac{\partial F(\bfX^n)}{\partial \bfW_V}\\
    =&-y\frac{1}{B}\sum_{n\in\mathcal{B}_b}\frac{1}{|\mathcal{S}^n|}\sum_{l\in\mathcal{S}^n} \sum_{i=1}^m a_{(l)_i}^*\mathbbm{1}[\bfW_{O_{(i,\cdot)}}\bfW_V\bfX^n\text{softmax}({\bfX^n}^\top\bfW_K^\top\bfW_Q\bfx_l^n)\geq0]\\
    &\cdot{\bfW_{O_{(i,\cdot)}}}^\top\text{softmax}({\bfX^n}^\top\bfW_K^\top\bfW_Q\bfx_l^n)^\top{\bfX^n}^\top\\
\end{aligned}
\end{equation}

\noindent Consider a data $\{\bfX^n,y^n\}$ where $y^n=1$. Let $l\in\mathcal{S}_1^n$
\begin{equation}
    \sum_{s\in\mathcal{S}_1^n}\text{softmax}({\bfx_s^n}^\top{\bfW_K^{(t)}}^\top\bfW_Q^{(t)}\bfx_l^n)\geq p_n(t)
\end{equation}
Then for $j\in\mathcal{S}_1^g,\ g\in[N]$,
\begin{equation}
\begin{aligned}
    &\frac{1}{B}\sum_{n\in\mathcal{B}_b}\frac{\partial \textbf{Loss}(\bfX^n,y^n)}{\partial \bfW_V^{(t)}}\Big|\bfW_V^{(t)}\bfx_j^g\\
    =&\frac{1}{B}\sum_{n\in\mathcal{B}_b}(-y^n)\frac{1}{|\mathcal{S}^n|}\sum_{l\in\mathcal{S}^n} \sum_{i=1}^m a_{(l)_i}\mathbbm{1}[\bfW_{O_{(i,\cdot)}}^{(t)}\sum_{s\in\mathcal{S}^n}\text{softmax}({\bfx_s^n}^\top{\bfW_K^{(t)}}^\top\bfW_Q^{(t)}{\bfx_l^n})\bfW_V^{(t)}{\bfx_s^n}\geq0]\\
    &\cdot{\bfW_{O_{(i,\cdot)}}^{(t)}}^\top\sum_{s\in\mathcal{S}^n}\text{softmax}({\bfx_s^n}^\top{\bfW_K^{(t)}}^\top\bfW_Q^{(t)}\bfx_l){\bfx_s^n}^\top\bfx_j^g\\
   = & \sum_{i\in\mathcal{W}_{l,n}(0)} V_i(t){\bfW_{O_{(i,\cdot)}}}^\top+\sum_{i\notin\mathcal{W}_{l,n}(0)} \lambda V_i(t){\bfW_{O_{(i,\cdot)}}}^\top,
\end{aligned}
\end{equation}
When $t=0$, only $\bfW_{O_{(i,\cdot)}}^{(t)}$ from $i\in\mathcal{W}_{l,n}(0)$ can ensure the indicator be $1$ for $l\in\mathcal{S}_1^n$. 
If $i\in\mathcal{W}_{l,n}(0)$, by the fact that $\mathcal{S}_\#^n$ contributes more to $V_i(t)$ compared to $\mathcal{S}_l^n$ for $l\geq 3$ and Assumption \ref{asmp: lambda}, we have
\begin{equation}
\begin{aligned}
    V_i(t)\lesssim &\frac{1}{2B}\sum_{n\in{\mathcal{B}_b}_+}-\frac{|\mathcal{S}_1^n|}{a|\mathcal{S}^n|}p_n(t)+\frac{|\mathcal{S}_2^n|}{a|\mathcal{S}^n|}|\lambda|\nu_n(t)(|\mathcal{S}^n|-|\mathcal{S}_1^n|)\\
    \lesssim &\frac{1}{2B}\sum_{n\in{\mathcal{B}_b}_+}-\frac{|\mathcal{S}_1^n|}{a|\mathcal{S}^n|}p_n(t)\label{Vi_W}
\end{aligned}
\end{equation}
Similarly, if $i\in\mathcal{U}_{l,n}(0)$,
\begin{equation}
    V_i(t)\gtrsim \frac{1}{2B}\sum_{n\in{\mathcal{B}_b}_-}\frac{|\mathcal{S}_2^n|}{a|\mathcal{S}^n|}p_n(t)\label{Vi_U}
\end{equation}
if $i$ is an unlucky neuron, by Hoeffding's inequality in (\ref{hoeffding}), we have
\begin{equation}
\begin{aligned}
    |V_i(t)|\leq& \frac{1}{\sqrt{B}}\cdot \frac{1}{a}\\
    \lesssim &\frac{1}{\sqrt{B} a}\label{Vi_non}
\end{aligned}
\end{equation}
which is smaller than $V_i(t)$ for $i\in\mathcal{W}_{l,n}(0)$ and $i\in\mathcal{U}_{l,n}(0)$ given $B\geq \Omega(1)$. When $t$ is large, since unlucky neurons can activate tokens from both $+1$ and $-1$ data, we still have that (\ref{Vi_W}), (\ref{Vi_U}) and (\ref{Vi_non}) hold. Then, 
for $i\in\mathcal{W}_{l,n}(0)$, we have
\begin{equation}
    \begin{aligned}
        &-\eta \sum_{b=1}^t\bfW_{O_{(i,\cdot)}}^{(b)}\sum_{j\in\mathcal{W}_{l,n}(0)}V_j(b){\bfW_{O_{(j,\cdot)}}^{(b)}}^\top\\
        \gtrsim &\eta  m  \frac{1}{2B}\sum_{b=1}^t\sum_{n\in{\mathcal{B}_b}_+}\frac{|\mathcal{S}_1^n|}{a|\mathcal{S}^n|}p_n(b)\\
        &\cdot ((\frac{\xi}{aB}\sum_{n\in\mathcal{B}_b}\frac{\eta (t+1)^2 |\mathcal{S}_1^n|}{|\mathcal{S}^n|}\frac{1}{4B}\sum_{n\in\mathcal{B}_b}\frac{|\mathcal{S}_1^n|m}{|\mathcal{S}^n|a}p_n(t)  )^2
    \end{aligned}
\end{equation}
\begin{equation}
    \begin{aligned}
        &|\eta \sum_{b=1}^t\bfW_{O_{(i,\cdot)}}^{(b)}\sum_{j\in\mathcal{U}_{l,n}(0)}V_j(b){\bfW_{O_{(j,\cdot)}^{(b)}}}^\top|\\
        \lesssim  &\frac{\eta}{B}\sum_{b=1}^t\sum_{n\in\mathcal{B}_b}\frac{|\mathcal{S}_2^n|p_n(b)m}{|\mathcal{S}^n|a}  \|\bfW_{O_{(i,\cdot)}}^{(t)}\|^2\|\bfp_1\|^2
    \end{aligned}
\end{equation}
\begin{equation}
    \begin{aligned}
        -\eta t\bfW_{O_{(i,\cdot)}}\sum_{j\notin(\mathcal{W}_{l,n}(0)\cup\mathcal{U}_{l,n}(0))}V_j(t){\bfW_{O_{(j,\cdot)}}}^\top
        \lesssim \frac{\eta t m\|\bfp\|^2}{Ba}\|\bfW_{O_{(i,\cdot)}}^{(t)}\|^2
    \end{aligned}
\end{equation}

\noindent Hence, \\
\noindent (1) If $j\in\mathcal{S}_1^n$ for one $n\in[N]$,
\begin{equation}
\begin{aligned}
    \bfW_V^{(t+1)}\bfx_j^n&=\bfW_V^{(t)}\bfx_j^n-\eta\Big(\frac{\partial L}{\partial \bfW_V}\Big|\bfW_V^{(t)}\Big)\bfx_j^n\\
    &=\bfp_1-\eta\sum_{b=1}^{t+1}\sum_{i\in\mathcal{W}_{l,n}(0)} V_i(b){\bfW_{O_{(i,\cdot)}}^{(b)}}^\top-\eta\sum_{b=1}^{t+1}\sum_{i\notin\mathcal{W}_{l,n}(0)} \lambda V_i(b){\bfW_{O_{(i,\cdot)}}^{(b)}}^\top+\bfz_j(t)
\end{aligned}
\end{equation}
(2) If $j\in\mathcal{S}_2^n$, we have
\begin{equation}
\begin{aligned}
    \bfW_V^{(t+1)}\bfx_j&=\bfW_V^{(0)}\bfx_j^n-\eta\Big(\frac{\partial L}{\partial \bfW_V}\Big|\bfW_V^{(0)}\Big)\bfx_j^n\\
    &=\bfp_2-\eta\sum_{b=1}^{t+1}\sum_{i\in\mathcal{U}_{l,n}(0)} V_i(b){\bfW_{O_{(i,\cdot)}}^{(b)}}^\top-\eta\sum_{b=1}^{t+1}\sum_{i\notin\mathcal{U}_{l,n}(0)} \lambda V_i(b){\bfW_{O_{(i,\cdot)}}^{(b)}}^\top+\bfz_j(t)
\end{aligned}
\end{equation}
(3) If $j\in\mathcal{S}^n/(\mathcal{S}_1^n\cup\mathcal{S}_2^n)$, we have
\begin{equation}
\begin{aligned}
    \bfW_V^{(t+1)}\bfx_j^n&=\bfW_V^{(0)}\bfx_j^n-\eta\Big(\frac{\partial L}{\partial \bfW_V}\Big|\bfW_V^{(0)}\Big)\bfx_j^n\\
    &=\bfp_j-\eta\sum_{b=1}^{t+1}\sum_{i=1}^m \lambda V_i(b){\bfW_{O_{(i,\cdot)}}^{(b)}}^\top+\bfz_j(t)
\end{aligned}
\end{equation}
Here \begin{equation}
    \|\bfz_j(t)\|\leq (\sigma+\tau)
\end{equation}
for $t\geq1$. 
Note that this claim also holds when $t=1$.

\section{Other useful lemmas}\label{sec: lemmas}

\begin{lemma}\label{lemma: initial_WU}
The number of lucky neurons at the initialization $|\mathcal{W}_{l,n}(0)|$, $|\mathcal{U}_{l,n}(0)|$ satisfies
\begin{equation}
    |\mathcal{W}_{l,n}(0)|,\ |\mathcal{U}_{l,n}(0)|\geq \Omega(m)
\end{equation}
\end{lemma}
\textbf{Proof:}\\
We know that the Gaussian initialization of $\bfW_{O_{(i,\cdot)}}^{(0)}$ generates a uniform distribution on the $m_a-1$-sphere. Therefore,
\begin{equation}
    \Pr(i\in\mathcal{W}_{l,n}(0))=A_{m_a}^{cap}(\phi)/A_{m_a},
\end{equation}
where $A_{m_a}$ is the surface area of an $m_a-1$-sphere. $A_{m_a}^{cap}(\phi)$ is the surface area of a $m_a-1$-spherical cap with $\phi$ as the colatitude angle. By Equation 1 in \citep{L10}, we have
\begin{equation}
    \Pr(i\in\mathcal{W}_{l,n}(0))=\frac{1}{2}I_{\sin^2{\phi}}(\frac{m_a-1}{2}, \frac{1}{2})=\frac{\int_{0}^{\sin^2{\phi}}t^{\frac{m_a-3}{2}}(1-t)^{-\frac{1}{2}}dt}{\int_{0}^{1}t^{\frac{m_a-3}{2}}(1-t)^{-\frac{1}{2}}dt},
\end{equation}
where $I_{\cdot}(\cdot,\cdot)$ is the regularized incomplete beta function. Since that 
\begin{equation}
    \phi\leq \pi/2-\sigma-\tau=\pi/2-\Theta(1/M),
\end{equation}
we have that 
\begin{equation}
    \begin{aligned}
        &\frac{\int_{0}^{\sin^2{\phi}}t^{\frac{m_a-3}{2}}(1-t)^{-\frac{1}{2}}dt}{\int_{0}^{1}t^{\frac{m_a-3}{2}}(1-t)^{-\frac{1}{2}}dt}\\
        \geq & \frac{\int_{0}^{\cos^2{1/M}}t^{\frac{m_a-3}{2}}(1-t)^{-\frac{1}{2}}dt}{\int_{0}^{1}t^{\frac{m_a-3}{2}}(1-t)^{-\frac{1}{2}}dt}\\
        \geq & 1-\frac{\int_{1-\Theta(1/M^2)}^{1}t^{\frac{m_a-3}{2}}(1-t)^{-\frac{1}{2}}dt}{\int_{0}^{1}t^{\frac{m_a-3}{2}}(1-t)^{-\frac{1}{2}}dt}\\
        \geq & 1-\frac{\int_{1-\Theta(1/M^2)}^{1}(1-t)^{-\frac{1}{2}}dt}{\int_{0}^{1}t^{\frac{m_a-3}{2}}dt}\\
        =& 1-\frac{\Theta(\frac{2}{M})}{\frac{1}{\frac{m_a-1}{2}}}\\
        \geq & \Theta(1)
    \end{aligned}
\end{equation}
where the last step is by $m_a=\Theta(M)$. This implies that
\begin{equation}
    |\mathcal{W}_{l,n}(0)|\geq \Omega(m)
\end{equation}
Likewise, the conclusion holds for $\mathcal{U}_{l,n}(0)$.

\begin{lemma}\label{lemma: update_WU}
Under the condition that $m\gtrsim M^2\log N$, we have the following result.\\
For $i\in\mathcal{W}_{l,n}(0)$ and $l\in\mathcal{S}_1^n$, we have
\begin{equation}
    \mathbbm{1}[\bfW_{O_{(i,\cdot)}}^{(t)}\bfV_l^n(t)]=1;
\end{equation}
For $i\in\mathcal{U}_{l,n}(0)$ and $l\in\mathcal{S}_2^n$, we have
\begin{equation}
    \mathbbm{1}[\bfW_{O_{(i,\cdot)}}^{(t)}\bfV_l^n(t)]=1;
\end{equation}
\end{lemma}
\textbf{Proof:}\\
We prove this lemma by induction.\\
\noindent When $t=0$. For $i\in\mathcal{W}_l^n(0)$ and $l\in\mathcal{S}_1^n$, we have that
\begin{equation}
    \bfW_{O_{(i,\cdot)}}^{(0)}(\sum_{s\in\mathcal{S}_1^n}\text{softmax}({\bfx_s^n}^\top{\bfW_K^{(t)}}^\top\bfW_Q^{(t)}\bfx_l^n)\bfp_1+\bfz(0)+\sum_{j\neq 1}W_j^n(0)\bfp_j)\gtrsim \xi(\Theta(1)-\sigma-\tau)>0
\end{equation}
Hence, the conclusion holds. When $t=1$, we have
\begin{equation}
    \begin{aligned}
        &\bfW_{O_{(i,\cdot)}}^{(t)}\bfV_l^n(t)\\
        =& \bfW_{O_{(i,\cdot)}}^{(t)}\Big(\sum_{s\in\mathcal{S}_1^n}\text{softmax}({\bfx_s^n}^\top{\bfW_K^{(t)}}^\top\bfW_Q^{(t)}\bfx_l^n)\bfp_1+\bfz(t)+\sum_{j\neq 1}W_j^n(t)\bfp_j\\
&-\eta \sum_{b=0}^{t-1}(\sum_{i\in \mathcal{W}_{l,n}(0)}V_i(b){\bfW_{O_{(i,\cdot)}}^{(b)}}^\top+\sum_{i\notin \mathcal{W}_{l,n}(0)}V_i(b)\lambda{\bfW_{O_{(i,\cdot)}}^{(b)}}^\top)\Big)
    \end{aligned}
\end{equation}
Denote $\theta_{l,n}^i$ as the angle between $\bfV_l^n(0)$ and $\bfW_{O_{(i,\cdot)}}^{(0)}$. Since that $\bfW_{O_{(j,\cdot)}}^{(0)}$ is initialized uniformed on the $m_a-1$-sphere, we have $\mathbb{E}[\theta_{l,n}^i]=0$. By Hoeffding's inequality (\ref{hoeffding}), we have
\begin{equation}
    \Big\|\frac{1}{|\mathcal{W}_{l,n}(0)|}\sum_{i\in\mathcal{W}_{l,n}(0)}\theta_{l,n}^i-\mathbb{E}[\theta_{l,n}^i]\Big\|=\Big\|\frac{1}{|\mathcal{W}_{l,n}(0)|}\sum_{i\in\mathcal{W}_{l,n}(0)}\theta_{l,n}^i\Big\|\leq \sqrt{\frac{\log N}{m}},\label{concentration_theta}
\end{equation}
with probability of at least $1-N^{-10}$. When $m\gtrsim M^2\log N$, we can obtain that 
\begin{equation}
    \Big\|\frac{1}{|\mathcal{W}_{l,n}(0)|}\sum_{i\in\mathcal{W}_{l,n}(0)}\theta_{l,n}^i-\mathbb{E}[\theta_{l,n}^i]\Big\|\leq O(\frac{1}{M}).
\end{equation}
Therefore, for $i\in\mathcal{W}_{l,n}(0)$, we have
\begin{equation}
    \bfW_{O_{(i,\cdot)}}\sum_{b=0}^{t-1}\sum_{i\in\mathcal{W}_{l,n}(0)}\bfW_{O_{(i,\cdot)}}^{(b)}>0
\end{equation}
Similarly, we have that $\sum_{b=0}^{t-1}\sum_{i\notin \mathcal{W}_{l,n}(0)}\bfW_{O_{(i,\cdot)}}^{(b)}$ is close to $-\bfV_l^n{(0)}$. Given that $\lambda<1$, we can approximately acquire that 
\begin{equation}
    -\bfW_{O_{(i,\cdot)}}^{(0)}\eta \sum_{b=0}^{t-1}(\sum_{i\in \mathcal{W}_{l,n}(0)}V_i(b){\bfW_{O_{(i,\cdot)}}^{(b)}}^\top+\sum_{i\notin \mathcal{W}_{l,n}(0)}V_i(b)\lambda{\bfW_{O_{(i,\cdot)}}^{(b)}}^\top)>0
\end{equation}
Since that $i\in\mathcal{W}_{l,n}(0)$, we have
\begin{equation}
    \bfW_{O_{(i,\cdot)}}^{(0)}(\sum_{s\in\mathcal{S}_1^n}\text{softmax}({\bfx_s^n}^\top{\bfW_K^{(t)}}^\top\bfW_Q^{(t)}\bfx_l^n)\bfp_1+\bfz(t)+\sum_{j\neq 1}W_j^n(t)\bfp_j)>0
\end{equation}
Therefore, we have
\begin{equation}
\bfW_{O_{(i,\cdot)}}^{(0)}\bfV_l^n(t)>0
\end{equation}
Meanwhile, the addition from $\bfW_{O_{(i,\cdot)}}^{(0)}$ to $\bfW_{O_{(i,\cdot)}}^{(1)}$ is approximately a summation of multiple $\bfV_j^n(0)$ such that $\bfW_{O_{(i,\cdot)}}^{(0)}\bfV_j^n(0)>0$ and $j\in\mathcal{S}_1^n$. Therefore, ${\bfV_j^n(0)}^\top\bfV_l^n(0)>0$. Therefore, we can obtain
\begin{equation}
    \bfW_{O_{(i,\cdot)}}^{(t)}\bfV_l^n(t)>0
\end{equation}
\noindent \textbf{(2) Suppose that the conclusion holds when $t=s$. When $t=s+1$}, we can follow the derivation of the case where $t=1$. Although the unit vector of $\bfW_{O_{(i,\cdot)}}^{(t)}$ no longer follows a uniform distribution, we know that (\ref{concentration_theta}) holds since the angle is bounded and has a mean which is very close to $\bfV_l^n(0)$. Then, the conclusion still holds.

\noindent One can develop the proof for $\mathcal{U}_{l,n}(0)$ following the above steps.

\section{Extension to more general cases}\label{sec: extension}
\subsection{Extension to multi-classification}\label{subsec: ext_multi}

Consider the classification problem with four classes, we use the label $y\in\{+1,-1\}^2$ to denote the corresponding class. Similarly to the previous setup, there are four orthogonal discriminative patterns. In the output layer, $a_{l_{(i)}}$ for the data $(\bfX^n, y^n)$ is changed into an $\mathbb{R}^2$ vector $\bfa_{l_{(i)}}$ for $l\in[|\mathcal{S}^n|]$ and $i\in[m]$. Hence, we define
\begin{equation}
    \bfF(\bfX^n)=\frac{1}{|\mathcal{S}^n|}\sum_{l\in\mathcal{S}^n}\bfa_{l_{(i)}}\text{Relu}(\bfW_{O}\bfW_{V}\bfX^n\text{softmax}({\bfX^n}^\top\bfW_{Q}^\top\bfW_{K}\bfx_l^n))\label{new_network}
\end{equation}
\begin{equation}
    F_1(\bfX^n)=\frac{1}{|\mathcal{S}^n|}\sum_{l\in\mathcal{S}^n}a_{l_{1(i)}}\text{Relu}(\bfW_{O}\bfW_{V}\bfX^n\text{softmax}({\bfX^n}^\top\bfW_{Q}^\top\bfW_{K}\bfx_l^n))
\end{equation}
\begin{equation}
    F_2(\bfX^n)=\frac{1}{|\mathcal{S}^n|}\sum_{l\in\mathcal{S}^n}a_{l_{2(i)}}\text{Relu}(\bfW_{O}\bfW_{V}\bfX^n\text{softmax}({\bfX^n}^\top\bfW_{Q}^\top\bfW_{K}\bfx_l^n))
\end{equation}
The dataset $\mathcal{D}$ can be divided into four groups as
\begin{equation}
    \begin{aligned}
    \mathcal{D}_1=&\{(\bfX^n,\bfy^n)|\bfy^n=(1,1)\}\\
    \mathcal{D}_2=&\{(\bfX^n,\bfy^n)|\bfy^n=(1,-1)\}\\
    \mathcal{D}_3=&\{(\bfX^n,\bfy^n)|\bfy^n=(-1,1)\}\\
    \mathcal{D}_4=&\{(\bfX^n,\bfy^n)|\bfy^n=(-1,-1)\}
    \end{aligned}
\end{equation}
The hinge loss function for data $(\bfX^n, \bfy^n)$ will be
\begin{equation}
    \text{Loss}(\bfX^n,\bfy^n)=\max\{1-{\bfy^n}^\top \bfF(\bfX^n),0\}
\end{equation}
We can divide the weights $\bfW_{O_{(i,\cdot)}}$ ($i\in[m]$) into two groups, respectively. 
\begin{equation}
    \begin{aligned}
    \mathcal{W}_1=&\{i|\bfa_{l_{(i)}}=\frac{1}{\sqrt{m}}\cdot(1,1)\}\\
    \mathcal{W}_2=&\{i|\bfa_{l_{(i)}}=\frac{1}{\sqrt{m}}\cdot(1,-1)\}\\
    \mathcal{W}_3=&\{i|\bfa_{l_{(i)}}=\frac{1}{\sqrt{m}}\cdot(-1,1)\}\\
    \mathcal{W}_4=&\{i|\bfa_{l_{(i)}}=\frac{1}{\sqrt{m}}\cdot(-1,-1)\}\\
    \end{aligned}
\end{equation}Therefore, for $\bfW_{O_u}$ in the network (\ref{new_network}), we have
\begin{equation}
    \frac{\partial \text{Loss}(\bfX^n,\bfy^n)}{\partial {\bfW_{O_{(i,\cdot)}}}^\top}=-y^n_{1} \frac{\partial F_1(\bfX^n)}{\partial \bfW_{O_{1(i,\cdot)}}}-y^n_{2}\frac{\partial F_2(\bfX^n)}{\bfW_{O_{2(i,\cdot)}}}
\end{equation}
where the derivation of $\frac{\partial F_1(\bfX^n)}{\partial \bfW_{O_{1(i,\cdot)}}}$ and $\frac{\partial F_2(\bfX^n)}{\partial \bfW_{O_{2(i,\cdot)}}}$ can be found in the analysis of binary classification above. For any $i\in\mathcal{W}_2$, suppose that we only care about $\bfp_i, i=1,2,3,4$ in the gradient. Following the proof of Claim \ref{clm: W_O} of Lemma \ref{lm: training}, if the data $(\bfX^n,y^n)\in\mathcal{D}_2$, we have
\begin{equation}
    -\frac{\partial \text{Loss}(\bfX^n,\bfy^n)}{\partial {\bfW_{O_{(i,\cdot)}}}^\top}=y^n_{1} \frac{\partial F_1(\bfX^n)}{\partial \bfW_{O_{1(i,\cdot)}}}+y^n_{2}\frac{\partial F_2(\bfX^n)}{\bfW_{O_{2(i,\cdot)}}}\approx\propto1\cdot\frac{1}{\sqrt{m}}\bfp_2-1\cdot(-\frac{1}{\sqrt{m}})\bfp_2=\frac{2}{\sqrt{m}}\bfp_2
\end{equation}
\begin{equation}
    (\bfW_{O_{(i,\cdot)}}^{(t+1)}-\bfW_{O_{(i,\cdot)}}^{(t)})\bfp_2\propto \|\bfp_2\|^2>0
\end{equation}
if $(\bfX^n,y^n)\in\mathcal{D}_1$, we have
\begin{equation}
    -\frac{\partial \text{Loss}(\bfX^n,\bfy^n)}{\partial {\bfW_{O_{(i,\cdot)}}}^\top}\approx \propto 1\cdot\frac{1}{\sqrt{m}}\bfp_1+1\cdot(-\frac{1}{\sqrt{m}})\bfp_1=0
\end{equation}
\begin{equation}
    (\bfW_{O_{(i,\cdot)}}^{(t+1)}-\bfW_{O_{(i,\cdot)}}^{(t)})\bfp_1\approx=0
\end{equation}
if $(\bfX^n,y^n)\in\mathcal{D}_3$, we have
\begin{equation}
    -\frac{\partial \text{Loss}(\bfX^n,\bfy^n)}{\partial {\bfW_{O_{(i,\cdot)}}}^\top}\approx \propto -1\cdot\frac{1}{\sqrt{m}}\bfp_3+1\cdot(-\frac{1}{\sqrt{m}})\bfp_3=-\frac{2}{\sqrt{m}}\bfp_3
\end{equation}
\begin{equation}
    (\bfW_{O_{(i,\cdot)}}^{(t+1)}-\bfW_{O_{(i,\cdot)}}^{(t)})\bfp_3\leq0
\end{equation}
if $(\bfX^n,y^n)\in\mathcal{D}_4$, we have
\begin{equation}
    -\frac{\partial \text{Loss}(\bfX^n,\bfy^n)}{\partial {\bfW_{O_{(i,\cdot)}}}^\top}\approx \propto -1\cdot\frac{1}{\sqrt{m}}\bfp_4-1\cdot(-\frac{1}{\sqrt{m}})\bfp_4=0
\end{equation}
\begin{equation}
    (\bfW_{O_{(i,\cdot)}}^{(t+1)}-\bfW_{O_{(i,\cdot)}}^{(t)})\bfp_4\approx0
\end{equation}
By the algorithm, $\bfW_{O_{(i,\cdot)}}$ will update along the direction of $\bfp_2$ for $i\in\mathcal{W}_2$. We can analyze $\bfW_V$, $\bfW_K$ and $\bfW_Q$ similarly.

\subsection{Extension to a more general data model}\label{subsec: general_data}
We generalize the patterns from vectors to sets of vectors. Consider that there are $M\ (2<M<m_a, m_b)$ distinct sets $\{\mathcal{M}_1,\mathcal{M}_2,\cdots,\mathcal{M}_M\}$ where $\mathcal{M}_l=\{\bfmu_{l,1},\bfmu_{l,2},\cdots,\bfmu_{l,l_m}\}$, $l_m\geq1$. $\mathcal{M}_1$, $\mathcal{M}_2$ denote sets of discriminative patterns for the binary labels, and $\mathcal{M}_3,\cdots,\mathcal{M}_M$ are sets of non-discriminative patterns.  
\begin{equation}
    \kappa=\min_{1\leq i\neq j\leq M, 1\leq a\leq i_m, 1\leq b\leq j_m}\|\bfmu_{i,a}-\bfmu_{j,b}\|>0
\end{equation}
is the minimum distance between patterns of different sets. Each token $\bfx_l^n$ of $\bfX^n$ is a noisy version of one pattern, i.e.,
\begin{equation}
    \min_{j\in[M], b\in[j_m]}\|\bfx_l^n-\bfmu_{j,b}\|\leq \tau
\end{equation}
Define that for $l$, $s$ corresponding to $b_1$, $b_2$, respectively,
\begin{equation}
    \min_{j\in[M], b_1, b_2\in[j_m]}\|\bfx_l^n-\bfx_s^n\|\leq \Delta,
\end{equation}
we have $2\tau+\Delta< \kappa$.\\
To simplify our theoretical analysis, one can similarly rescale all tokens a little bit like in Assumption \ref{asmp: lambda} such that tokens corresponding to patterns in the same pattern set has an inner product larger than 1, while tokens corresponding to patterns from different pattern sets has an inner product smaller than $\lambda<1$. \\
Assumption \ref{asmp: VQK_initialization} can be modified such that 
\begin{equation}
    \|\bfW_V^{(0)}\bfmu_{j,b}-\bfp_{j,b}\|\leq \sigma
\end{equation}
\begin{equation}
    \|\bfW_K^{(0)}\bfmu_{j,b}-\bfq_{j,b}\|\leq \delta
\end{equation}
\begin{equation}
    \|\bfW_Q^{(0)}\bfmu_{j,b}-\bfr_{j,b}\|\leq \delta
\end{equation}
where $\bfp_{j,b}\perp\bfp_{i,a}$ for any $i,j\in[M]$, $b\in[j_m]$, and $a\in[i_m]$. Likewise, $\bfq_{j,b}\perp\bfq_{i,a}$ for any $i,j\in[M]$, $b\in[j_m]$, and $a\in[i_m]$. $\bfr_{j,b}\perp\bfr_{i,a}$ for any $i,j\in[M]$, $b\in[j_m]$, and $a\in[i_m]$.\\
Therefore, we make sure that the initial query, key and value features from different sets of patterns are still close to be orthogonal to each other. Then, we can follow our main proof idea. To be more specific, for label-relevant tokens $\bfx_l^n$, by computing (\ref{grad_Wk}) and (\ref{grad_W_Q}), $\bfW_K^{(t)}\bfx_l^n$, $\bfW_Q^{(t)}\bfx_l^n$ will grow in the direction of a fixed linear combination of $\bfq_{l,1},\cdots,\bfq_{l,l_m}$, and $\bfr_{l,1},\cdots,\bfr_{l,l_m}$. The coefficient of the linear combination is a function of fractions of different pattern vectors $\bfmu_{l,b}$ in $\mathcal{M}_l$. One can still derive a sparse attention map with weights of non-discriminative patterns decreasing to be close to zero during the training.

\subsection{Extension to multi-head networks}\label{subsec: multi-head}
Suppose there are $H$ heads in total. The network is modified to
\begin{equation}
\begin{aligned}
    F(\bfX^n)&=\frac{1}{|\mathcal{S}^n|}\sum_{l\in\mathcal{S}^n}\bfa_{(l)}^\top\text{Relu}(\bfW_O\bigparallel_{h=1}^H\sum_{s\in\mathcal{S}^n}\bfW_{V_h}\bfx_s^n\text{softmax}({\bfx_s^n}^\top\bfW_{K_h}^\top\bfW_{Q_h}\bfx_l^n))\\
    &=\frac{1}{|\mathcal{S}^n|}\sum_{l\in\mathcal{S}^n}\bfa_{(l)}^\top\text{Relu}(\sum_{h=1}^H\bfW_{O_h}\sum_{s\in\mathcal{S}^n}\bfW_{V_h}\bfx_s^n\text{softmax}({\bfx_s^n}^\top\bfW_{K_h}^\top\bfW_{Q_h}\bfx_l^n))
\end{aligned}
\end{equation}
where $\bfW_{V_h}\in\mathbb{R}^{m_a\times d}$, $\bfW_O=(\bfW_{O_1},\bfW_{O_2},\cdots \bfW_{O_H})\in\mathbb{R}^{m\times Hm_a}$, and $\bfW_{O_h}\in\mathbb{R}^{m\times m_a}$ for $h\in[H]$.\\
One can make similar assumptions for $\bfW_{Q_h}^{(0)}$, $\bfW_{K_h}^{(0)}$, and $\bfW_{V_h}^{(0)}$ as in Assumption \ref{asmp: VQK_initialization}. Note that $\{\bfp_1,\bfp_2,\cdots,\bfp_M\}$ needs to be changed into $\{\bfp_{h_1},\bfp_{h_2},\cdots,\bfp_{h_M}\}$, and the set $\{\bfp_{h_1},\bfp_{h_2},\cdots,\bfp_{h_M}\}$ can vary for different $h\in[H]$. It is also the same for $\{\bfq_{h_1},\bfq_{h_2},\cdots,\bfq_{h_M}\}$ and $\{\bfr_{h_1},\bfr_{h_2},\cdots,\bfr_{h_M}\}$ for different $h\in[H]$.\\
Based on the modified assumption with $H$ heads, the backbone of the proof remains the same. Lucky neurons in $\bfW_O$ tend to learn $(\bfp_{1_1}^\top,\bfp_{2_1}^\top,\cdots,\bfp_{H_1}^\top)^\top$ and $(\bfp_{1_2}^\top,\bfp_{2_2}^\top,\cdots,\bfp_{H_2}^\top)^\top$. Hence, the properties of the Relu activation are almost the same as the single-head case because luck neurons are still activated by either of two label-relevant patterns with a high probability. In fact, one can expect a more stable training process by multiple heads due to a more stable Relu gate for lucky neurons.

\subsection{Extension to skip connections and normalization}\label{subsec: skip_connection}
Consider a basic case where a skip connection is added after the self-attention layer. Let $m_a=d$. The network is changed into
\begin{equation}
\begin{aligned}
    F(\bfX^n)&=\frac{1}{|\mathcal{S}^n|}\sum_{l\in\mathcal{S}^n}\bfa_{(l)}^\top\text{Relu}(\bfW_O(\sum_{s\in\mathcal{S}^n}\bfW_{V}\bfx_s^n\text{softmax}({\bfx_s^n}^\top\bfW_{K}^\top\bfW_{Q}\bfx_l^n)+\bfx_l^n))\label{skip_connection}
\end{aligned}
\end{equation}
The assumption of $\bfW_V^{(0)}$ in Assumption \ref{asmp: VQK_initialization} should be changed into 
\begin{equation}
    \|(\bfW_V^{(0)}+\bfI)\bfmu_j-\bfp_j\|\leq \sigma,
\end{equation}
while the assumption of $\bfW_Q^{(0)}$ and $\bfW_K^{(0)}$ remain the same.\\
One can easily verify that the gradients of $\bfW_K$, $\bfW_Q$, and $\bfW_V$ for (\ref{skip_connection}) are almost the same as those for (\ref{eqn: attention}) except for the Relu gate. The major differences come from the gradient of $\bfW_O$, which also helps to determine the Relu gate. One needs to redefine 
\begin{equation}
\begin{aligned}
\bfV_l^n(t)&=\bfW_V^{(t)}\bfX^n\text{softmax}({\bfX^n}^\top{\bfW_K^{(t)}}^\top\bfW_Q^{(t)}\bfx_l^n)+\bfx_l^n\\
&=\sum_{s\in\mathcal{S}_1}\text{softmax}({\bfx_s^n}^\top{\bfW_K^{(t)}}^\top\bfW_Q^{(t)}\bfx_l^n)\bfp_1+\bfz(t)+\sum_{j\neq 1}W_j^n(t)(\bfx_j^n+\bfx_l^n)\\
&-\eta(\sum_{i\in \mathcal{W}_{l,n}(0)}V_i(t){\bfW_{O_{(i,\cdot)}}^{(t)}}^\top+\sum_{i\notin \mathcal{W}_{l,n}(0)}V_i(t)\lambda{\bfW_{O_{(i,\cdot)}}^{(t)}}^\top)
\end{aligned}
\end{equation}
for $l\in\mathcal{S}_1^n$. The inner product between the lucky neuron and the term $\sum_{j\neq 1}W_j^n(t)(\bfx_j^n+\bfx_l^n)$ can still be upper bounded by the inner product between the lucky neuron and the term $\sum_{s\in\mathcal{S}_1}\text{softmax}({\bfx_s^n}^\top{\bfW_K^{(t)}}^\top\bfW_Q^{(t)}\bfx_l^n)\bfp_1$ given good initialization of $\bfW_K$ and $\bfW_Q$. Therefore, (\ref{skip_connection}) can be analyzed following our proof techniques.\\
For layer normalization, one usually use that approach to normalize each data. It is consistent with our normalization of $\bfx_l^n$, which plays an important role in our proof. By normalization, the training process becomes more stable because of the unified norm of all tokens.

\end{document}